\documentclass{article}

    \usepackage[final,nonatbib]{neurips_2025}

\usepackage[utf8]{inputenc} %
\usepackage[T1]{fontenc}    %
\usepackage{hyperref}       %
\usepackage{url}            %
\usepackage{booktabs}       %
\usepackage{amsfonts}       %
\usepackage{nicefrac}       %
\usepackage{microtype}      %
\usepackage{xcolor}         %
\usepackage[table]{xcolor}
\usepackage[numbers]{natbib}
\usepackage{subcaption}
\usepackage{wrapfig}

\usepackage{amsmath}
\usepackage{amssymb}
\usepackage{mathtools}
\usepackage{amsthm}
\usepackage{bm}

\usepackage{multirow}
\usepackage{stmaryrd}  %
\usepackage{float}
\usepackage{algorithm}
\usepackage[noend]{algpseudocode}

\usepackage[capitalize,noabbrev]{cleveref}

\newcommand{\method}{\textsc{AutoGraph}}
\newcommand{\ie}{\textit{i.e.,\ }}
\newcommand{\eg}{\textit{e.g,\ }}
\newcommand{\st}{\textit{s.t.\ }}

\def\Tcal{{\mathcal{T}}}
\def\TNcal{{\mathcal{T}^\mathcal{N}}}
\def\Scal{{\mathcal{S}}}
\def\SNcal{\mathcal{S}^\mathcal{N}}

\def\Ncal{{\mathcal{N}}}

\def\concat{\mathbin\Vert}
\DeclareMathOperator*{\avg}{avg}

\newsavebox{\tablebox}

\definecolor{green}{RGB}{168, 213, 160}
\definecolor{red}{RGB}{244, 163, 165}

\usepackage[conf={end, restate, no link to proof}]{proof-at-the-end}
\theoremstyle{plain}
\newtheorem{theorem}{Theorem}[section]

\theoremstyle{definition}
\newtheorem{definition}[theorem]{Definition}

\theoremstyle{remark}

\title{Flatten Graphs as Sequences: \\Transformers are Scalable Graph Generators}

\author{%
  Dexiong Chen \\
  Max Planck Institute of Biochemistry\\
  Martinsried, Germany \\
  \texttt{dchen@biochem.mpg.de} \\
  \And
  Markus Krimmel \\
  Max Planck Institute of Biochemistry\\
  Martinsried, Germany \\
  \texttt{krimmel@biochem.mpg.de} \\
  \AND
  Karsten Borgwardt \\
  Max Planck Institute of Biochemistry\\
  Martinsried, Germany \\
  \texttt{borgwardt@biochem.mpg.de} \\
}

\begin{document}

\maketitle

\begin{abstract}

We introduce \method{}, a scalable autoregressive model for attributed graph generation using decoder-only transformers. By flattening graphs into random sequences of tokens through a reversible process, \method{} enables modeling graphs as sequences without relying on additional node features that are expensive to compute, in contrast to diffusion-based approaches. This results in sampling complexity and sequence lengths that scale optimally linearly with the number of edges, making it scalable and efficient for large, sparse graphs. A key success factor of \method{} is that its sequence prefixes represent induced subgraphs, creating a direct link to sub-sentences in language modeling. Empirically, \method{} achieves state-of-the-art performance on synthetic and molecular benchmarks, with up to 100x faster generation and 3x faster training than leading diffusion models. It also supports substructure-conditioned generation without fine-tuning and shows promising transferability, bridging language modeling and graph generation to lay the groundwork for graph foundation models. \\ Our code is available at \url{https://github.com/BorgwardtLab/AutoGraph}.
\end{abstract}

\section{Introduction}\label{sec:introduction}

Recent advancements in deep generative models have revolutionized various domains of artificial intelligence, demonstrating remarkable capabilities in generating complex data types such as images~\citep{rombach2022high}, natural language~\citep{gpt3,touvron2023llama,touvron2023llama2}, and audio~\citep{dhariwal2020jukebox,huang2024audiogpt}. These achievements have been primarily driven by the development of advanced architectures or methods such as transformers and diffusion models, alongside increasingly large-scale data resources. However, the generation of graph-structured data, which is fundamental to numerous scientific applications including drug discovery~\citep{vignac2023digress,lim2020scaffold}, protein design~\citep{ingraham2019generative}, and program synthesis~\citep{brockschmidt2019generative}, remains a significant challenge. This disparity primarily stems from the inherent complexity of preserving structural validity, maintaining invariance properties within graphs, and achieving scalability in real-world graph generation tasks.

To this end, diffusion-based models have emerged as a promising direction for graph generation, demonstrating effectiveness in synthesizing both classic unattributed graphs and molecules~\citep{jo2022score,vignac2023digress}. These approaches typically implement a denoising process in discrete graph space, simultaneously predicting edge connectivity and attributes. Yet, their practical applications are constrained by fundamental scalability limitations. The requirement for full adjacency matrix operations imposes quadratic memory complexity with respect to the number of nodes. Moreover, computing additional node features in each denoising step, such as spectral features, often involving cubic complexity, further increases the computational overhead.

Autoregressive approaches represent an alternative paradigm, constructing graphs sequentially by generating nodes and edges in a step-by-step manner~\citep{liao2019efficient,you2018graphrnn}. These models have demonstrated strong performance in generating small to medium-sized graphs by leveraging their ability to maintain structural validity through the generation process. Nevertheless, these models face inherent limitations: their sequences are not composed of tokens and thereby require specialized architectures, primarily based on recurrent neural networks, to process their complex ad-hoc sequential representations, preventing them from directly leveraging the remarkable advances in large language models (LLMs). Moreover, these specialized architectures often struggle with long-range dependencies and global structural consistency, leading to significantly inferior performance compared to recent diffusion models~\citep{vignac2023digress,jo2024grum}. This representational and architectural constraint not only limits their scalability but also creates a growing performance gap as general-purpose LLMs continue to advance rapidly.

In light of these challenges, we introduce a novel paradigm that bridges the gap between graph generation and LLMs through a graph-to-sequence transformation. Our approach advances previous random walk-based methods by representing graphs as sequences of tokens while maintaining their topological properties. Instead of requiring specialized architectures or operating directly on graph structures, we propose a method to linearize graphs into random sequences that encode local connectivity patterns. This transformation enables direct utilization of language models for graph generation while achieving optimally linear complexity with respect to the number of edges in both computational and memory requirements. Our approach effectively addresses the limitations of both diffusion-based and autoregressive methods: it maintains structural validity while enabling efficient scaling to large graphs and leveraging the powerful capabilities of modern language models.

Our work presents several technical contributions to the field of graph generation. (1) We introduce the concept of segmented Eulerian neighborhood trails (SENTs), a specialized class of Eulerian trails that permit breaks and incorporate neighborhood information. We establish sufficient conditions under which they can be employed for effective graph generation. (2) We propose an efficient flattening algorithm that transforms graphs into sequences and vice versa by sampling these SENTs, enabling lossless sequence representation of graphs. (3) Our method, termed \method{}, achieves state-of-the-art (SOTA) performance across diverse synthetic and molecular graph generation benchmarks, delivering a 100-fold generation and a 3-fold training speedup compared to diffusion-based models while maintaining the ability to scale to graphs of possibly immense size. (4) Additionally, \method{} demonstrates strong transfer learning capabilities and supports substructure-conditioned generation without additional fine-tuning. Our work not only advances the field of graph generation but also opens new avenues for applying LLMs to graph-centric tasks, paving the way for building foundation models for graphs.

\section{Methods}\label{sec:methods}

In this section, we present an approach to transforming graphs into sequences, enabling their modeling akin to natural language. Our method hinges on a specialized class of random trail segments that ensure complete graph coverage. We begin by introducing the concept of segmented Eulerian trails (SET) and demonstrate theoretically why this representation alone is insufficient for effective graph generation. Subsequently, we propose an extension of SET, namely the segmented Eulerian neighborhood trail (SENT), which additionally incorporates neighborhood information alongside the trails. We elucidate sufficient conditions for effective generation and develop an efficient sampling strategy to obtain such SENTs. The section concludes with extensions and discusses how to model the SENTs autoregressively using language models, thus bridging the gap between graph learning and language modeling paradigms. An overview of \method{} is illustrated in Figure~\ref{fig:overview}, and backgrounds and proofs are provided in Appendix~\ref{app:sec:remarks} and \ref{app:sec:proofs}.

\begin{figure}
    \centering
    \vskip -0.1in
    \includegraphics[width=\textwidth]{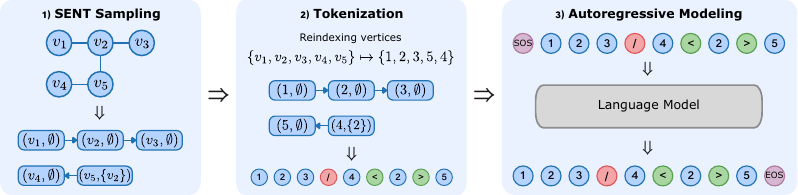}
    \caption{Overview of \method{}: (1) We use Algorithm~\ref{algo:sent_sampling} to sample a SENT $s$ from the input graph: $s=(s_1,s_2)$ with $s_1=((v_1,\emptyset),(v_2,\emptyset),(v_3,\emptyset))$ and $s_2=((v_5, \{v_2\}), (v_4,\emptyset))$. (2) We tokenize it by reindexing the vertices based on their first occurrence order in $s$ and adding special tokens (`\textbf{/}' represents breakage between segments, `$\bm{<}$' and `$\bm{>}$' indicate the start and end of a neighborhood set). (3) We perform the next token prediction on the tokenized sequences using a decoder-only transformer or any language model.}
    \label{fig:overview}
\end{figure}

\subsection{Segmented Eulerian Trail}
To formalize our approach, we begin by introducing fundamental concepts in graph theory.
Let $V$ be a set of vertices and $E:=V\times V$ a set of edges. A graph is defined as a tuple $G=(V_G, E_G)$, where $V_G\subseteq V$ is a finite set of vertices and $E_G\subseteq V_G\times V_G$ is the set of edges. For simplicity and without loss of generality, we restrict our attention to undirected graphs without isolated vertices, where each edge is represented as an unordered pair $(u,v)$ for $u,v \in V$. We begin by defining the concept of a trail in a graph:
\begin{definition}[Walk and trail]
    A walk is a sequence of nodes connected by edges in $G$ and a trail is a walk in which all edges are distinct. Given a graph $G$, the set of trails in $G$ is denoted as $\Tcal_G$.
\end{definition}
Next, we generalize the concept of trails beyond the context of a specific graph:
\begin{definition}[Generalized trail]
    A generalized trail of length $k$ is defined as a sequence of nodes $w:=(w_0,\dots,w_k)\in V^{k+1}$ for $k\geq 0$ \st $(w_{i-1},w_{i})\neq (w_{j-1},w_{j})$, $\forall i,j\in [k]$ and $i\neq j$.
\end{definition}
The set of all generalized trails is denoted as $\Tcal$, noting that $\Tcal_G\subseteq \Tcal$ for any $G$. For a generalized trail $w\in\Tcal$, we define $V_w\subseteq V$ and $E_w\subseteq E$ as the sets of vertices and edges traversed by $w$, respectively, termed the \emph{generated sets} of $w$.
An \emph{Eulerian trail} is a trail that visits every edge in a graph exactly once. Such trails are of particular interest as they capture the complete topology of the graph. However, the existence of an Eulerian trail depends on specific conditions related to vertex degrees and connectivity~\citep{biggs1986graph}. To generalize this concept to arbitrary graphs, we introduce the notion of trail segments:
\begin{definition}[Segmented Eulerian trail (SET)]
    A segmented Eulerian trail (SET) in $G$ is a sequence of trail segments such that each edge is visited exactly once across all segments, and segments do not need to be connected. 
    Formally, a SET of size $k$ in $G$ is defined as $s:=(s_1,\dots,s_k)$ \st $s_i\in \Tcal_G$, and the generated edge sets of its segments form a partition of $E_G$, \ie $\cup_{i=1}^k E_{s_i}=E_G$ and $E_{s_i}\cap E_{s_j}=\emptyset, \forall i,j\in [k], i\neq j$. 
    Similarly, a SET (without relying on a specific graph) is defined as a sequence of generalized trails whose generated edge sets are disjoint.
\end{definition}
The set of all SETs in $G$ is denoted as $\Scal_G$, and the set of all SETs is denoted as $\Scal$.
For a SET $s=(s_i)_{i=1}^k$, we define the \emph{generated node and edge sets} as $V_s:=\cup_{i=1}^k V_{s_i}$ and $E_s:=\cup_{i=1}^k E_{s_i}$. The graph $G_s:=(V_s,E_s)$ is termed \emph{generated graph} of $s$. It is easy to show that $s$ is a SET in $G$ if $G_s\simeq G$.
Moreover, SETs can be classified into equivalence classes based on graph isomorphism, as formalized below:
\begin{definition}[SET isomorphism]
    For any two SETs $s,t\in \Scal$, we say they are isomorphic $s\simeq t$ if there is a bijection $\pi:V_s\to V_t$ between their generated node sets and $\pi(s)=t$ where $\pi$ applies elementwise to all nodes in $s$.
\end{definition}
This isomorphism partitions $\Scal$ into equivalence classes. Moreover, we have the following relationship between SETs and graphs, relevant for our tokenization (Sec.~\ref{sec:tokenization}):
\begin{theoremEnd}{theorem}\label{thm:set_isomorphism}
    For any SETs $s,t\in \Scal$, their generated graphs are isomorphic, \ie $G_s\simeq G_t$, if $s\simeq t$. Conversely, if two graphs $G\simeq H$, then for any SET $s\in\Scal_G$, there exists a SET $t\in\Scal_H$ \st $s\simeq t$.
\end{theoremEnd}
\begin{proofEnd}
    By definition of the isomorphism between $s$ and $t$, there exists a bijection $\pi:V_s\to V_t$ \st $\pi(s)=t$. Now if $u, v\in V_s$ are adjacent in $G_s$, \ie $(u,v)\in E_s$, then $(\pi(u),\pi(v))$ is an edge visited by $\pi(s)=t$, thus $(\pi(u),\pi(v))\in E_t$. Similarly, the reverse is also true. Consequently, $G_s\simeq G_t$.

    Now assume that $G\simeq H$ with an isomorphism $\pi$ and $s\in\Scal_G$. It is easy to show that $\pi(s)$ is also a SET and its generated graph $G_{\pi(s)}=H$. By taking $t=\pi(s)$, we obtain the result.
\end{proofEnd}
While a SET in $G$ fully characterizes its structure, we show below the prefixes of the SET do not necessarily describe the substructures of $G$, a critical property for effective autoregressive graph generation.
\begin{definition}[Flattening]
    The flattening of a sequence of sequences $s$ is the concatenation of all its sequences, denoted as $\concat s$.
\end{definition}
\begin{definition}[Prefix of a SET]
    For $s\in \Scal$, we call $t$ a prefix of $s$ if $\concat t$ is a prefix of $\concat s$.
\end{definition}
\begin{theoremEnd}{lemma}
    For any graph $G$ and SET $s$ in $G$, the generated graph of any prefix of $s$ is a subgraph of $G$, but not necessarily an induced subgraph.
\end{theoremEnd}
\begin{proofEnd}
    Assume that $t$ is a prefix of $s$. Then $V_t\subseteq V_s=V_G$ and $E_t\subseteq E_s$. However, $G_t$ is not necessarily an induced subgraph of $G_s$. We consider the following counter-example: $s=((1,2,3,4,1,3))$, $V_s=\{1,2,3,4\}$, and $E_s=\{(1,2),(2,3),(3,4),(1,4),(1,3)\}$. Let $t=((1,2,3,4,1))$. $t$ is clearly a prefix of $s$, but its generated graph is not an induced subgraph of $G_s$ as its generated edge set does contain $(1,3)$.
\end{proofEnd}
This result motivates us to extend the definition of generalized trails to incorporate the full structural information of the \emph{induced subgraphs}, rather than arbitrary subgraphs, to constrain the generation space better and address long-range dependency challenges. 
Without this extension, dependencies between neighboring nodes may span a long sequence of generation steps, making it more difficult for the model to learn such dependencies. Empirically, we show that SET fails to generate structurally valid graphs in Section~\ref{sec:ablation_experiments}.

\subsection{Segmented Eulerian Neighborhood Trail}
To make the prefixes of a SET encode richer information, we need to extend SET to contain neighborhood information in a graph. Thus, we consider the following definitions:
\begin{definition}[Neighborhood sequence]
    A neighborhood sequence is a sequence of tuples $w:=(w_0,\dots,w_k)$ where $w_i=(v_i,A_i)$ with a node $v_i\in V$ and a neighborhood set $A_i\subseteq V$, $\forall i\in \{0,\dots,k\}$. $w$ is called Hamiltonian if its node sequence $n(w):=(v_0,\dots,v_k)$ has non-repeated elements. $w$ is called \emph{causal} if $A_i$ only contains visited nodes, \ie $A_i\subseteq \{v_0,\dots,v_{i-1}\}$ $\forall i\in [k]$.
\end{definition}
\begin{definition}[Neighborhood trail]
    A neighborhood trail is a neighborhood sequence that satisfies two conditions. (i) $n(w)$ is a generalized trail. (ii) If we define the generated edge set of $w_i$ as $E_{w_i}=\{(v_i,u)\, |\, u\in A_i\}$, the family $\{ E_{n(w)}, E_{w_1},\dots, E_{w_k}\}$ is pairwise disjoint. Its union is called the generated edge set of $w$.
\end{definition}
The set of all neighborhood trails is denoted by $\TNcal$. For any $w\in\TNcal$, we denote by $G_w:=(V_w,E_w)$ the generated graph of $w$ where $V_w:=(\cup_{i=1}^k A_i)\cup V_{n(w)}$ is the generated node set and $E_w$ is the generated edge set. Note that a generalized trail is a neighborhood trail with $A_i=\emptyset, \forall i$. We extend SET to incorporate neighborhood information:
\begin{definition}[Segmented Eulerian neighborhood trail (SENT)]
    A segmented Eulerian neighborhood trail (SENT) of size $k$ is a sequence of neighborhood trails $s:=(s_1,\dots,s_k)$ with pairwise disjoint generated edge sets, \ie $s_i\in\TNcal$ and $E_{s_i}\cap E_{s_j}=\emptyset, \forall i,j\in [k], i\neq j$. 
\end{definition}
Similarly to SETs, the generated graph of a SENT $s$ is denoted by $G_s=(V_s,E_s)$. If a graph $G\simeq G_s$, we say that $s$ is a SENT in $G$. We denote by $\SNcal$ and $\SNcal_G$ the set of SENTs and SENTs in $G$. Analogously to SETs, we define an isomorphism over $\SNcal$ and obtain the same relationship as in Thm.~\ref{thm:set_isomorphism}. A prefix of a SENT is defined similarly to that of a SET. We give below conditions to force generated graphs of prefixes of a SENT to be induced subgraphs.
\begin{definition}[Causal SENT]
    A SENT $s$ is called causal if its flattening $\concat s$ is causal. 
\end{definition}
\begin{definition}[Hamiltonian and semi-hamiltonian SENT]
    A SENT $s$ is called Hamiltonian if its flattening $\concat s$ is Hamiltonian. $s$ is called semi-hamiltonian if $s$ is Hamiltonian, or for any nodes visited more than once, their occurrences after the first time should be in a start tuple of a neighborhood trail and their associated neighborhood sets are empty.
\end{definition}

\begin{theoremEnd}{theorem}\label{thm:induced_subgraph}
    For any causal SENT $s\in\SNcal$, the generated graph of any prefix $t$ of $s$ is an induced subgraph of $G_s$ if and only if $s$ is semi-hamiltonian. In this case, $s$ is called subgraph-induced. %
\end{theoremEnd}
\begin{proofEnd}
    Let us first introduce some notations.
    We denote by $R_s$ the sequence of the start tuples across all neighborhood trails in $s$, which is also a neighborhood sequence. By definition of semi-hamiltonian, the occurrences after the first time of a node in $s$ should be in $R_s$. We denote by $n(s)$ the associated node sequence of SENT $s$, \ie $n(s):=n(\concat s)$.

    Let us first assume that $s$ is semi-hamiltonian.
    
    Assume that $t$ is a prefix of $s$. It is easy to show that $G_t$ is a subgraph of $G_s$. Now assume that $u,v\in V_t$ \st $(u,v)\in E_s$, we want to show that $(u,v)\in E_t$. There are two cases:

    1) Assume that $u,v\in n(t)$. Since $s$ is semi-hamiltonian, $n(s)\setminus n(t)$ either does not contain $u$ or $v$, or even if one of them, say $u\in n(s)\setminus n(t)$, we have $u\in n(R_s)$ and its associated neighborhood set is empty. In both cases, the edge $(u,v)$ does not belong to the generated edge set of the neighborhood subsequence after $\concat t$. By the disjointness of the generated edge sets of $s$, it can only be included in the generated edge set of $t$, we thus have $(u,v)\in E_t$.

    2) Assume that one of them, say $u\notin n(t)$. There exists a neighborhood set $A$ in a tuple of $\concat t$ such as $u\in A$. Since $t$ is causal, we have $u\in n(t)$ which contradicts the assumption.

    In all the above cases, we have $(u,v)\in E_t$.

    Now let us assume that the generated graph of any prefix of $s$ is an induced subgraph of $G_s$.

    Let us prove that $s$ is semi-hamiltonian by contradiction. Assume that there exist two tuples in $\concat s$ with the same nodes $s_i=(v, A_i)$ and $s_j=(v,A_{j})$ with $i<j$. There are two cases: 1) $s_j\notin R_s$. A tuple $(u,A_u)$ exists one step before $s_j$ in the same neighborhood trail. We consider the prefix $t$ ending at $(u, A_u)$. We have $v,u\in V_t$ and $(u,v)\in E_s$, but $(u,v)\notin E_t$, by the disjointness of $s$ and since $(u,v)$ is visited at $s_j$ after $t$. 2) $s_j\notin R_s$ and $A_j\neq\emptyset$. Since $s$ is causal, there exists $s_u:=(u,A_u)$ before $s_j$ \st $u\in A_j$. We consider the prefix $t$ ending at exactly this tuple. We have $u,v\in V_t$ and $(u,v)\in E_s$, but $(u,v)\notin E_t$, by the disjointness of $s$ and since $(u,v)$ is an edge visited at $(v,A_j)$ after $t$.
\end{proofEnd}
Now let us find the conditions for a causal and Hamiltonian SENT. For any SENT $s$ and a tuple $w:=(v,A)$ in $s$, we denote by $V_s(w)$ the set of nodes visited by $s$ before $w$, excluding the node linked to $v$ through the trail if it exists. We have the following necessary and sufficient conditions:
\begin{theoremEnd}{theorem}\label{thm:causal_hamiltonian_sent}
    For $s\in\SNcal_G$, $s$ is causal and Hamiltonian if and only if every tuple $w:=(v, A_v)$ in $\concat s$ satisfies $A_v=\Ncal_G(v)\cap V_s(w)$. In this case, every node is visited exactly once. Moreover, $s$ is causal and semi-hamiltonian if and only if every tuple $w:=(v, A_v)$ in $s$ satisfies either $A_v=\Ncal_G(v)\cap V_s(w)$ or $A_v=\emptyset$.
\end{theoremEnd}
\begin{proofEnd}
    Let us first assume that for any tuple $w:=(v, A_v)$ in $\concat s$, $A_v=\Ncal_G(v)\cap V_s(w)$.
    Since $A_v\subseteq V_s(w)$ which is a subset of the set of visited nodes, $s$ is causal. Now we prove $s$ is Hamiltonian by contradiction. Assume that there exist two tuples in $\concat s$, $s_u:=(u,A_u)$ and a later visited one $s_v:=(v,A_v)$ \st $u=v$. Then, $A_v=\Ncal_G(v)\cap V_s(s_v)=\Ncal_G(u)\cap V_s(s_v)$ should contain the node visited before that is a neighbor of $u$ (either through a trail or the neighborhood set of $u$), denoted by $u'$. Thus, the edge $(u,u')$ has been visited twice, which contradicts the disjointness of $s$.

    Assuming that $A_v=\Ncal_G(v)\cap V_s(w)$ or $A_v=\emptyset$ for any tuple $(v, A_v)$ in $s$, we can also prove $s$ is semi-hamiltonian by contradiction. Assume that there exist two tuples in $\concat s$, $s_u:=(u,A_u)$ and a later visited one $s_v:=(v,A_v)$ \st $u=v$ and $A_v\neq \emptyset$. Then, $A_v=\Ncal_G(v)\cap V_s(s_v)=\Ncal_G(u)\cap V_s(s_v)$ by assumption. And using the same argument as above, we have a contradiction.

    Now assume that $s$ is causal and Hamiltonian. Let us prove the other direction by contradiction. There exists a tuple $w:=(v,A_v)$ in $s$ \st $A_v\neq \Ncal_G(v)\cap V_s(w)$. As $s$ is causal, $A_v\subseteq V_s(w)$. $A_v\subseteq \Ncal_G(v)$ as $s\in\SNcal_G$. Thus, $A_v\subset \Ncal_G(v)\cap V_s(w)$, which means that there exists $u\in \Ncal_G(v)\cap V_s(w)$ and $u\notin A_v$. Hence, $(u, v)\in E_G$ and $u$ is visited before $v$. However, as $u\notin A_v$, $(u, v)\in E_G$, and $s$ is Hamiltonian, there exists a tuple $(u, A_u)$ in $\concat s$ \st $v\in A_u$. By causality of $s$, $v$ is visited before $u$, which contradicts the fact that $s$ is Hamiltonian.

    Assuming that $s$ is causal and semi-hamiltonian. Let us prove the other direction by contradiction. There exists a tuple $w:=(v,A_v)$ in $s$ \st $A_v\neq \Ncal_G(v)\cap V_s(w)$ and $A_v\neq \emptyset$. Using the same arguments as above, there exists $(u, v)\in E_G$, and $u$ is visited before $v$. However, as $u\notin A_v$ and $(u, v)\in E_G$, $s$ should visit the edge $(u,v)$ at some point. Since $s$ is semi-hamiltonian, if $s$ visits again $u,v$ they can only be the first nodes and their associated neighborhood sets are empty. Hence, there is no means for $s$ to visit $(u,v)$ after $v$, leading to a contradiction.
\end{proofEnd}
This theorem offers a simple sufficient condition for subgraph-induced SENTs. We provide an implementation in the following through a random path sampling strategy. 

\subsection{Sampling Algorithm for SENT}
Thm.~\ref{thm:causal_hamiltonian_sent} offers a simple strategy to sample a causal and Hamiltonian SENT: one needs to traverse the graph and choose the neighborhood set as all neighbors of the current node that have been visited. The traversing strategy could be achieved through a random path sampling or a depth-first search. In Algorithm~\ref{algo:sent_sampling}, we provide a sampling strategy based on random path sampling with breaks.
\textbf{Complexity analysis.}
The length of a SENT, including the sizes of neighborhood sets (in other words, tokenized SENT defined in Section~\ref{sec:tokenization}), is bounded by the number of nodes and edges, as each node and edge can be visited exactly once. Therefore, both the time and space complexity of sampling a SENT from graph $G$ are $\mathcal{O}(m)$ where $m$ is the number of edges. 

\subsection{Tokenization of SENT}\label{sec:tokenization}
\begin{wrapfigure}{r}{.55\textwidth}
\begin{minipage}{.55\textwidth}
\vspace{-.3in}
\begin{algorithm}[H]
    \caption{Causal and Hamiltonian SENT Sampling}\label{algo:sent_sampling}
    \begin{algorithmic}[1]
        \Require $G=(V, E)$
        \Ensure A SENT $s$ in $G$
        \State Set of unvisited nodes $U \gets V$
        \State $s \gets []$
        \State $v \gets \texttt{RandomSample}(U)$; $U \gets U\setminus \{v\}$
        \State $t \gets [(v,\emptyset)]$ \Comment{first neighborhood trail}
        \While{$U \neq \emptyset$}
        \If{$\Ncal_G(v)\cap U=\emptyset$} \Comment{start a new trail}
        \State $s.\texttt{append}(t)$
        \State $v \gets \texttt{RandomSample}(U)$; $U \gets U\setminus \{v\}$
        \State $A \gets \Ncal_G(v)\cap (V\setminus U)$
        \State $t \gets [(v, A)]$
        \Else \Comment{sample the next node in the trail}
        \State $u \gets \texttt{RandomSample}(\Ncal_G(v)\cap U)$
        \State $U \gets U\setminus \{u\}$
        \State $A \gets (\Ncal_G(u)\setminus \{v\})\cap (V\setminus U)$
        \State $t.\texttt{append}((u, A))$
        \State $v \gets u$
        \EndIf
        \EndWhile
    \end{algorithmic}
\end{algorithm}
\vspace{-.3in}
\end{minipage}
\vspace{-.1in}
\end{wrapfigure}
Previous works have explored related concepts of sequences in graphs. For example,
\citet{you2018graphrnn} investigated causal Hamiltonian neighborhood sequences generated through breadth-first search, while \citet{liao2019efficient,goyal2020graphgen} constructed SENT-like sequences using depth-first search. However, neither of these works interpreted these sequences as a language. Here, we present a method to bridge the gap between graph generation and language modeling.

The tokenization process starts by mapping all isomorphic SENTs to the same sequence, by reindexing the vertices according to their first occurrence order within the sequence. Specifically, if we denote this ordering function for a SENT $s$ by $\pi: V_s\to \{1,\dots,|V_s|\}$, $s$ is then replaced with its ordered representation $\pi(s)$. Thanks to the isomorphism property of SENT (Thm.~\ref{thm:set_isomorphism}), $\pi(s)$ generates a graph isomorphic to $G_s$ while ensuring the obtained sequence is invariant to the node ordering of the input graph.

To convert an (ordered) SENT into a machine-readable sequence, we tokenize it into a sequence of indices using special tokens. These tokens include symbols such as ` \textbf{/} ' to indicate a breakage between segments, and `$\bm{<}$' and `$\bm{>}$' to mark the start and end of a neighborhood set. Specifically, for any $s:=(s_1,\dots, s_k)\in\SNcal$, we define the tokenization function \texttt{Token} as follows:
\begin{equation*}
    \texttt{Token}(s):=\texttt{Token}(s_1)\concat [~\textbf{/}~] \concat 
    \cdots \concat [~\textbf{/}~] \concat \texttt{Token}(s_k),\text{ where } \texttt{Token}(s_i):=\concat_{w\in s_i} \texttt{Token}(w),
\end{equation*}
and for each tuple $w:=(v,A)$ with the \emph{sorted set} $A= \{u_1,\dots,u_p\}$ (due to the reindexing by $\pi$), we define:
\begin{equation*}
    \texttt{Token}(w):=\left[v, \bm{<}, u_1,\dots, u_p, \bm{>}\right].
\end{equation*}
This process converts a SENT into a sequence of tokens that a language model can effectively model. Using an equivalent form, the resulting tokenization induces a \emph{non-Markovian} random walk in the graph, incorporating additional virtual nodes labeled with the above special tokens (see Appendix~\ref{app:sec:random_walk_interpretation} for more details). \emph{Language modeling of SENTs aims to learn the state transition probabilities}.

\subsection{Extension to Attributed Graphs}
Our method can be easily extended to graphs with categorical (or discretized) attributes by inserting node and edge attributes in an interleaved fashion into the tokenized SENT sequence. Specifically, let $L_{\mathrm{node}}(v)$ and $L_{\mathrm{edge}}(u,v)$ be the attributes of a node $v$ and an edge $(u,v)$ respectively. Using the same notation as above, we define for any $s_i:=(w_1,\dots, w_q)\in\TNcal$ with $w_i=(v_i,\cdot)$:
\begin{equation*}
    \begin{aligned}
        \texttt{Token}(s_i)&:=\texttt{Token}(w_1) \concat [L_{\mathrm{edge}}(v_1,v_2)] \concat \texttt{Token}(w_2) \concat \dots \concat \texttt{Token}(w_q),  \\
        \texttt{Token}(w)&:=[v, L_{\mathrm{node}}(v), \bm{<}, L_{\mathrm{edge}}(v,u_1), u_1, \dots, L_{\mathrm{edge}}(v,u_p),u_p, \bm{>}].
    \end{aligned}
\end{equation*}

\subsection{Autoregressive Modeling of Tokenized SENTs}
The sampling and tokenization of SENTs in graphs allows for transforming graphs into sequences, which could be modeled by language models. Specifically, given a graph $G$ represented as a SENT $s$, which consists of a sequence of tokens $(s_1,\dots, s_n)$, a standard language modeling objective is to maximize the following log-likelihood:
\begin{equation}
    p(s)=\sum_{i=1}^n \log p_\theta(s_i\,|\, s_1,\dots, s_{i-1}),
\end{equation}
where the conditional probability $p_\theta$ is modeled using a neural network with parameters $\theta$. The architecture of the neural network can be any state-of-the-art sequence model.

\section{Related Work}
\vspace{-.005in}
\textbf{Autoregressive models for graph generation.}
Autoregressive models generate graphs by sequentially adding nodes and edges. GraphRNN~\citep{you2018graphrnn} pioneered this approach by framing graph generation as a sequence prediction task, demonstrating the capacity of recurrent neural networks~(RNNs)~\citep{chung2014empirical} to capture complex structures. DeepGMG~\citep{li2018learning} introduced a probabilistic policy framework for conditional generation, while GRAN~\citep{liao2019efficient} and BiGG~\citep{dai2020bigg} enhanced efficiency and scalability by generating multiple nodes and edges in parallel. ANFM~\citep{krimmel2025towards} leverages filtration to improve efficiency.

Recent research has focused on optimizing the generation order.  \citet{chen2021order} highlighted that the ordering of node and edge additions impacts graph quality, and GraphARM~\citep{kong2023autoregressive} applied reinforcement learning to dynamically refine this order. \citet{goyal2020graphgen} incorporated logical constraints to improve domain-specific generation, and \citet{bacciu2020edge} proposed Bayesian reasoning to better capture graph dependencies. BwR~\citep{diamant2023improving} and GEEL~\citep{jang2024simple} investigated node ordering based on optimized bandwidth.

These models, while efficient on synthetic datasets, do not explicitly represent graphs as token sequences, preventing direct application of LLM techniques. More significantly, their sequences are not guaranteed to be subgraph-induced (see Thm.~\ref{thm:induced_subgraph}). In contrast, our approach enables substructured-conditioned generation analogous to prompt-based generation in language models, establishing a more fundamental connection between graph and language modeling paradigms.

\textbf{Other graph generative models.}
Other graph generative models include variational, GAN-based, flow-based, and diffusion-based approaches. GraphVAEs~\citep{kipf2016variational,simonovsky2018graphvae} employ variational autoencoders to learn latent representations, effectively generating small graphs but struggling with more complex structures. GAN-based models, such as NetGAN~\citep{bojchevski2018netgan} and SPECTRE~\citep{martinkus2022spectre}, generate graphs by modeling graph descriptors like random walks and spectral features. Flow-based methods such as~\citep{shi2020graphaf} have shown the ability to generate small molecular graphs.

Diffusion-based models iteratively refine noise into structured graphs through reverse diffusion steps. Continuous diffusion models~\citep{niu2020permutation,jo2022score} adapt denoising diffusion probabilistic models for graph generation. To leverage graph sparsity and structure, discrete diffusion models~\citep{vignac2023digress,kong2023autoregressive,jo2024grum,xu2024discrete} have been developed. However, a key challenge for these models is the slow sampling process due to the long reverse diffusion chain. To mitigate this limitation, several efficient diffusion techniques have been proposed, including EDGE~\citep{chen2023efficient}, HiGen~\citep{karami2024higen}, ESGG~\citep{bergmeister2024efficient}, and Pard~\citep{zhao2024pard}.

\textbf{Random walks for graph learning.}
Random walks have been widely used in graph learning due to their strong expressive power. GCKN~\citep{chen2020convolutional} and RWGNN~\citep{nikolentzos2020random} utilize path and walk kernels to learn graph representations. Several recent works~\citep{ivanov2018anonymous,wang2021inductive,yin2022algorithm} explicitly integrate random walk sequences with positional encodings, inspiring subsequent methods such as CRaWL~\citep{tonshoff2023crawl}, NeuralWalker~\citep{chen2024neuralwalker} and RWNN~\citep{kim2024revisiting}. GraphGPT~\citep{zhao2023graphgpt} leverages Eulerian paths to improve graph property prediction. 
Some graph transformers~\citep{mialon2021graphit,chen2022structure} also leverage features based on random walks. Moreover, graph-to-sequence representations have been used to assist LLMs in understanding graphs~\citep{fatemi2024talk,chen2024llaga}. 
Our work explores sequence representations of graphs for graph generation, introducing a novel perspective on combining random walks and language modeling for scalable graph generation.

\section{Experiments}\label{sec:experiments}
In this section, we evaluate the performance of \method{} on several graph generation benchmarks, including both small and large graphs, and synthetic and real-world molecular datasets. Our experiments compare its performance to several SOTA methods and particularly focus on evaluating the following aspects: (1) We show its ability to generate relatively small graphs with a 100-fold inference speedup compared to diffusion-based models while maintaining or even improving structural validity. (2) We show its ability to scale to large graphs without loss of performance. (3) We demonstrate its effectiveness in generating real-world graphs with attributes with a focus on molecular generation, outperforming SOTA diffusion models. (4) We showcase its strong transfer capabilities and its ability to perform substructure-conditioned generation without any additional fine-tuning.
Additional details on experimental settings and evaluation are provided in Appendix~\ref{app:sec:experimental_details}.

\begin{table}[tbp]
    \centering
    \vskip -0.1in
    \caption{Benchmarking \method{} on Planar and SBM}
    \label{tab:planar_sbm}
    \begin{small}
    \begin{sc}
    \resizebox{\textwidth}{!}{
    \begin{tabular}{lcccccc|cccccc}\toprule
         & \multicolumn{6}{c}{Planar Graphs} & \multicolumn{6}{c}{Stochastic Block Models} \\ 
         & \multicolumn{6}{c}{$n_{\mathrm{graphs}}=128$, $|V|=64$} & \multicolumn{6}{c}{$n_{\mathrm{graphs}}=128$, $|V|_{\max}=187$, $|V|_{\avg}\approx 104$}  \\ \cmidrule{2-13}
        Model & Deg. & Clus. & Orbit & Spec. & Ratio & VUN & Deg. & Clus. & Orbit & Spec. & Ratio & VUN \\ \midrule
        Training set &  0.0002 & 0.0310 & 0.0005 & 0.0038 & 1.0 & -- & 0.0008 & 0.0332 & 0.0255 & 0.0027 & 1.0 & --  \\ \midrule
        GraphRNN~\cite{you2018graphrnn} &  0.0049 & 0.2779 & 1.2543 & 0.0459 & 638.5 & 0.0 & 0.0055 & 0.0584 & 0.0785 & 0.0065 & 3.5 & 5.0 \\
        GRAN~\citep{liao2019efficient} & 0.0007 & 0.0426 & 0.0009 & 0.0075 & 2.1 & 0.0 & 0.0113 & 0.0553 & 0.0540 & 0.0054 & 5.0 & 25.0 \\
        SPECTRE~\citep{martinkus2022spectre} &  0.0005 & 0.0785 & 0.0012 & 0.0112 & 2.6 & 25.0 & 0.0015 & 0.0521 & 0.0412 & 0.0056 & 1.8 & 52.5 \\
        EDGE~\citep{bacciu2020edge} &  0.0761 & 0.3229 & 0.7737 & 0.0957 & 490.9 & 0.0 & 0.0279 & 0.1113 & 0.0854 & 0.0251 & 12.7 & 0.0 \\
        GraphGen~\citep{goyal2020graphgen} & 0.0328 & 0.2106 & 0.4236 & 0.0430 & 257.3 & 7.5 & 0.0550 & 0.0623 & 0.1189 & 0.0182 & 20.5 & 5.0 \\
        BiGG~\citep{dai2020bigg} & 0.0007 & 0.0570 & 0.0367 & 0.0105 & 20.4 & 5.0 & 0.0012 & 0.0604 & 0.0667 & 0.0059 & 2.0 & 10.0 \\
        DiGress~\citep{vignac2023digress} &  0.0007 & 0.0780 & 0.0079 & 0.0098 & 6.1 & 77.5 & 0.0018 & \textbf{0.0485} & \textbf{0.0415} & 0.0045 & 1.8 & 60.0 \\ 
        GruM~\citep{jo2024grum} & 0.0005 & \textbf{0.0353} & 0.0009 & \textbf{0.0062} & 1.8 & 90.0 & \textbf{0.0007} & 0.0492 & 0.0448 & 0.0050 & \textbf{1.5} & 85.0 \\
        GEEL~\citep{jang2024simple} & 0.0039 & 0.0013 & 0.0062 & 0.0234	& 9.5 & 0.0 & 0.0106 & 0.0616 & 0.0023 & 0.0381 & 7.3 & 5.0 \\
        ESGG~\citep{bergmeister2024efficient} & 0.0005 & 0.0626 & 0.0017 & 0.0075 & 2.5 & \textbf{95.0} & 0.0119 & 0.0517 & 0.0669 & 0.0067 & 5.4 & 45.0 \\ \midrule
        \method{} & \textbf{0.0004} & 0.0605 & \textbf{0.0003} & 0.0064 & \textbf{1.5} & 87.5 & 0.0077 & 0.0519 & 0.0439 & \textbf{0.0040} & 3.4 & \textbf{92.5} \\
        \bottomrule
    \end{tabular}
    }
    \end{sc}
    \end{small}
\end{table}

\textbf{Implementation details.}
We employ the LLaMA model with 12 layers and a hidden dimension of 768 as our sequence model backbone across all experiments, aligning with the architecture of GPT-2's smallest variant~\citep{radford2019gpt2}. Although prior works have used smaller models, we argue that our approach still demonstrates better scalability and faster training and inference speeds compared to diffusion models. For inference, we adopt the commonly used top-k sampling strategy~\citep{fan2018topk}. Our implementation leverages the Hugging Face framework~\citep{jain2022huggingface}, providing users with a flexible interface to experiment with SOTA language models for graph generation.

\textbf{Evaluation.}
For fair comparison, we align our evaluation methodology with established practices from prior works~\citep{you2018graphrnn,martinkus2022spectre,vignac2023digress}. Our evaluation compares generated samples against the test set using maximum mean discrepancy (MMD)~\citep{gretton2012kernel}, computed across multiple graph descriptors: node degree distributions (\textsc{Deg.}), clustering coefficients (\textsc{Clus.}), orbit count statistics (\textsc{Orbit}), and eigenvalue spectra (\textsc{Spec.}). 
As a reference, we also compute MMDs between the training and test sets and report the average ratio between generated and training MMDs (\textsc{Ratio}) following~\citet{bergmeister2024efficient}.

For synthetic datasets, we additionally assess model performance using the VUN metric, namely the proportion of generated graphs that are simultaneously valid, unique, and novel compared to the training graphs. Our efficiency analysis includes two measurements: inference speed, calculated as the per-graph generation time when producing 1024 graphs, and training speed, measured as the time required to achieve a VUN score of 75.0 for the Planar dataset and 60.0 for the SBM dataset. All efficiency measurements are performed on a single NVIDIA H100 GPU.

For molecular generation datasets, we strictly follow the evaluation metrics used in DiGress~\citep{vignac2023digress} and use the evaluation tools from the official codebase~\citep{polykovskiy2020moses,brown2019guacamol}. More details are provided in App.~\ref{app:sec:evaluation_metrics}.

\subsection{Comparison to State-of-the-Art Methods}\label{sec:compare_sota}
We evaluate the performance of \method{} compared to other SOTA graph generative models using the standard setting without pre-training. 
The comparison partners include GraphRNN~\citep{you2018graphrnn}, GRAN~\citep{liao2019efficient}, SPECTRE~\citep{martinkus2022spectre}, EDGE~\citep{chen2023efficient}, GraphGen~\citep{goyal2020graphgen}, BiGG~\citep{dai2020bigg}, DiGress~\citep{vignac2023digress}, GruM~\citep{jo2024grum}, GEEL~\citep{jang2024simple}, and ESGG~\citep{bergmeister2024efficient}.

\textbf{Small synthetic graph generation.}
We first evaluate our method on the small synthetic graph datasets introduced by~\citet{martinkus2022spectre}, including the Planar and SBM datasets. 
As shown in Table~\ref{tab:planar_sbm}, \method{} demonstrates competitive MMDs while ranking second-best and best in terms of VUN scores on the Planar and SBM datasets, respectively. Importantly, only GruM and \method{} exhibit strong structural validity (VUN $\geq 80.0$) on the SBM dataset. Previous state-of-the-art autoregressive models, particularly GEEL, completely fail on both datasets in terms of VUN scores, as they largely memorize some subset of the training data. It is worth noting that the relatively low MMD ratio of \method{} is expected, as we selected the best model based on the VUN score. More results with error bars are provided in App.~\ref{app:sec:additional_results_synthetic}.

Additionally, we assess the training and inference times of \method{} against representative models, including DiGress, GRAN, and ESGG. As presented in Table~\ref{tab:time_comparison}, \method{} is approximately 3 times faster during training and 100 times faster during inference compared to diffusion-based models. This substantial speedup over diffusion-based models is even more pronounced than that observed in other data modalities such as images~\citep{tian2024var}.

\begin{table*}[tbp]
    \centering
    \caption{Benchmarking \method{} on Proteins and Point Clouds. OOM indicates out of memory. Note that for GEEL~\citep{jang2024simple}, we fail to reproduce their experiments on the Point Clouds dataset using their official codebase.}
    \label{tab:protein_and_pointcloud}
    \begin{small}
    \begin{sc}
    \resizebox{.9\textwidth}{!}{
    \begin{tabular}{lccccc|ccccc}\toprule
         & \multicolumn{5}{c}{Proteins} &  \multicolumn{5}{c}{Point Clouds} \\ %
         & \multicolumn{5}{c}{$n_{\mathrm{graphs}}=587$, $|V|_{\max}=500$, $|V|_{\avg}\approx 258$} & \multicolumn{5}{c}{$n_{\mathrm{graphs}}=26$, $|V|_{\max}=5037$, $|V|_{\avg}\approx 1332$} \\ \cmidrule{2-11}
        Model & Deg. & Clus. & Orbit & Spec. & Ratio & Deg. & Clus. & Orbit & Spec. & Ratio \\ \midrule
        Training set & 0.0003 & 0.0068 & 0.0032 & 0.0005 & 1.0 & 0.0000 & 0.1768 & 0.0049 & 0.0043 & 1.0 \\ \midrule
        GraphRNN~\citep{you2018graphrnn} & 0.0040 & 0.1475 & 0.5851 & 0.0152 & 62.1 & OOM & OOM & OOM & OOM & OOM \\
        GRAN~\citep{liao2019efficient} & 0.0479 & 0.1234 & 0.3458 & 0.0125 & 77.7 & 0.0201 & 0.4330 & 0.2625 & 0.0051 & 19.1 \\
        SPECTRE~\citep{martinkus2022spectre} & 0.0056 & 0.0843 & 0.0267 & 0.0052 & 12.5 & OOM & OOM & OOM & OOM & OOM \\
        EDGE~\citep{bacciu2020edge} & 0.1863 & 0.3406 & 0.6786 & 0.1075 & 274.5 & 0.4441 & 0.3298 & 1.0730 &  0.4006 & 104.7 \\
        GraphGen~\citep{goyal2020graphgen} & 0.0159 & 0.1677 & 0.3789 & 0.0181 & 58.1 & OOM & OOM & OOM & OOM & OOM \\
        BiGG~\citep{dai2020bigg} & 0.0070 & 0.1150 & 0.4696 & 0.0067 & 50.1 & 0.0994 & 0.6035 & 0.3633 & 0.1589 & 38.2 \\ 
        DiGress~\citep{vignac2023digress} &  0.0041 & 0.0489 & 0.1286 & 0.0018 & 16.2 & OOM & OOM & OOM & OOM & OOM \\ 
        GruM~\citep{jo2024grum} & 0.0019 & 0.0660 & 0.0345 & 0.0030 & 8.2 & OOM & OOM & OOM & OOM & OOM \\
        GEEL~\citep{jang2024simple} & 0.2110 & 0.3753 & 0.1768 & 0.1689 & 287.9 & -- & -- & -- & -- & -- \\
        ESGG~\citep{bergmeister2024efficient} & 0.0030 & 0.0309 & 0.0047 & 0.0013 & 4.7 & 0.0139 & 0.5775 & 0.0780 & 0.0055 & 6.8 \\ \midrule
        \method{} & 0.0004 & 0.0244 & 0.0056 & 0.0013 & \textbf{2.3} & 0.0307 & 0.3031 & 0.0167 & 0.0171 & \textbf{3.0} \\
        \bottomrule
    \end{tabular}
    }
    \end{sc}
    \end{small}
\end{table*}

\begin{wraptable}{r}{.5\textwidth}
    \centering
    \caption{Time comparison of {\scriptsize \method{}} to representative models. OOT indicates the model never reaches the target VUN.}
    \label{tab:time_comparison}
    \begin{small}
    \begin{sc}
    \resizebox{.5\textwidth}{!}{
    \begin{tabular}{llcccc}\toprule
        Dataset & Time& DiGress & Gran & ESGG & \method{} \\ \midrule
        \multirow{2}{*}{Planar} & Training & 25.9h & OOT & 7.4h & \textbf{6.2h (4.2$\times$)} \\
        & Inference & 2.84s & 0.03s & 4.60s & \textbf{0.01s (284$\times$)} \\ \midrule
        \multirow{2}{*}{SBM} & Training & 47.7h & OOT & OOT & \textbf{13.8h (3.5$\times$)} \\
        & Inference & 13.05s & \textbf{0.13s} & 30.0s & \textbf{0.14s (93$\times$)} \\
        \bottomrule
    \end{tabular}
    }
    \end{sc}
    \end{small}
\end{wraptable}
\textbf{Large graph generation.}
To understand the scalability of \method{}, we evaluate its performance on the Proteins and Point Clouds datasets used by~\citet{liao2019efficient}. The results, shown in Table~\ref{tab:protein_and_pointcloud}, demonstrate that even when using a context window shorter than the longest sequence during training, \method{} achieves MMD ratios comparable to those observed on the Planar and SBM datasets. Furthermore, \method{} outperforms all existing methods in terms of MMD ratio, achieving a twofold or more improvement over the previous best model, ESGG. More significantly, while ESGG was specifically designed for generating unattributed graphs, \method{} demonstrates versatility by being applicable to both unattributed and attributed graphs. Finally, previous state-of-the-art autoregressive models, such as GEEL, again fail to achieve competitive performance on these datasets.

\textbf{Molecular graph generation.}
We demonstrate the applicability of our method to generating real-world attributed graphs, such as molecular structures. We evaluate \method{} on the same datasets used by DiGress~\citep{vignac2023digress}, including QM9 (all atoms)~\citep{wu2018moleculenet}, MOSES~\citep{polykovskiy2020moses}, and GuacaMol~\citep{brown2019guacamol}. Following the data splits and experimental setup from DiGress, we benchmark \method{} against a variety of SOTA models, including DiGress, VAE on SMILES~\citep{polykovskiy2020moses}, JT-VAE~\citep{jin2018junction}, GraphINVENT~\citep{mercado2021graphinvent}, NAGVAE~\citep{kwon2020nagvae}, LSTM and MCTS~\citep{brown2019guacamol}. On the QM9 dataset (Table~\ref{tab:qm9}), \method{} outperforms DiGress across all metrics except uniqueness, showing its superiority for attributed graphs.

For the more challenging MOSES and GuacaMol datasets, \method{} also demonstrates superior performance, achieving higher validity and improved distributional alignment as measured by metrics like FCD, as shown in Table~\ref{tab:molecules}. %
Notably, to our best knowledge, \method{} is the first autoregressive model for graphs to surpass diffusion-based approaches on these datasets. It is worth mentioning that all metrics were computed using SMILES representations rather than molecular graphs. Due to the non-reversible nature of converting SMILES to graphs and back, where approximately 20\% of molecules cannot be mapped back to their original SMILES~\citep{vignac2023digress}, some discrepancies are introduced when calculating these metrics. Despite these challenges, \method{} achieves validity and FCD scores comparable to SMILES-based methods.

Furthermore, \method{} demonstrates remarkable efficiency, with training times of less than one day on both datasets, compared to up to one week for DiGress~\citep{vignac2023digress}. This substantial reduction underscores \method{}'s practical advantages in large-scale molecular graph generation tasks.

\begin{table}[tbp]
    \centering
    \vskip -0.1in
    \caption{Benchmarking \method{} on the molecular generation datasets, more results in App.~\ref{app:sec:additional_results_molecules}. \method{}$^*$ was first pretrained on the PubChem-10M dataset~\citep{chithrananda2020chemberta}.}\label{tab:molecules}
    \begin{sc}
    \resizebox{.525\textwidth}{!}{
    \begin{tabular}{lcccccc}\toprule
         & \multicolumn{6}{c}{MOSES} \\ %
         & \multicolumn{6}{c}{$n_{\mathrm{graphs}}= 1.58$M, $|V|_{\max}=27$, $|V|_{\avg}\approx 22$} \\ \cmidrule{2-7}
        Model - Type & Valid$\shortuparrow$ & Unique$\shortuparrow$ & Novel$\shortuparrow$ & Filters$\shortuparrow$ & FCD$\shortdownarrow$ & SNN$\shortdownarrow$ \\ \midrule %
        VAE - SMILES & 97.7 & 99.8 & 69.5 & 99.7 & 0.57 & 0.58 \\ %
        JT-VAE - Fragments & 100 & 100 & 99.9 & 97.8 & 1.00 & 0.53 \\ %
        GraphINVENT - Graph & 96.4 & 99.8 & – & 95.0  & 1.22 & 0.54 \\ %
        DiGress - Graph & 85.7 & 100 & 95.0 & 97.1 & 1.19 & 0.52 \\ \midrule %
        \method{} - Graph & 87.4 & 100 & 85.9 & 98.6 & 0.91 & 0.55 \\  %
        \bottomrule
    \end{tabular}
    }
    \resizebox{.45\textwidth}{!}{
    \begin{tabular}{lccccc}\toprule
         & \multicolumn{5}{c}{GuacaMol} \\ %
         & \multicolumn{5}{c}{$n_{\mathrm{graphs}}=1.1$M, $|V|_{\max}=88$, $|V|_{\avg}\approx 28$} \\ \cmidrule{2-6}
        Model - Type & Valid$\shortuparrow$ & Unique$\shortuparrow$ & Novel$\shortuparrow$ & KL div$\shortuparrow$ & FCD$\shortuparrow$ \\ \midrule
        LSTM - SMILES & 95.9 & 100 & 91.2 & 99.1 & 91.3 \\
        NAGVAE - Graph & 92.7 & 95.5 & 100 & 38.4 & 0.9 \\
        MCTS - Graph & 100 & 100 & 99.4 & 52.2 & 1.5 \\
        DiGress - Graph & 85.2 & 100 & 99.9 & 92.9 & 68.0 \\ \midrule
        \method{} - Graph & 91.6 & 100 & 97.7 & 97.5 & 79.2  \\
        \method{}$^*$ - Graph & 95.9 & 100 & 95.5 & 98.1 & 91.4 \\
        \bottomrule
    \end{tabular}
    }
    \end{sc}
    \vskip -0.1in
\end{table}

\subsection{Transfer Performance of \method{}}
We evaluate the transferability of \method{} by pre-training it on a large dataset of synthetic graphs generated using NetworkX~\citep{hagberg2008networkx} and fine-tuning it on the unattributed graph datasets. Dataset and experimental details are provided in Appendix~\ref{app:sec:experimental_details}. As shown in Table~\ref{tab:transferability}, the pre-trained model consistently outperforms the baseline on small synthetic datasets in terms of the VUN score, achieving near-perfect validity. On larger graph datasets, the pre-trained model also surpasses the baseline across MMD metrics, demonstrating its ability to generalize to more complex structures. However, on small synthetic datasets, the pre-trained model shows a slight decline in MMD metrics compared to the baseline. These findings highlight the potential of building foundation models for graph generation and underscore the need for more comprehensive benchmarks beyond synthetic datasets.

We also test the transferability of \method{} on molecular graphs, by pre-training it on the PubChem-10M dataset~\citep{chithrananda2020chemberta} and fine-tuning on GuacaMol. The pre-trained model substantially outperforms the baseline, as shown in Table~\ref{tab:molecules}.

\begin{table}[tbp]
    \centering
    \caption{Transfer performance on downstream tasks using \method{} pre-trained on the NetworkX dataset. Red and green colors indicate relative decreases and increases, respectively, compared to \method{} without pre-training.}
    \label{tab:transferability}
    \begin{small}
    \begin{sc}
    \resizebox{.6\textwidth}{!}{
    \begin{tabular}{lcccccc}\toprule
        Dataset & Deg. & Clus. & Orbit & Spec. & Ratio & VUN (improvement) \\ \midrule
        NetworkX  & 0.0016 & 0.0073 & 0.0068 & 0.0020 & -- & -- \\ \midrule
        Planar & \cellcolor{red!75} 0.0007 & \cellcolor{red!34} 0.0811 & \cellcolor{red!66} 0.0005 & \cellcolor{green!5} 0.0061 & \cellcolor{red!47} 2.2 & \cellcolor{green!9} 95.0 (+7.5) \\
        SBM & \cellcolor{red!29} 0.0099 & \cellcolor{red!9} 0.0566 & \cellcolor{red!95} 0.0854 & \cellcolor{red!62.5} 0.0065 & \cellcolor{red!41} 4.8 & \cellcolor{green!5} 97.5 (+5) \\
        Proteins & \cellcolor{green!100} 0.0002 & \cellcolor{green!33} 0.0183 & \cellcolor{green!47} 0.0038 & \cellcolor{green!9} 0.0012 & \cellcolor{green!35} 1.7 & -- \\
        Point Clouds & \cellcolor{green!100} 0.0154 & \cellcolor{green!17} 0.2591 & \cellcolor{green!100} 0.0076 & \cellcolor{red!38} 0.0236 & \cellcolor{green!7} 2.8 & --  \\ \bottomrule
    \end{tabular}
    }
    \end{sc}
    \end{small}
\end{table}

\subsection{Substructure Conditioned Generation}\label{sec:substructure_conditioned_generation}
\begin{wraptable}{r}{.41\textwidth}
    \vskip-0.15in
    \caption{Motif scaffolding}
    \label{tab:substructure_conditioned_generation}
    \begin{sc}
    \resizebox{.4\textwidth}{!}{
    \begin{tabular}{lccc}\toprule
        \# Motif copies & Valid & Unique & Novelty \\ \midrule
        1 & 92.0 & 98.8 & 99.6 \\
        2 & 88.8 & 99.7 & 100.0\\
        5 & 66.0 & 100.0 & 100.0 \\ \bottomrule
    \end{tabular}
    }
    \end{sc}
\end{wraptable}
We explore the ability of \method{} to perform substructure-conditioned generation without requiring fine-tuning. Given a subgraph $S$ (which could represent a functional motif of interest in drug discovery), we flatten the subgraph into a SENT sequence and condition the generation process on this sequence. This approach guarantees that the generated graph will contain $S$ as an induced subgraph (Thm.~\ref{thm:induced_subgraph}). As a proof-of-concept, we follow the methodology of~\citet{vignac2023digress,maziarz2022learning} and generate molecular graphs starting from a specific motif, called 1,4-Dihydroquinoline\footnote{\url{https://pubchem.ncbi.nlm.nih.gov/compound/1_4-Dihydroquinoline}}, using the model pre-trained on the GuacaMol dataset. Our results in Table~\ref{tab:substructure_conditioned_generation} demonstrate that this approach maintains similar validity, uniqueness, and novelty to unconditional generation (Table~\ref{tab:molecules}). To further showcase the flexibility of this method, we test more extreme cases by replicating the same motif multiple times before performing the conditional generation. While validity decreases significantly when using an unrealistically large number of copies (\eg 5), the model still generates some visually plausible molecules (Appendix~\ref{app:sec:substructure_conditioned_generation}), showing superior flexibility over \citet{vignac2023digress}. These results highlight the potential of \method{} for important applications in drug discovery, particularly in motif scaffolding. Additional experiments on multiple but different motifs are provided in App.~\ref{app:sec:substructure_conditioned_generation}.

\subsection{Ablation Experiments}\label{sec:ablation_experiments}

In this study, we aim to understand the effectiveness of the key components in \method{}.

\textbf{Comparison of sequence model architectures.}
\method{} provides a novel framework for evaluating the capability of current LLM architectures in graph generation and, more broadly, in structural reasoning tasks. In Table~\ref{tab:architecture}, we compare several state-of-the-art architectures on the Planar dataset, including GPT-2~\citep{radford2019gpt2}, Mamba~\citep{gu2023mamba}, and LLaMA~\citep{touvron2023llama}. While all models achieve comparable MMD ratios, transformer-based architectures, particularly LLaMA, demonstrate significantly better performance in terms of VUN scores compared to state-space models. These findings highlight the potential of \method{} to serve as a valuable benchmark for assessing sequence/language models' capabilities in graph generation tasks.

\begin{table}[tbp]
    \centering
    \vskip -0.1in
    \caption{Comparison of sequence model architectures on the Planar dataset.}
    \label{tab:architecture}
    \begin{small}
    \begin{sc}
    \resizebox{.6\textwidth}{!}{
    \begin{tabular}{lcccccc}\toprule
        Architecture & Deg. & Clus. & Orbit & Spec. & Ratio & VUN  \\ \midrule
        GPT-2 & 0.0004 & 0.0720 & 0.0010 & 0.0053 & 1.8 & 85.0 \\
        Mamba & 0.0002 &  0.0429 & 0.0014 & 0.0087 & 1.6 & 55.0 \\
        LLaMA & 0.0005 & 0.0651 & 0.0005 & 0.0056 & \textbf{1.6} & \textbf{90.0} \\ \bottomrule
    \end{tabular}
    }
    \end{sc}
    \end{small}
    \vskip -0.1in
\end{table}

\begin{figure}[tbp]
    \centering
    \includegraphics[width=0.24\linewidth]{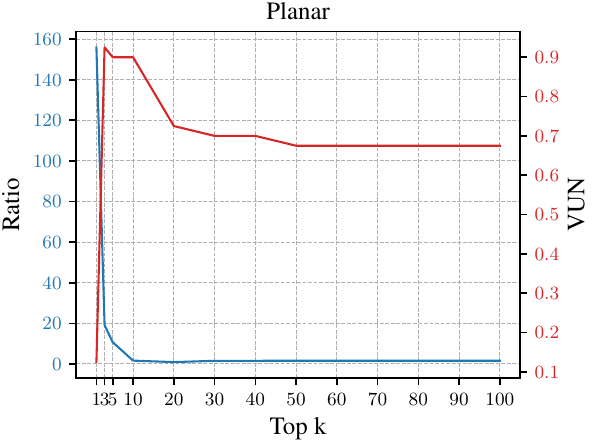}
    \includegraphics[width=0.24\linewidth]{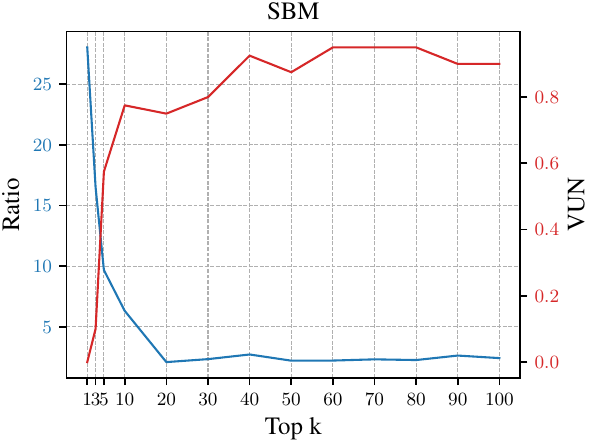}
    \includegraphics[width=.5\linewidth]{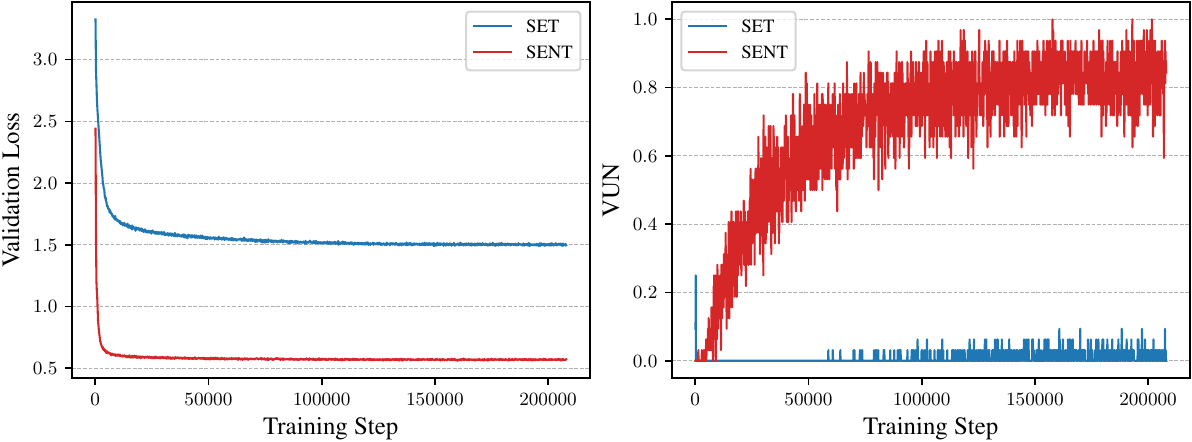}
    \caption{Ablation experiments. Left: the effect of top-k sampling on the Planar and SBM datasets. Right: the validation loss and VUN scores when using SET and SENT on the Planar dataset.}
    \label{fig:ablation_exp}
\end{figure}

\textbf{Effect of top-k sampling.}
A key advantage of \method{} over diffusion-based approaches is the flexibility to apply top-k sampling~\citep{fan2018topk} during inference, which can improve generation quality. As shown on the left of Figure~\ref{fig:ablation_exp}, a smaller $k$ improves the VUN score on the Planar dataset, whereas it is not beneficial on the SBM dataset. In contrast, increasing $k$ generally improves MMD ratios on both datasets. These results suggest that top-k sampling can be optimized based on dataset characteristics. In our experiments, we select the best $k$ that maximizes the VUN score for small synthetic datasets and minimizes the validation MMD ratios for other datasets. This flexibility allows practitioners to select $k$ based on specific performance criteria they aim to prioritize.

\textbf{Comparison of SET and SENT.}
As discussed in Section~\ref{sec:methods}, SENT is preferred over SET for graph generation, as incorporating neighborhood information is essential to ensure structural coherence. To empirically validate this, we compare the performance of SENT and SET on the Planar dataset and present the training curves on the right of Figure~\ref{fig:ablation_exp}. Consistent with our theoretical analysis, SET fails to produce high-validity graphs, resulting in a VUN score close to zero, whereas SENT successfully generates valid planar graphs.

\section{Conclusion}\label{sec:conclusion}
We proposed \method{}, a scalable and efficient autoregressive model for attributed graph generation that handles large graphs while maintaining high quality. Our approach enables substructure-conditioned generation without additional fine-tuning and demonstrates promising transfer capabilities. Crucially, \method{} establishes the first fundamental connection between graph and language modeling--where graphs are losslessly represented as token sequences, and prefixes in these sequences serve as meaningful patterns in both paradigms--representing a significant step toward applying language modeling techniques to graph generation and broader graph learning challenges.

\textbf{Limitations.}
While \method{} demonstrates strong scalability on current graph generation benchmarks, we acknowledge that the datasets used in our study remain relatively small-scale compared to those used in pre-training LLMs. To push the boundaries of more powerful graph generative models or eventually foundation models for graphs, we draw the community's attention to building more comprehensive graph generation benchmarks and well-curated pre-training datasets.

\section*{Acknowledgements}
The authors thank Dr.\ Till Hendrik Schulz, Philip Hartout, and Błażej Banaszewski for their insightful discussions and valuable feedback on the manuscript. The authors thank Dr.\ Till Hendrik Schulz for presenting the poster at NeurIPS, as the first author was unable to attend due to a visa issue.

\bibliographystyle{plainnat}
\bibliography{mybib}

\begin{thebibliography}{74}
\providecommand{\natexlab}[1]{#1}
\providecommand{\url}[1]{\texttt{#1}}
\expandafter\ifx\csname urlstyle\endcsname\relax
  \providecommand{\doi}[1]{doi: #1}\else
  \providecommand{\doi}{doi: \begingroup \urlstyle{rm}\Url}\fi

\bibitem[Bacciu et~al.(2020)Bacciu, Micheli, and Podda]{bacciu2020edge}
Davide Bacciu, Alessio Micheli, and Marco Podda.
\newblock Edge-based sequential graph generation with recurrent neural networks.
\newblock \emph{Neurocomputing}, 416:\penalty0 177--189, 2020.

\bibitem[Bergmeister et~al.(2024)Bergmeister, Martinkus, Perraudin, and Wattenhofer]{bergmeister2024efficient}
Andreas Bergmeister, Karolis Martinkus, Nathana{\"e}l Perraudin, and Roger Wattenhofer.
\newblock Efficient and scalable graph generation through iterative local expansion.
\newblock In \emph{International Conference on Learning Representations (ICLR)}, 2024.

\bibitem[Biggs et~al.(1986)Biggs, Lloyd, and Wilson]{biggs1986graph}
Norman Biggs, E~Keith Lloyd, and Robin~J Wilson.
\newblock \emph{Graph Theory, 1736-1936}.
\newblock Oxford University Press, 1986.

\bibitem[Bojchevski et~al.(2018)Bojchevski, Shchur, Z{\"u}gner, and G{\"u}nnemann]{bojchevski2018netgan}
Aleksandar Bojchevski, Oleksandr Shchur, Daniel Z{\"u}gner, and Stephan G{\"u}nnemann.
\newblock Netgan: Generating graphs via random walks.
\newblock In \emph{International Conference on Machine Learning (ICML)}, pages 610--619, 2018.

\bibitem[Brockschmidt et~al.(2019)Brockschmidt, Allamanis, Gaunt, and Polozov]{brockschmidt2019generative}
Marc Brockschmidt, Miltiadis Allamanis, Alexander~L Gaunt, and Oleksandr Polozov.
\newblock Generative code modeling with graphs.
\newblock In \emph{International Conference on Learning Representations (ICLR)}, 2019.

\bibitem[Brown et~al.(2019)Brown, Fiscato, Segler, and Vaucher]{brown2019guacamol}
Nathan Brown, Marco Fiscato, Marwin~HS Segler, and Alain~C Vaucher.
\newblock Guacamol: benchmarking models for de novo molecular design.
\newblock \emph{Journal of chemical information and modeling}, 59\penalty0 (3):\penalty0 1096--1108, 2019.

\bibitem[Brown et~al.(2020)Brown, Mann, Ryder, Subbiah, Kaplan, Dhariwal, Neelakantan, Shyam, Sastry, Askell, Agarwal, Herbert-Voss, Krueger, Henighan, Child, Ramesh, Ziegler, Wu, Winter, Hesse, Chen, Sigler, Litwin, Gray, Chess, Clark, Berner, McCandlish, Radford, Sutskever, and Amodei]{gpt3}
Tom Brown, Benjamin Mann, Nick Ryder, Melanie Subbiah, Jared~D Kaplan, Prafulla Dhariwal, Arvind Neelakantan, Pranav Shyam, Girish Sastry, Amanda Askell, Sandhini Agarwal, Ariel Herbert-Voss, Gretchen Krueger, Tom Henighan, Rewon Child, Aditya Ramesh, Daniel Ziegler, Jeffrey Wu, Clemens Winter, Chris Hesse, Mark Chen, Eric Sigler, Mateusz Litwin, Scott Gray, Benjamin Chess, Jack Clark, Christopher Berner, Sam McCandlish, Alec Radford, Ilya Sutskever, and Dario Amodei.
\newblock Language models are few-shot learners.
\newblock In \emph{Advances in Neural Information Processing Systems (NeurIPS)}, volume~33, pages 1877--1901, 2020.

\bibitem[Chen et~al.(2020)Chen, Jacob, and Mairal]{chen2020convolutional}
Dexiong Chen, Laurent Jacob, and Julien Mairal.
\newblock Convolutional kernel networks for graph-structured data.
\newblock In \emph{International Conference on Machine Learning (ICML)}, pages 1576--1586, 2020.

\bibitem[Chen et~al.(2022)Chen, O’Bray, and Borgwardt]{chen2022structure}
Dexiong Chen, Leslie O’Bray, and Karsten Borgwardt.
\newblock Structure-aware transformer for graph representation learning.
\newblock In \emph{International Conference on Machine Learning (ICML)}, pages 3469--3489. PMLR, 2022.

\bibitem[Chen et~al.(2024{\natexlab{a}})Chen, Schulz, and Borgwardt]{chen2024neuralwalker}
Dexiong Chen, Till~Hendrik Schulz, and Karsten Borgwardt.
\newblock Learning long range dependencies on graphs via random walks.
\newblock \emph{arXiv preprint arXiv:2406.03386}, 2024{\natexlab{a}}.

\bibitem[Chen et~al.(2024{\natexlab{b}})Chen, Zhao, Jaiswal, Shah, and Wang]{chen2024llaga}
Runjin Chen, Tong Zhao, Ajay~Kumar Jaiswal, Neil Shah, and Zhangyang Wang.
\newblock Llaga: Large language and graph assistant.
\newblock In \emph{International Conference on Machine Learning (ICML)}, pages 7809--7823. PMLR, 2024{\natexlab{b}}.

\bibitem[Chen et~al.(2021)Chen, Han, Hu, Ruiz, and Liu]{chen2021order}
Xiaohui Chen, Xu~Han, Jiajing Hu, Francisco Ruiz, and Liping Liu.
\newblock Order matters: Probabilistic modeling of node sequence for graph generation.
\newblock In \emph{International Conference on Machine Learning (ICML)}, pages 1630--1639, 2021.

\bibitem[Chen et~al.(2023)Chen, He, Han, and Liu]{chen2023efficient}
Xiaohui Chen, Jiaxing He, Xu~Han, and Liping Liu.
\newblock Efficient and degree-guided graph generation via discrete diffusion modeling.
\newblock In \emph{International Conference on Machine Learning (ICML)}, pages 4585--4610, 2023.

\bibitem[Chithrananda et~al.(2020)Chithrananda, Grand, and Ramsundar]{chithrananda2020chemberta}
Seyone Chithrananda, Gabriel Grand, and Bharath Ramsundar.
\newblock Chemberta: large-scale self-supervised pretraining for molecular property prediction.
\newblock \emph{arXiv preprint arXiv:2010.09885}, 2020.

\bibitem[Chung et~al.(2014)Chung, Gulcehre, Cho, and Bengio]{chung2014empirical}
Junyoung Chung, Caglar Gulcehre, Kyunghyun Cho, and Yoshua Bengio.
\newblock Empirical evaluation of gated recurrent neural networks on sequence modeling.
\newblock In \emph{NIPS Workshop on Deep Learning}, 2014.

\bibitem[Dai et~al.(2020)Dai, Nazi, Li, Dai, and Schuurmans]{dai2020bigg}
Hanjun Dai, Azade Nazi, Yujia Li, Bo~Dai, and Dale Schuurmans.
\newblock Scalable deep generative modeling for sparse graphs.
\newblock In \emph{International Conference on Machine Learning (ICML)}, pages 2302--2312, 2020.

\bibitem[Dhariwal et~al.(2020)Dhariwal, Jun, Payne, Kim, Radford, and Sutskever]{dhariwal2020jukebox}
Prafulla Dhariwal, Heewoo Jun, Christine Payne, Jong~Wook Kim, Alec Radford, and Ilya Sutskever.
\newblock Jukebox: A generative model for music.
\newblock \emph{arXiv preprint arXiv:2005.00341}, 2020.

\bibitem[Diamant et~al.(2023)Diamant, Tseng, Chuang, Biancalani, and Scalia]{diamant2023improving}
Nathaniel~Lee Diamant, Alex~M Tseng, Kangway~V Chuang, Tommaso Biancalani, and Gabriele Scalia.
\newblock Improving graph generation by restricting graph bandwidth.
\newblock In \emph{International Conference on Machine Learning (ICML)}. PMLR, 2023.

\bibitem[Diestel(2005)]{diestel2005graph}
Reinhard Diestel.
\newblock \emph{Graph Theory}.
\newblock Electronic library of mathematics. Springer, 2005.

\bibitem[Dobson and Doig(2003)]{dobson2003distinguishing}
Paul~D Dobson and Andrew~J Doig.
\newblock Distinguishing enzyme structures from non-enzymes without alignments.
\newblock \emph{Journal of molecular biology}, 330\penalty0 (4):\penalty0 771--783, 2003.

\bibitem[Fan et~al.(2018)Fan, Lewis, and Dauphin]{fan2018topk}
Angela Fan, Mike Lewis, and Yann Dauphin.
\newblock Hierarchical neural story generation.
\newblock In \emph{Proceedings of the Annual Meeting of the Association for Computational Linguistics (ACL)}, pages 889--898, 2018.

\bibitem[Fatemi et~al.(2024)Fatemi, Halcrow, and Perozzi]{fatemi2024talk}
Bahare Fatemi, Jonathan Halcrow, and Bryan Perozzi.
\newblock Talk like a graph: Encoding graphs for large language models.
\newblock In \emph{International Conference on Learning Representations (ICLR)}, 2024.

\bibitem[Goyal et~al.(2020)Goyal, Jain, and Ranu]{goyal2020graphgen}
Nikhil Goyal, Harsh~Vardhan Jain, and Sayan Ranu.
\newblock Graphgen: A scalable approach to domain-agnostic labeled graph generation.
\newblock In \emph{Proceedings of The Web Conference}, pages 1253--1263, 2020.

\bibitem[Gretton et~al.(2012)Gretton, Borgwardt, Rasch, Sch{\"o}lkopf, and Smola]{gretton2012kernel}
Arthur Gretton, Karsten~M Borgwardt, Malte~J Rasch, Bernhard Sch{\"o}lkopf, and Alexander Smola.
\newblock A kernel two-sample test.
\newblock \emph{Journal of Machine Learning Research (JMLR)}, 13\penalty0 (1):\penalty0 723--773, 2012.

\bibitem[Gu and Dao(2023)]{gu2023mamba}
Albert Gu and Tri Dao.
\newblock Mamba: Linear-time sequence modeling with selective state spaces.
\newblock \emph{arXiv preprint arXiv:2312.00752}, 2023.

\bibitem[Hagberg et~al.(2008)Hagberg, Swart, and Schult]{hagberg2008networkx}
Aric Hagberg, Pieter~J Swart, and Daniel~A Schult.
\newblock Exploring network structure, dynamics, and function using networkx.
\newblock Technical report, Los Alamos National Laboratory (LANL), Los Alamos, NM (United States), 2008.

\bibitem[Hoogeboom et~al.(2022)Hoogeboom, Satorras, Vignac, and Welling]{hoogeboom2022equivariant}
Emiel Hoogeboom, V{\i}ctor~Garcia Satorras, Cl{\'e}ment Vignac, and Max Welling.
\newblock Equivariant diffusion for molecule generation in 3d.
\newblock In \emph{International Conference on Machine Learning (ICML)}, pages 8867--8887, 2022.

\bibitem[Huang et~al.(2024)Huang, Li, Yang, Shi, Chang, Ye, Wu, Hong, Huang, Liu, et~al.]{huang2024audiogpt}
Rongjie Huang, Mingze Li, Dongchao Yang, Jiatong Shi, Xuankai Chang, Zhenhui Ye, Yuning Wu, Zhiqing Hong, Jiawei Huang, Jinglin Liu, et~al.
\newblock Audiogpt: Understanding and generating speech, music, sound, and talking head.
\newblock In \emph{Proceedings of the AAAI Conference on Artificial Intelligence}, pages 23802--23804, 2024.

\bibitem[Ingraham et~al.(2019)Ingraham, Garg, Barzilay, and Jaakkola]{ingraham2019generative}
John Ingraham, Vikas Garg, Regina Barzilay, and Tommi Jaakkola.
\newblock Generative models for graph-based protein design.
\newblock In \emph{Advances in Neural Information Processing Systems (NeurIPS)}, volume~32, 2019.

\bibitem[Ivanov and Burnaev(2018)]{ivanov2018anonymous}
Sergey Ivanov and Evgeny Burnaev.
\newblock Anonymous walk embeddings.
\newblock In \emph{International Conference on Machine Learning (ICML)}, 2018.

\bibitem[Jain(2022)]{jain2022huggingface}
Shashank~Mohan Jain.
\newblock Hugging face.
\newblock In \emph{Introduction to transformers for NLP: With the hugging face library and models to solve problems}, pages 51--67. Springer, 2022.

\bibitem[Jang et~al.(2024)Jang, Lee, and Ahn]{jang2024simple}
Yunhui Jang, Seul Lee, and Sungsoo Ahn.
\newblock A simple and scalable representation for graph generation.
\newblock In \emph{International Conference on Learning Representations (ICLR)}, 2024.

\bibitem[Jin et~al.(2018)Jin, Barzilay, and Jaakkola]{jin2018junction}
Wengong Jin, Regina Barzilay, and Tommi Jaakkola.
\newblock Junction tree variational autoencoder for molecular graph generation.
\newblock In \emph{International Conference on Machine Learning (ICML)}, 2018.

\bibitem[Jo et~al.(2022)Jo, Lee, and Hwang]{jo2022score}
Jaehyeong Jo, Seul Lee, and Sung~Ju Hwang.
\newblock Score-based generative modeling of graphs via the system of stochastic differential equations.
\newblock In \emph{International Conference on Machine Learning (ICML)}, pages 10362--10383, 2022.

\bibitem[Jo et~al.(2024)Jo, Kim, and Hwang]{jo2024grum}
Jaehyeong Jo, Dongki Kim, and Sung~Ju Hwang.
\newblock Graph generation with diffusion mixture.
\newblock In \emph{International Conference on Machine Learning (ICML)}. PMLR, 2024.

\bibitem[Karami(2024)]{karami2024higen}
Mahdi Karami.
\newblock Higen: Hierarchical graph generative networks.
\newblock In \emph{International Conference on Learning Representations (ICLR)}, 2024.

\bibitem[Kim et~al.(2024)Kim, Zaghen, Suleymanzade, Ryou, and Hong]{kim2024revisiting}
Jinwoo Kim, Olga Zaghen, Ayhan Suleymanzade, Youngmin Ryou, and Seunghoon Hong.
\newblock Revisiting random walks for learning on graphs.
\newblock \emph{arXiv preprint arXiv:2407.01214}, 2024.

\bibitem[Kipf and Welling(2016)]{kipf2016variational}
Thomas~N Kipf and Max Welling.
\newblock Variational graph auto-encoders.
\newblock \emph{arXiv preprint arXiv:1611.07308}, 2016.

\bibitem[Kong et~al.(2023)Kong, Cui, Sun, Zhuang, Prakash, and Zhang]{kong2023autoregressive}
Lingkai Kong, Jiaming Cui, Haotian Sun, Yuchen Zhuang, B~Aditya Prakash, and Chao Zhang.
\newblock Autoregressive diffusion model for graph generation.
\newblock In \emph{International Conference on Machine Learning (ICML)}, pages 17391--17408, 2023.

\bibitem[Krimmel et~al.(2025)Krimmel, Wiens, Borgwardt, and Chen]{krimmel2025towards}
Markus Krimmel, Jenna Wiens, Karsten Borgwardt, and Dexiong Chen.
\newblock Towards fast graph generation via autoregressive noisy filtration modeling.
\newblock \emph{arXiv preprint arXiv:2502.02415}, 2025.

\bibitem[Kwon et~al.(2020)Kwon, Lee, Choi, Shin, and Kang]{kwon2020nagvae}
Youngchun Kwon, Dongseon Lee, Youn-Suk Choi, Kyoham Shin, and Seokho Kang.
\newblock Compressed graph representation for scalable molecular graph generation.
\newblock \emph{Journal of Cheminformatics}, 12:\penalty0 1--8, 2020.

\bibitem[Li et~al.(2018)Li, Vinyals, Dyer, Pascanu, and Battaglia]{li2018learning}
Yujia Li, Oriol Vinyals, Chris Dyer, Razvan Pascanu, and Peter Battaglia.
\newblock Learning deep generative models of graphs.
\newblock \emph{arXiv preprint arXiv:1803.03324}, 2018.

\bibitem[Liao et~al.(2019)Liao, Li, Song, Wang, Hamilton, Duvenaud, Urtasun, and Zemel]{liao2019efficient}
Renjie Liao, Yujia Li, Yang Song, Shenlong Wang, Will Hamilton, David~K Duvenaud, Raquel Urtasun, and Richard Zemel.
\newblock Efficient graph generation with graph recurrent attention networks.
\newblock In \emph{Advances in Neural Information Processing Systems (NeurIPS)}, 2019.

\bibitem[Lim et~al.(2020)Lim, Hwang, Moon, Kim, and Kim]{lim2020scaffold}
Jaechang Lim, Sang-Yeon Hwang, Seokhyun Moon, Seungsu Kim, and Woo~Youn Kim.
\newblock Scaffold-based molecular design with a graph generative model.
\newblock \emph{Chemical Science}, 11\penalty0 (4):\penalty0 1153--1164, 2020.

\bibitem[Martinkus et~al.(2022)Martinkus, Loukas, Perraudin, and Wattenhofer]{martinkus2022spectre}
Karolis Martinkus, Andreas Loukas, Nathana{\"e}l Perraudin, and Roger Wattenhofer.
\newblock Spectre: Spectral conditioning helps to overcome the expressivity limits of one-shot graph generators.
\newblock In \emph{International Conference on Machine Learning (ICML)}, pages 15159--15179, 2022.

\bibitem[Maziarz et~al.(2022)Maziarz, Jackson-Flux, Cameron, Sirockin, Schneider, Stiefl, Segler, and Brockschmidt]{maziarz2022learning}
Krzysztof Maziarz, Henry~Richard Jackson-Flux, Pashmina Cameron, Finton Sirockin, Nadine Schneider, Nikolaus Stiefl, Marwin Segler, and Marc Brockschmidt.
\newblock Learning to extend molecular scaffolds with structural motifs.
\newblock In \emph{International Conference on Learning Representations (ICLR)}, 2022.

\bibitem[Mercado et~al.(2021)Mercado, Rastemo, Lindel{\"o}f, Klambauer, Engkvist, Chen, and Bjerrum]{mercado2021graphinvent}
Roc{\'\i}o Mercado, Tobias Rastemo, Edvard Lindel{\"o}f, G{\"u}nter Klambauer, Ola Engkvist, Hongming Chen, and Esben~Jannik Bjerrum.
\newblock Graph networks for molecular design.
\newblock \emph{Machine Learning: Science and Technology}, 2\penalty0 (2):\penalty0 025023, 2021.

\bibitem[Mialon et~al.(2021)Mialon, Chen, Selosse, and Mairal]{mialon2021graphit}
Gr{\'e}goire Mialon, Dexiong Chen, Margot Selosse, and Julien Mairal.
\newblock Graphit: Encoding graph structure in transformers.
\newblock \emph{arXiv preprint arXiv:2106.05667}, 2021.

\bibitem[Neumann et~al.(2013)Neumann, Moreno, Antanas, Garnett, and Kersting]{neumann2013graph}
Marion Neumann, Plinio Moreno, Laura Antanas, Roman Garnett, and Kristian Kersting.
\newblock Graph kernels for object category prediction in task-dependent robot grasping.
\newblock In \emph{Online proceedings of the eleventh workshop on mining and learning with graphs}, pages 0--6. ACM Chicago, Illinois, USA, 2013.

\bibitem[Nikolentzos and Vazirgiannis(2020)]{nikolentzos2020random}
Giannis Nikolentzos and Michalis Vazirgiannis.
\newblock Random walk graph neural networks.
\newblock In \emph{Advances in Neural Information Processing Systems (NeurIPS)}, volume~33, pages 16211--16222, 2020.

\bibitem[Niu et~al.(2020)Niu, Song, Song, Zhao, Grover, and Ermon]{niu2020permutation}
Chenhao Niu, Yang Song, Jiaming Song, Shengjia Zhao, Aditya Grover, and Stefano Ermon.
\newblock Permutation invariant graph generation via score-based generative modeling.
\newblock In \emph{International Conference on Artificial Intelligence and Statistics (AISTATS)}, pages 4474--4484, 2020.

\bibitem[O'Bray et~al.(2022)O'Bray, Horn, Rieck, and Borgwardt]{obray2022evaluation}
Leslie O'Bray, Max Horn, Bastian Rieck, and Karsten~M. Borgwardt.
\newblock Evaluation metrics for graph generative models: Problems, pitfalls, and practical solutions.
\newblock In \emph{International Conference on Learning Representations (ICLR)}, 2022.

\bibitem[Polykovskiy et~al.(2020)Polykovskiy, Zhebrak, Sanchez-Lengeling, Golovanov, Tatanov, Belyaev, Kurbanov, Artamonov, Aladinskiy, Veselov, et~al.]{polykovskiy2020moses}
Daniil Polykovskiy, Alexander Zhebrak, Benjamin Sanchez-Lengeling, Sergey Golovanov, Oktai Tatanov, Stanislav Belyaev, Rauf Kurbanov, Aleksey Artamonov, Vladimir Aladinskiy, Mark Veselov, et~al.
\newblock Molecular sets (moses): a benchmarking platform for molecular generation models.
\newblock \emph{Frontiers in pharmacology}, 11:\penalty0 565644, 2020.

\bibitem[Radford et~al.(2019)Radford, Wu, Child, Luan, Amodei, Sutskever, et~al.]{radford2019gpt2}
Alec Radford, Jeffrey Wu, Rewon Child, David Luan, Dario Amodei, Ilya Sutskever, et~al.
\newblock Language models are unsupervised multitask learners.
\newblock \emph{OpenAI blog}, 1\penalty0 (8):\penalty0 9, 2019.

\bibitem[Rombach et~al.(2022)Rombach, Blattmann, Lorenz, Esser, and Ommer]{rombach2022high}
Robin Rombach, Andreas Blattmann, Dominik Lorenz, Patrick Esser, and Bj{\"o}rn Ommer.
\newblock High-resolution image synthesis with latent diffusion models.
\newblock In \emph{Proceedings of the Conference on Computer Vision and Pattern Recognition (CVPR)}, pages 10684--10695, 2022.

\bibitem[Shazeer(2020)]{shazeer2020glu}
Noam Shazeer.
\newblock Glu variants improve transformer.
\newblock \emph{arXiv preprint arXiv:2002.05202}, 2020.

\bibitem[Shi et~al.(2020)Shi, Xu, Zhu, Zhang, Zhang, and Tang]{shi2020graphaf}
Chence Shi, Minkai Xu, Zhaocheng Zhu, Weinan Zhang, Ming Zhang, and Jian Tang.
\newblock Graphaf: a flow-based autoregressive model for molecular graph generation.
\newblock In \emph{International Conference on Learning Representations (ICLR)}, 2020.

\bibitem[Simonovsky and Komodakis(2018)]{simonovsky2018graphvae}
Martin Simonovsky and Nikos Komodakis.
\newblock Graphvae: Towards generation of small graphs using variational autoencoders.
\newblock In \emph{International Conference on Artificial Neural Networks}, pages 412--422, 2018.

\bibitem[Su et~al.(2024)Su, Ahmed, Lu, Pan, Bo, and Liu]{su2024rope}
Jianlin Su, Murtadha Ahmed, Yu~Lu, Shengfeng Pan, Wen Bo, and Yunfeng Liu.
\newblock Roformer: Enhanced transformer with rotary position embedding.
\newblock \emph{Neurocomputing}, 568:\penalty0 127063, 2024.

\bibitem[Thompson et~al.(2022)Thompson, Knyazev, Ghalebi, Kim, and Taylor]{thompson2022evaluation}
Rylee Thompson, Boris Knyazev, Elahe Ghalebi, Jungtaek Kim, and Graham~W Taylor.
\newblock On evaluation metrics for graph generative models.
\newblock In \emph{International Conference on Learning Representations (ICLR)}, 2022.

\bibitem[Tian et~al.(2024)Tian, Jiang, Yuan, Peng, and Wang]{tian2024var}
Keyu Tian, Yi~Jiang, Zehuan Yuan, Bingyue Peng, and Liwei Wang.
\newblock Visual autoregressive modeling: Scalable image generation via next-scale prediction.
\newblock In \emph{Advances in Neural Information Processing Systems (NeurIPS)}, 2024.

\bibitem[T{\"o}nshoff et~al.(2023)T{\"o}nshoff, Ritzert, Wolf, and Grohe]{tonshoff2023crawl}
Jan T{\"o}nshoff, Martin Ritzert, Hinrikus Wolf, and Martin Grohe.
\newblock Walking out of the weisfeiler leman hierarchy: Graph learning beyond message passing.
\newblock \emph{Transactions on Machine Learning Research (TMLR)}, 2023.

\bibitem[Touvron et~al.(2023{\natexlab{a}})Touvron, Lavril, Izacard, Martinet, Lachaux, Lacroix, Rozi{\`e}re, Goyal, Hambro, Azhar, et~al.]{touvron2023llama}
Hugo Touvron, Thibaut Lavril, Gautier Izacard, Xavier Martinet, Marie-Anne Lachaux, Timoth{\'e}e Lacroix, Baptiste Rozi{\`e}re, Naman Goyal, Eric Hambro, Faisal Azhar, et~al.
\newblock Llama: Open and efficient foundation language models.
\newblock \emph{arXiv preprint arXiv:2302.13971}, 2023{\natexlab{a}}.

\bibitem[Touvron et~al.(2023{\natexlab{b}})Touvron, Martin, Stone, Albert, Almahairi, Babaei, Bashlykov, Batra, Bhargava, Bhosale, et~al.]{touvron2023llama2}
Hugo Touvron, Louis Martin, Kevin Stone, Peter Albert, Amjad Almahairi, Yasmine Babaei, Nikolay Bashlykov, Soumya Batra, Prajjwal Bhargava, Shruti Bhosale, et~al.
\newblock Llama 2: Open foundation and fine-tuned chat models.
\newblock \emph{arXiv preprint arXiv:2307.09288}, 2023{\natexlab{b}}.

\bibitem[Vaswani et~al.(2017)Vaswani, Shazeer, Parmar, Uszkoreit, Jones, Gomez, Kaiser, and Polosukhin]{vaswani2017attention}
Ashish Vaswani, Noam Shazeer, Niki Parmar, Jakob Uszkoreit, Llion Jones, Aidan~N Gomez, {\L}ukasz Kaiser, and Illia Polosukhin.
\newblock Attention is all you need.
\newblock In \emph{Advances in Neural Information Processing Systems (NeurIPS)}, 2017.

\bibitem[Vignac et~al.(2023)Vignac, Krawczuk, Siraudin, Wang, Cevher, and Frossard]{vignac2023digress}
Cl{\'{e}}ment Vignac, Igor Krawczuk, Antoine Siraudin, Bohan Wang, Volkan Cevher, and Pascal Frossard.
\newblock Digress: Discrete denoising diffusion for graph generation.
\newblock In \emph{International Conference on Learning Representations (ICLR)}, 2023.

\bibitem[Wang et~al.(2021)Wang, Chang, Liu, Leskovec, and Li]{wang2021inductive}
Yanbang Wang, Yen-Yu Chang, Yunyu Liu, Jure Leskovec, and Pan Li.
\newblock Inductive representation learning in temporal networks via causal anonymous walks.
\newblock \emph{arXiv preprint arXiv:2101.05974}, 2021.

\bibitem[Wu et~al.(2018)Wu, Ramsundar, Feinberg, Gomes, Geniesse, Pappu, Leswing, and Pande]{wu2018moleculenet}
Zhenqin Wu, Bharath Ramsundar, Evan~N Feinberg, Joseph Gomes, Caleb Geniesse, Aneesh~S Pappu, Karl Leswing, and Vijay Pande.
\newblock Moleculenet: a benchmark for molecular machine learning.
\newblock \emph{Chemical science}, 9\penalty0 (2):\penalty0 513--530, 2018.

\bibitem[Xu et~al.(2024)Xu, Qiu, Chen, Chen, Fan, Pan, Zeng, Das, and Tong]{xu2024discrete}
Zhe Xu, Ruizhong Qiu, Yuzhong Chen, Huiyuan Chen, Xiran Fan, Menghai Pan, Zhichen Zeng, Mahashweta Das, and Hanghang Tong.
\newblock Discrete-state continuous-time diffusion for graph generation.
\newblock In \emph{Advances in Neural Information Processing Systems (NeurIPS)}, 2024.

\bibitem[Yin et~al.(2022)Yin, Zhang, Wang, Wang, and Li]{yin2022algorithm}
Haoteng Yin, Muhan Zhang, Yanbang Wang, Jianguo Wang, and Pan Li.
\newblock Algorithm and system co-design for efficient subgraph-based graph representation learning.
\newblock \emph{Proceedings of the VLDB Endowment}, 15\penalty0 (11):\penalty0 2788--2796, 2022.

\bibitem[You et~al.(2018)You, Ying, Ren, Hamilton, and Leskovec]{you2018graphrnn}
Jiaxuan You, Rex Ying, Xiang Ren, William Hamilton, and Jure Leskovec.
\newblock {GraphRNN}: Generating realistic graphs with deep auto-regressive models.
\newblock In \emph{International Conference on Machine Learning (ICML)}, 2018.

\bibitem[Zhang and Sennrich(2019)]{zhang2019rmsnorm}
Biao Zhang and Rico Sennrich.
\newblock Root mean square layer normalization.
\newblock In \emph{Advances in Neural Information Processing Systems (NeurIPS)}, volume~32, 2019.

\bibitem[Zhao et~al.(2024)Zhao, Ding, and Akoglu]{zhao2024pard}
Lingxiao Zhao, Xueying Ding, and Leman Akoglu.
\newblock Pard: Permutation-invariant autoregressive diffusion for graph generation.
\newblock In \emph{Advances in Neural Information Processing Systems (NeurIPS)}, 2024.

\bibitem[Zhao et~al.(2023)Zhao, Ren, Li, Xu, and Liu]{zhao2023graphgpt}
Qifang Zhao, Weidong Ren, Tianyu Li, Xiaoxiao Xu, and Hong Liu.
\newblock Graphgpt: Graph learning with generative pre-trained transformers.
\newblock \emph{arXiv preprint arXiv:2401.00529}, 2023.

\end{thebibliography}

\clearpage
\newpage
\section*{NeurIPS Paper Checklist}

\begin{enumerate}

\item {\bf Claims}
    \item[] Question: Do the main claims made in the abstract and introduction accurately reflect the paper's contributions and scope?
    \item[] Answer: \answerYes{} %
    \item[] Justification: The claims are fully justified in our theoretical and empirical results in Section~\ref{sec:methods} and \ref{sec:experiments}.
    \item[] Guidelines:
    \begin{itemize}
        \item The answer NA means that the abstract and introduction do not include the claims made in the paper.
        \item The abstract and/or introduction should clearly state the claims made, including the contributions made in the paper and important assumptions and limitations. A No or NA answer to this question will not be perceived well by the reviewers. 
        \item The claims made should match theoretical and experimental results, and reflect how much the results can be expected to generalize to other settings. 
        \item It is fine to include aspirational goals as motivation as long as it is clear that these goals are not attained by the paper. 
    \end{itemize}

\item {\bf Limitations}
    \item[] Question: Does the paper discuss the limitations of the work performed by the authors?
    \item[] Answer: \answerYes{} %
    \item[] Justification: The limitations are discussed in Appendix~\ref{sec:conclusion}.
    \item[] Guidelines:
    \begin{itemize}
        \item The answer NA means that the paper has no limitation while the answer No means that the paper has limitations, but those are not discussed in the paper. 
        \item The authors are encouraged to create a separate "Limitations" section in their paper.
        \item The paper should point out any strong assumptions and how robust the results are to violations of these assumptions (e.g., independence assumptions, noiseless settings, model well-specification, asymptotic approximations only holding locally). The authors should reflect on how these assumptions might be violated in practice and what the implications would be.
        \item The authors should reflect on the scope of the claims made, e.g., if the approach was only tested on a few datasets or with a few runs. In general, empirical results often depend on implicit assumptions, which should be articulated.
        \item The authors should reflect on the factors that influence the performance of the approach. For example, a facial recognition algorithm may perform poorly when image resolution is low or images are taken in low lighting. Or a speech-to-text system might not be used reliably to provide closed captions for online lectures because it fails to handle technical jargon.
        \item The authors should discuss the computational efficiency of the proposed algorithms and how they scale with dataset size.
        \item If applicable, the authors should discuss possible limitations of their approach to address problems of privacy and fairness.
        \item While the authors might fear that complete honesty about limitations might be used by reviewers as grounds for rejection, a worse outcome might be that reviewers discover limitations that aren't acknowledged in the paper. The authors should use their best judgment and recognize that individual actions in favor of transparency play an important role in developing norms that preserve the integrity of the community. Reviewers will be specifically instructed to not penalize honesty concerning limitations.
    \end{itemize}

\item {\bf Theory assumptions and proofs}
    \item[] Question: For each theoretical result, does the paper provide the full set of assumptions and a complete (and correct) proof?
    \item[] Answer: \answerYes{} %
    \item[] Justification: The background for the theoretical results is provided in Section~\ref{app:sec:background_theory}. The full set of assumptions and complete proofs is provided in Appendix~\ref{app:sec:proofs}.
    \item[] Guidelines:
    \begin{itemize}
        \item The answer NA means that the paper does not include theoretical results. 
        \item All the theorems, formulas, and proofs in the paper should be numbered and cross-referenced.
        \item All assumptions should be clearly stated or referenced in the statement of any theorems.
        \item The proofs can either appear in the main paper or the supplemental material, but if they appear in the supplemental material, the authors are encouraged to provide a short proof sketch to provide intuition. 
        \item Inversely, any informal proof provided in the core of the paper should be complemented by formal proofs provided in appendix or supplemental material.
        \item Theorems and Lemmas that the proof relies upon should be properly referenced. 
    \end{itemize}

    \item {\bf Experimental result reproducibility}
    \item[] Question: Does the paper fully disclose all the information needed to reproduce the main experimental results of the paper to the extent that it affects the main claims and/or conclusions of the paper (regardless of whether the code and data are provided or not)?
    \item[] Answer: \answerYes{} %
    \item[] Justification: Experimental details, including dataset description, evaluation metrics, computing details, and hyperparameter choices, are provided in Appendix~\ref{app:sec:experimental_details}. We will release the full code upon publication.
    \item[] Guidelines:
    \begin{itemize}
        \item The answer NA means that the paper does not include experiments.
        \item If the paper includes experiments, a No answer to this question will not be perceived well by the reviewers: Making the paper reproducible is important, regardless of whether the code and data are provided or not.
        \item If the contribution is a dataset and/or model, the authors should describe the steps taken to make their results reproducible or verifiable. 
        \item Depending on the contribution, reproducibility can be accomplished in various ways. For example, if the contribution is a novel architecture, describing the architecture fully might suffice, or if the contribution is a specific model and empirical evaluation, it may be necessary to either make it possible for others to replicate the model with the same dataset, or provide access to the model. In general. releasing code and data is often one good way to accomplish this, but reproducibility can also be provided via detailed instructions for how to replicate the results, access to a hosted model (e.g., in the case of a large language model), releasing of a model checkpoint, or other means that are appropriate to the research performed.
        \item While NeurIPS does not require releasing code, the conference does require all submissions to provide some reasonable avenue for reproducibility, which may depend on the nature of the contribution. For example
        \begin{enumerate}
            \item If the contribution is primarily a new algorithm, the paper should make it clear how to reproduce that algorithm.
            \item If the contribution is primarily a new model architecture, the paper should describe the architecture clearly and fully.
            \item If the contribution is a new model (e.g., a large language model), then there should either be a way to access this model for reproducing the results or a way to reproduce the model (e.g., with an open-source dataset or instructions for how to construct the dataset).
            \item We recognize that reproducibility may be tricky in some cases, in which case authors are welcome to describe the particular way they provide for reproducibility. In the case of closed-source models, it may be that access to the model is limited in some way (e.g., to registered users), but it should be possible for other researchers to have some path to reproducing or verifying the results.
        \end{enumerate}
    \end{itemize}

\item {\bf Open access to data and code}
    \item[] Question: Does the paper provide open access to the data and code, with sufficient instructions to faithfully reproduce the main experimental results, as described in supplemental material?
    \item[] Answer: \answerYes{} %
    \item[] Justification: We will release the full code and documentation upon publication.
    \item[] Guidelines:
    \begin{itemize}
        \item The answer NA means that paper does not include experiments requiring code.
        \item Please see the NeurIPS code and data submission guidelines (\url{https://nips.cc/public/guides/CodeSubmissionPolicy}) for more details.
        \item While we encourage the release of code and data, we understand that this might not be possible, so “No” is an acceptable answer. Papers cannot be rejected simply for not including code, unless this is central to the contribution (e.g., for a new open-source benchmark).
        \item The instructions should contain the exact command and environment needed to run to reproduce the results. See the NeurIPS code and data submission guidelines (\url{https://nips.cc/public/guides/CodeSubmissionPolicy}) for more details.
        \item The authors should provide instructions on data access and preparation, including how to access the raw data, preprocessed data, intermediate data, and generated data, etc.
        \item The authors should provide scripts to reproduce all experimental results for the new proposed method and baselines. If only a subset of experiments are reproducible, they should state which ones are omitted from the script and why.
        \item At submission time, to preserve anonymity, the authors should release anonymized versions (if applicable).
        \item Providing as much information as possible in supplemental material (appended to the paper) is recommended, but including URLs to data and code is permitted.
    \end{itemize}

\item {\bf Experimental setting/details}
    \item[] Question: Does the paper specify all the training and test details (e.g., data splits, hyperparameters, how they were chosen, type of optimizer, etc.) necessary to understand the results?
    \item[] Answer: \answerYes{} %
    \item[] Justification: All the training and test details are provided in Appendix~\ref{app:sec:experimental_details}.
    \item[] Guidelines:
    \begin{itemize}
        \item The answer NA means that the paper does not include experiments.
        \item The experimental setting should be presented in the core of the paper to a level of detail that is necessary to appreciate the results and make sense of them.
        \item The full details can be provided either with the code, in appendix, or as supplemental material.
    \end{itemize}

\item {\bf Experiment statistical significance}
    \item[] Question: Does the paper report error bars suitably and correctly defined or other appropriate information about the statistical significance of the experiments?
    \item[] Answer: \answerYes{}{} %
    \item[] Justification: While reporting error bars is not yet standard in the field of graph generation, we report the error bars on small synthetic datasets in Appendix~\ref{app:sec:additional_results_synthetic} to mitigate the impact of small test sample size.
    \item[] Guidelines:
    \begin{itemize}
        \item The answer NA means that the paper does not include experiments.
        \item The authors should answer "Yes" if the results are accompanied by error bars, confidence intervals, or statistical significance tests, at least for the experiments that support the main claims of the paper.
        \item The factors of variability that the error bars are capturing should be clearly stated (for example, train/test split, initialization, random drawing of some parameter, or overall run with given experimental conditions).
        \item The method for calculating the error bars should be explained (closed form formula, call to a library function, bootstrap, etc.)
        \item The assumptions made should be given (e.g., Normally distributed errors).
        \item It should be clear whether the error bar is the standard deviation or the standard error of the mean.
        \item It is OK to report 1-sigma error bars, but one should state it. The authors should preferably report a 2-sigma error bar than state that they have a 96\% CI, if the hypothesis of Normality of errors is not verified.
        \item For asymmetric distributions, the authors should be careful not to show in tables or figures symmetric error bars that would yield results that are out of range (e.g. negative error rates).
        \item If error bars are reported in tables or plots, The authors should explain in the text how they were calculated and reference the corresponding figures or tables in the text.
    \end{itemize}

\item {\bf Experiments compute resources}
    \item[] Question: For each experiment, does the paper provide sufficient information on the computer resources (type of compute workers, memory, time of execution) needed to reproduce the experiments?
    \item[] Answer: \answerYes{} %
    \item[] Justification: Computing details are provided in Appendix~\ref{app:sec:computing_details}. Runtime to reproduce some experiments is given in Section~\ref{sec:compare_sota}.
    \item[] Guidelines:
    \begin{itemize}
        \item The answer NA means that the paper does not include experiments.
        \item The paper should indicate the type of compute workers CPU or GPU, internal cluster, or cloud provider, including relevant memory and storage.
        \item The paper should provide the amount of compute required for each of the individual experimental runs as well as estimate the total compute. 
        \item The paper should disclose whether the full research project required more compute than the experiments reported in the paper (e.g., preliminary or failed experiments that didn't make it into the paper). 
    \end{itemize}
    
\item {\bf Code of ethics}
    \item[] Question: Does the research conducted in the paper conform, in every respect, with the NeurIPS Code of Ethics \url{https://neurips.cc/public/EthicsGuidelines}?
    \item[] Answer: \answerYes{} %
    \item[] Justification: The research conducted in this paper conforms with the NeurIPS Code of Ethics.
    \item[] Guidelines:
    \begin{itemize}
        \item The answer NA means that the authors have not reviewed the NeurIPS Code of Ethics.
        \item If the authors answer No, they should explain the special circumstances that require a deviation from the Code of Ethics.
        \item The authors should make sure to preserve anonymity (e.g., if there is a special consideration due to laws or regulations in their jurisdiction).
    \end{itemize}

\item {\bf Broader impacts}
    \item[] Question: Does the paper discuss both potential positive societal impacts and negative societal impacts of the work performed?
    \item[] Answer: \answerYes{} %
    \item[] Justification: The paper discusses both aspects in Appendix~\ref{app:sec:broader_impacts}.
    \item[] Guidelines:
    \begin{itemize}
        \item The answer NA means that there is no societal impact of the work performed.
        \item If the authors answer NA or No, they should explain why their work has no societal impact or why the paper does not address societal impact.
        \item Examples of negative societal impacts include potential malicious or unintended uses (e.g., disinformation, generating fake profiles, surveillance), fairness considerations (e.g., deployment of technologies that could make decisions that unfairly impact specific groups), privacy considerations, and security considerations.
        \item The conference expects that many papers will be foundational research and not tied to particular applications, let alone deployments. However, if there is a direct path to any negative applications, the authors should point it out. For example, it is legitimate to point out that an improvement in the quality of generative models could be used to generate deepfakes for disinformation. On the other hand, it is not needed to point out that a generic algorithm for optimizing neural networks could enable people to train models that generate Deepfakes faster.
        \item The authors should consider possible harms that could arise when the technology is being used as intended and functioning correctly, harms that could arise when the technology is being used as intended but gives incorrect results, and harms following from (intentional or unintentional) misuse of the technology.
        \item If there are negative societal impacts, the authors could also discuss possible mitigation strategies (e.g., gated release of models, providing defenses in addition to attacks, mechanisms for monitoring misuse, mechanisms to monitor how a system learns from feedback over time, improving the efficiency and accessibility of ML).
    \end{itemize}
    
\item {\bf Safeguards}
    \item[] Question: Does the paper describe safeguards that have been put in place for responsible release of data or models that have a high risk for misuse (e.g., pretrained language models, image generators, or scraped datasets)?
    \item[] Answer: \answerNA{} %
    \item[] Justification: Our work focuses on generative modeling for general graphs, but does not release ready-to-use models for real-world applications.
    \item[] Guidelines:
    \begin{itemize}
        \item The answer NA means that the paper poses no such risks.
        \item Released models that have a high risk for misuse or dual-use should be released with necessary safeguards to allow for controlled use of the model, for example by requiring that users adhere to usage guidelines or restrictions to access the model or implementing safety filters. 
        \item Datasets that have been scraped from the Internet could pose safety risks. The authors should describe how they avoided releasing unsafe images.
        \item We recognize that providing effective safeguards is challenging, and many papers do not require this, but we encourage authors to take this into account and make a best faith effort.
    \end{itemize}

\item {\bf Licenses for existing assets}
    \item[] Question: Are the creators or original owners of assets (e.g., code, data, models), used in the paper, properly credited and are the license and terms of use explicitly mentioned and properly respected?
    \item[] Answer: \answerYes{} %
    \item[] Justification: We have properly cited the code, data, and models used in this study (see Appendix~\ref{app:sec:datasets}).
    \item[] Guidelines:
    \begin{itemize}
        \item The answer NA means that the paper does not use existing assets.
        \item The authors should cite the original paper that produced the code package or dataset.
        \item The authors should state which version of the asset is used and, if possible, include a URL.
        \item The name of the license (e.g., CC-BY 4.0) should be included for each asset.
        \item For scraped data from a particular source (e.g., website), the copyright and terms of service of that source should be provided.
        \item If assets are released, the license, copyright information, and terms of use in the package should be provided. For popular datasets, \url{paperswithcode.com/datasets} has curated licenses for some datasets. Their licensing guide can help determine the license of a dataset.
        \item For existing datasets that are re-packaged, both the original license and the license of the derived asset (if it has changed) should be provided.
        \item If this information is not available online, the authors are encouraged to reach out to the asset's creators.
    \end{itemize}

\item {\bf New assets}
    \item[] Question: Are new assets introduced in the paper well documented and is the documentation provided alongside the assets?
    \item[] Answer: \answerYes{} %
    \item[] Justification: The new NetworkX dataset and its generation is fully documented in Appendix~\ref{app:sec:datasets}. It conforms with the License of the NetworkX library.
    \item[] Guidelines:
    \begin{itemize}
        \item The answer NA means that the paper does not release new assets.
        \item Researchers should communicate the details of the dataset/code/model as part of their submissions via structured templates. This includes details about training, license, limitations, etc. 
        \item The paper should discuss whether and how consent was obtained from people whose asset is used.
        \item At submission time, remember to anonymize your assets (if applicable). You can either create an anonymized URL or include an anonymized zip file.
    \end{itemize}

\item {\bf Crowdsourcing and research with human subjects}
    \item[] Question: For crowdsourcing experiments and research with human subjects, does the paper include the full text of instructions given to participants and screenshots, if applicable, as well as details about compensation (if any)? 
    \item[] Answer: \answerNA{} %
    \item[] Justification: This work does not involve crowdsourcing nor research with human subjects.
    \item[] Guidelines:
    \begin{itemize}
        \item The answer NA means that the paper does not involve crowdsourcing nor research with human subjects.
        \item Including this information in the supplemental material is fine, but if the main contribution of the paper involves human subjects, then as much detail as possible should be included in the main paper. 
        \item According to the NeurIPS Code of Ethics, workers involved in data collection, curation, or other labor should be paid at least the minimum wage in the country of the data collector. 
    \end{itemize}

\item {\bf Institutional review board (IRB) approvals or equivalent for research with human subjects}
    \item[] Question: Does the paper describe potential risks incurred by study participants, whether such risks were disclosed to the subjects, and whether Institutional Review Board (IRB) approvals (or an equivalent approval/review based on the requirements of your country or institution) were obtained?
    \item[] Answer: \answerNA{} %
    \item[] Justification: This work does not involve crowdsourcing nor research with human subjects.
    \item[] Guidelines:
    \begin{itemize}
        \item The answer NA means that the paper does not involve crowdsourcing nor research with human subjects.
        \item Depending on the country in which research is conducted, IRB approval (or equivalent) may be required for any human subjects research. If you obtained IRB approval, you should clearly state this in the paper. 
        \item We recognize that the procedures for this may vary significantly between institutions and locations, and we expect authors to adhere to the NeurIPS Code of Ethics and the guidelines for their institution. 
        \item For initial submissions, do not include any information that would break anonymity (if applicable), such as the institution conducting the review.
    \end{itemize}

\item {\bf Declaration of LLM usage}
    \item[] Question: Does the paper describe the usage of LLMs if it is an important, original, or non-standard component of the core methods in this research? Note that if the LLM is used only for writing, editing, or formatting purposes and does not impact the core methodology, scientific rigorousness, or originality of the research, declaration is not required.
    \item[] Answer: \answerNA{} %
    \item[] Justification: The core method development in this research does not involve LLMs.
    \item[] Guidelines:
    \begin{itemize}
        \item The answer NA means that the core method development in this research does not involve LLMs as any important, original, or non-standard components.
        \item Please refer to our LLM policy (\url{https://neurips.cc/Conferences/2025/LLM}) for what should or should not be described.
    \end{itemize}

\end{enumerate}

\newpage
\appendix
\vspace*{0.3cm}
\begin{center}
    {\huge Appendix}
\end{center}
\vspace*{0.5cm}
This appendix provides both theoretical and experimental materials. It is organized as follows: Section~\ref{app:sec:background} provides additional background on sequence models. Section~\ref{app:sec:broader_impacts} provides a discussion about the broader impact of this work. Section~\ref{app:sec:remarks} provides additional details and remarks on our method. Section~\ref{app:sec:proofs} provides proofs for the theorems presented in the main manuscript. Section~\ref{app:sec:experimental_details} provides experimental details. Section~\ref{app:sec:additional_results} provides additional quantitative and qualitative results.

\section{Background on Sequence Model Architectures}\label{app:sec:background}
Our sequence model architectures are fully based on established natural language models. In particular, we consider three prominent models, including GPT-2~\citep{radford2019gpt2}, LLaMA~\citep{touvron2023llama,touvron2023llama2}, and Mamba~\citep{gu2023mamba} to demonstrate the effectiveness of our approach. Notably, our methodology is not restricted to these specific models; it can be applied to any sequence or language model.

\paragraph{GPT-2.}
GPT-2 represents one of the earliest large language models based on the transformer architecture~\citep{vaswani2017attention}. The model employs pre-normalization with LayerNorm, the GeLU activation function, and absolute positional embeddings to encode token positions in sequences. These design choices laid the foundation for many subsequent models.

\paragraph{LLaMA.}
LLaMA~\citep{touvron2023llama,touvron2023llama2} builds upon the transformer framework with several key enhancements. It incorporates pre-normalization through RMSNorm~\citep{zhang2019rmsnorm} and employs the SwiGLU activation function~\citep{shazeer2020glu}. Additionally, LLaMA replaces absolute positional embeddings with rotary positional embeddings~\citep{su2024rope}, enabling better generalization to longer sequences.

\paragraph{Mamba.}
Mamba~\citep{gu2023mamba} is a state-space model (SSM) that maps input sequences to outputs using continuous-time dynamics. It introduces a selection mechanism that dynamically controls how input data flows into hidden states, making the model parameters adaptive to time and data. This innovation enables Mamba to achieve superior performance compared to other SSMs across various tasks.

\section{Broader Impacts}\label{app:sec:broader_impacts}
Our research focuses on advancing the algorithmic development of graph generative models, strongly emphasizing their responsible and ethical application in specialized fields. In domains such as drug discovery and synthetic biology, ensuring the trustworthiness and appropriate use of our methods is essential to prevent potential misuse. Through our experiments, we showcase the potential of our approach in these fields, underscoring its promise to deliver meaningful societal benefits while acknowledging the need to address potential risks.

\section{Additional Details about \method{}}\label{app:sec:remarks}
\subsection{Background on Graph Theory}\label{app:sec:background_theory}
We provide additional background on graph theory necessary for the definitions and theories of SETs and SENTs. The background is largely based on~\citet{diestel2005graph}.

We first give the formal definition of \emph{graph isomorphism}:
\begin{definition}[Graph isomorphism]
    An isomorphism of graphs $G$ and $H$ is a bijection between the vertex sets of $G$ and $H$: $\pi:V_G\to V_H$ such that any two vertices $u$ and $v$ of $G$ are adjacent in $G$ if and only if $\pi(u)$ and $\pi(v)$ are adjacent in $H$, \ie $(u,v)\in E_G$ if and only if $(\pi(u),\pi(v))\in E_H$.
\end{definition}
Graph isomorphism is an equivalence relation on graphs and as such it partitions the class of all graphs into equivalence classes. A set of graphs isomorphic to each other is called an isomorphism class of graphs. It is worth noting that our SENT isomorphism also partitions the class of all SENTs into equivalence classes in a similar fashion.

We also provide the formal definition of \emph{induced subgraph}:
\begin{definition}[Induced subgraph]
    An induced subgraph of a graph is another graph, formed from a subset of the vertices of the graph and all of the edges, from the original graph, connecting pairs of vertices in that subset. Formally, let $S\subseteq V_G$ be any subset of vertices of $G:=(V_G,E_G)$. Then, the induced subgraph $G[S]$ is the graph whose vertex set is $S$ and whose edge set consists of all of the edges in $E_G$ that have endpoints in $S$. That is, for any two vertices $u, v\in S$, $(u,v)\in E_{G[S]}$ if and only if $(u,v)\in E_G$.
\end{definition}

\subsection{Remarks on Tokenized SENTs}\label{app:sec:random_walk_interpretation}
In Section~\ref{sec:tokenization}, we showed that an (ordered) SENT can be converted into a sequence of tokens. Here, we extend this idea by interpreting the tokenized sequence as a random walk on a slightly modified graph. We first introduce an alternative tokenization scheme that is equivalent to the one described earlier but offers enhanced interpretability. The proposed tokenization remains largely unchanged except for how tuples are handled. For each $w=(v,A)$ with $A=\{u_1,\dots,u_p\}$, we now define
\begin{equation*}
    \texttt{Token}(w):=\left[v, \bm{<}, u_1, \bm{<}, u_2\dots, \bm{<}, u_p, \bm{>}\right].
\end{equation*}
We now detail how to modify the original graph $G$: we introduce three virtual nodes, labeled $\textbf{/}$, $\bm{<}$, and $\bm{>}$ respectively. These virtual nodes are connected to all other virtual nodes and original nodes in the graph. This modification ensures that for any non-special token in the tokenized sequence, its subsequent token can either be one of its neighbors or one of the virtual nodes ($\textbf{/}$, $\bm{<}$, or $\bm{>}$). Consequently, each token has a direct connection to the node corresponding to the current token, and the language model amounts to learning the state transition functions for these random walks. Since these random walks are non-Markovian, this perspective further justifies our choice of using autoregressive models instead of one-step generative models. Furthermore, as random walks are random sequences on graphs, sampling random walks amounts to sampling from those random sequences. 

\subsection{Remarks on Model Inference}
The model inference is straightforward following the same process as LLMs such as LLaMA~\citep{touvron2023llama,touvron2023llama2}. An alternative way is to enforce the semantic correctness of the generated sequences of tokens by adjusting the logits at a certain token to obey the semantic rule of the tokenization. For instance, the token `$\bm{>}$' can only occur after a token `$\bm{<}$' or no special tokens can appear right after $\textbf{/}$. We manually implemented these transition constraints and incorporated them into the inference. We compared this strategy with the constraint-free counterpart. Surprisingly, our experiments demonstrate that the constraint-free variant could always generate semantically correct sequences and perform similarly to the one with the transition constraints on unattributed graphs. Therefore, we did not use any transition constraints during the inference in our unattributed graph experiments. On the other hand, for attributed graphs, we found that the constraint-free decoding might generate semantically incorrect sequences. As a consequence, we used constrained decoding in all molecular generation experiments. In practice, we recommend using constrained decoding for attributed and complex graphs. 

\section{Proofs}\label{app:sec:proofs}
In this section, we provide proof for the theorems stated in the manuscript.
\printProofs

\section{Experimental Details}\label{app:sec:experimental_details}

\subsection{Datasets}\label{app:sec:datasets}
We provide details of the datasets used in our experiments. We adopt the standard train/validation/test splits provided in the original sources. The statistics about the datasets are summarized in Table~\ref{app:tab:datasets}.

\paragraph{Small synthetic graphs: Planar and SBM.}
Both of these datasets are from \citet{martinkus2022spectre}. The Planar dataset consists of 200 planar graphs with 64 nodes each, generated via Delaunay triangulation on points uniformly sampled in the unit square. The SBM dataset contains 200 graphs comprising 2 to 5 communities, with each community having between 20 and 40 nodes. An edge is placed between two nodes with probability 0.3 if they belong to the same community, and 0.05 otherwise. We follow the same splits as \citet{martinkus2022spectre}.

\paragraph{Large graphs: Proteins and Point Clouds.}
The Proteins dataset includes graph representations (contact maps) of proteins from \citet{dobson2003distinguishing}. In these graphs, each node represents an amino acid, and an edge connects two nodes if their corresponding amino acids are within 6 angstroms of each other. We use the same data splits as \citet{liao2019efficient}. The Point Clouds dataset, also from \citet{liao2019efficient}, consists of 41 point clouds of household objects \citep{neumann2013graph}. As many of these graphs are disconnected, we retain only the largest connected component of each, following \citet{bergmeister2024efficient}, and again employ the splits used by \citet{liao2019efficient}.

\paragraph{QM9.} 
The QM9 dataset, from \citet{wu2018moleculenet}, comprises small molecules with up to nine heavy atoms (carbon, oxygen, nitrogen, and fluorine). In this work, we adopt the more challenging setting proposed by \citet{vignac2023digress}, where hydrogen atoms are modeled explicitly, and we follow the same data splits as in that reference.

\paragraph{MOSES and GuacaMol.} 
The MOSES and GuacaMol datasets are obtained from the respective benchmark tools of \citet{polykovskiy2020moses} and \citet{brown2019guacamol}. Both consist of drug-like molecules, with those in GuacaMol typically being larger on average. For each dataset, we convert generated molecular graphs to SMILES using the code from \citet{jo2022score}, which permits partial charges. We employ the standard data splits provided by the corresponding benchmarks.

\paragraph{PubChem-10M.}
PubChem-10M is a subset of about 10M molecules from PubChem curated by~\citet{chithrananda2020chemberta}.

\paragraph{NetworkX.}
We generate the graphs using the generators from the NetworkX library\footnote{\url{https://networkx.org/documentation/stable/reference/generators.html}}~\citep{hagberg2008networkx}, categories including ``Classic'', ``Lattice'', ``Small'', ``Random Graphs'', ``Geometric'', ``Trees'', ``Community'', ``Social Networks''. We ensure that this dataset \emph{does not contain any graphs in the downstream datasets}. The summary of the code for generating these graphs is provided in Table~\ref{app:tab:networkx_dataset}. Notably, the largest graph has up to 5999 nodes.

\begin{table}[ht]
    \centering
    \caption{Dataset statistics}
    \label{app:tab:datasets}
    \begin{sc}
    \resizebox{\textwidth}{!}{
    \begin{tabular}{lccccccc} \toprule
        Dataset & \multicolumn{3}{c}{$n_{\mathrm{graphs}}$} & $|V|_{\max}$ & $|V|_{\avg}$ & $|E|_{\max}$ & $|E|_{\avg}$ \\ \cmidrule{2-4}
        & Train & Val & Test & \\ \midrule
        Unattributed Graphs \\
        Planar & 128 & 32 & 40 & 64 & 64 & 181 & 178 \\ 
        SBM & 128 & 32 & 40 & 187 & 104 & 1129 & 500 \\
        Proteins & 587 & 147 & 184 & 500 & 258 & 1575 & 646 \\
        Point Clouds & 26 & 7 & 8 & 5037 & 1332 & 10886 & 2971 \\ \midrule
        Attributed Graphs \\
        QM9 & 97734 & 20042 & 13055 & 29 & 18 & 28 & 19 \\
        MOSES & 1584663 & 176225 & 176074 & 27 & 22 & 31 & 23  \\
        GuacaMol & 1118633 & 69926 & 209654 & 88 & 28 & 88 & 30  \\ \midrule
        Pre-training unattributed graphs \\
        NetworkX & 24957 & 2516 & --- & 5999 & 459 & 5999 & 751 \\ 
        \bottomrule
    \end{tabular}
    }
    \end{sc}
\end{table}

\begin{table}[ht]
    \centering
    \caption{Summary of the code for generating graphs in the NetworkX dataset}
    \label{app:tab:networkx_dataset}
    \begin{sc}
    \begin{lrbox}{\tablebox}
    \begin{tabular}{lcc}\toprule
        Generator & $n_{\mathrm{graphs}}$ & Python Code \\ \midrule
        \textbf{Category: Classic} \\
        balanced tree & 10 & \verb|nx.balanced_tree(2, np.random.randint(4, 10))| \\
        barbell graph & 100 & \verb|nx.barbell_graph(np.random.randint(3, 31), np.random.randint(41))| \\
        binomial tree &  10 & \verb|nx.binomial_tree(np.random.randint(2, 9))| \\
        complete graph & 10 & \verb|nx.complete_graph(np.random.randint(3, 31))| \\
        circular ladder graph & 300 & \verb|nx.circular_ladder_graph(np.random.randint(10, 501))| \\
        cycle graph & 2000 & \verb|nx.cycle_graph(np.random.randint(10, 6001))| \\
        Dorogovtsev Goltsev Mendes graph & 5 & \verb|nx.dorogovtsev_goltsev_mendes_graph(np.random.randint(2, 7))| \\
        ladder graph & 500 & \verb|nx.ladder_graph(np.random.randint(10, 1001))| \\
        lollipop graph & 200 & \verb|nx.lollipop_graph(np.random.randint(3, 21), np.random.randint(10, 51))| \\
        star graph & 200 & \verb|nx.star_graph(np.random.randint(10, 501))| \\
        turan graph & 100 & \verb|nx.turan_graph(np.random.randint(10, 41), 2)| \\
        wheel graph & 100 & \verb|nx.wheel_graph(np.random.randint(10, 201))| \\ \midrule
        \textbf{Category: Lattices} \\
        grid 2d graph & 400 & \verb|nx.grid_2d_graph(np.random.randint(5, 31), np.random.randint(5, 31))| \\
        triangular lattice graph & 400 & \verb|nx.triangular_lattice_graph(np.random.randint(5, 41), np.random.randint(5, 41))| \\ \midrule
        \textbf{Category: Small} \\
        all but the LCF graph & 1 (each) & \verb|nx.{method}()| \\ \midrule
        \textbf{Category: Random graphs} \\
        Erdos Renyi graph & 4000 & \verb|nx.erdos_renyi_graph(np.random.randint(20, 101), 0.2)| \\
        random regular graph & 2000 & \verb|nx.random_regular_graph(np.random.randint(3, 11), np.random.choice([20,30,...,500]))| \\
        Barabasi Albert graph & 4000 & \verb|nx.barabasi_albert_graph(np.random.randint(20, 501), np.random.randint(2, 6))| \\ 
        random lobster & 4000 & \verb|nx.random_lobster(80, 0.7, 0.7)| \\ \midrule
        \textbf{Category: Geometric} \\
        random geometric graph & 3000 & \verb|nx.random_geometric_graph(np.random.choice([20,30,...,100]), 0.3)| \\
        Waxman graph & 2000 & \verb|nx.waxman_graph(np.random.choice([50,100,150,...,300]))| \\ \midrule
        \textbf{Category: Trees} \\
        random unlabeled tree & 1000 & \verb|nx.random_unlabeled_tree(np.random.randint(20, 501))| \\ \midrule
        \textbf{Category: Community} \\ 
        connected Caveman graph & 300 & \verb|nx.connected_caveman_graph(np.random.randint(10, 101), np.random.randint(2, 5))| \\
        Windmill graph & 300 & \verb|nx.windmill_graph(np.random.randint(10, 101), np.random.randint(2, 5))| \\ \midrule
        \textbf{Category: Social networks} \\
        All social networks & 1 (each) & \verb|nx.{method}()| \\
        \bottomrule
    \end{tabular}
    \end{lrbox}
    \resizebox{\textwidth}{!}{\usebox{\tablebox}}
    \end{sc}
\end{table}

\subsection{Evaluation Metrics}\label{app:sec:evaluation_metrics}

We follow \citet{martinkus2022spectre} and \citet{vignac2023digress} in comparing our model’s performance with other graph generative approaches. Specifically, we measure the maximum mean discrepancy (MMD) between the generated and test graphs for degree distribution, clustering coefficient, orbit counts, and spectrum. As a reference, we also compute these metrics on the training set and report the mean ratio across all properties as a global indicator of statistical discrepancy between the generated samples and test samples. Note that for the Point Clouds dataset, which is defined by a $k$-nearest-neighbor structure, the degree MMD is always zero and is therefore excluded from the mean ratio. While we utilize these metrics to maintain consistency with previous research, we acknowledge their limitations, particularly regarding arbitrary kernel hyperparameter selection, as highlighted by~\citet{obray2022evaluation,thompson2022evaluation}. In short, MMD measures the distributional similarity between generated and real graphs. Lower MMD scores indicate that the generated graphs' statistics (e.g., degree, clustering coefficients) more closely match the training data. 

We additionally track uniqueness and novelty: \emph{uniqueness} is the fraction of generated graphs that are not isomorphic to each other, and \emph{novelty} is the fraction of generated graphs that are not isomorphic to any training graph.

Below, we describe additional metrics specific to each dataset.

\paragraph{Planar and SBM.} 
Following \citet{martinkus2022spectre}, we report a \emph{validity score} for synthetic datasets. For Planar graphs, it verifies whether the generated graphs remain planar; for SBM graphs, it measures how likely they are to be generated under the original SBM parameters. We integrate validity, novelty, and uniqueness into a single metric, VUN, which measures the fraction of generated graphs that are simultaneously valid, novel, and unique. In short, VUN assesses sample quality. 

\paragraph{QM9.} 
For QM9, we report the \emph{validity}, \emph{uniqueness}, and \emph{novelty} defined for general molecules, as described in the following paragraph.
We also report \emph{atom stability} and \emph{molecule stability} as defined by \citet{hoogeboom2022equivariant} and \citet{vignac2023digress}.

\paragraph{MOSES and GuacaMol.} Since MOSES \citep{polykovskiy2020moses} and GuacaMol \citep{brown2019guacamol} are benchmarking platforms, each comes with its own suite of metrics, which we use to evaluate our model. These include: 
\begin{itemize} 
\item \textbf{Validity}: Proportion of molecules passing basic valency checks. 
\item \textbf{Uniqueness}: Proportion of generated molecules with distinct SMILES strings (indicating non-isomorphic structures). 
\item \textbf{Novelty}: Proportion of generated molecules not present in the training set. 
\item \textbf{Filter score}: Proportion of molecules passing the same filters used to create the test set. 
\item \textbf{Fréchet ChemNet Distance (FCD)}: Similarity measure between generated and training sets based on learned neural embeddings. 
\item \textbf{SNN}: Similarity to the nearest neighbor, computed via Tanimoto distance. 
\item \textbf{Scaffold similarity}: Comparison of Bemis–Murcko scaffold frequencies. 
\item \textbf{KL divergence}: Differences in the distributions of various physicochemical descriptors. 
\end{itemize}

\subsection{Computing Details}\label{app:sec:computing_details}
We implemented our sequence models using the model hub of Hugging Face. Users can easily test their preferred sequence or language models using our code.
Experiments were conducted on a shared computing cluster with various CPU and GPU configurations,
including 16 NVIDIA H100 (80GB) GPUs. Each experiment was allocated
resources on a single GPU, along with 8 CPUs and up to 48GB of system RAM. The run-time of
each model was measured on a single NVIDIA H100 GPU.

\subsection{Hyperparameters}
Unlike prior studies that adjust model sizes across datasets, we maintain a consistent model architecture and size throughout all experiments, specifically using the small GPT configuration (768 hidden dimensions, 12 layers, 12 attention heads). Training hyperparameters are aligned with established practices from popular LLMs such as GPT-3~\citep{gpt3} and LLaMA~\citep{touvron2023llama}. We fix the context length to 2048 and use a batch size of 128 if possible, otherwise 64 for larger graphs. In particular, we employ the AdamW optimizer with a gradient clipping threshold of 1.0, a weight decay of 0.1, and a learning rate schedule with a linear warmup followed by cosine decay, peaking at 6e-4. The AdamW hyperparameters are set to $\beta=(0.9, 0.95)$. Due to the small dataset sizes of previous benchmarks, we tune the only training hyperparameter dropout in $\{0, 0.5\}$, and find the model achieves better validation loss with the value of 0.5 on the small synthetic datasets. Each model was trained for 200000, 400000, or 800000 iterations, depending on the dataset size.

Inference hyperparameters, including $k$ (top-k sampling) and $\tau$ (temperature), are reported in Table~\ref{app:tab:inference_hyperparameters} and analyzed in detail in Section~\ref{app:sec:ablation}. 

\begin{table}[tbp]
    \centering
    \caption{Inference hyperparameters for each dataset.}
    \label{app:tab:inference_hyperparameters}
    \begin{sc}
    \begin{tabular}{lcccc}\toprule
        & \multicolumn{2}{c}{w/o pre-training} & \multicolumn{2}{c}{w/ pre-training} \\ \cmidrule{2-5}
        Dataset & Top-$k$ & Temperature $\tau$ & Top-$k$ & Temperature $\tau$ \\  \midrule
        Planar & 10 & 1.0 & 30 & 0.9 \\
        SBM & 60 & 1.0 & 150 & 1.0 \\
        Proteins & 40 & 1.0 & 30 & 1.05 \\
        Point Clouds & 60 & 1.0 & 20 & 0.9 \\ 
        NetworkX & 120 & 1.0 & --- & --- \\ \midrule
        QM9 & 5 & 1.0 & --- & ---\\
        MOSES & 5 & 1.0 & --- & --- \\
        GuacaMol & 5 & 1.0 & --- & --- \\ \bottomrule
    \end{tabular}
    \end{sc}
\end{table}

\section{Additional Results}\label{app:sec:additional_results}

\subsection{Additional Results on Synthetic Datasets}\label{app:sec:additional_results_synthetic}
Due to the small number of samples in the Planar and SBM datasets, we observe significant variances in evaluation metrics. In order to mitigate the impact of the small test samples on the evaluation, we use trained models to generate samples with 10 different seeds and report the average metrics and error bars for DiGress, ESGG, and our \method{}. For DiGress and ESGG, we use either pretrained models released by the authors, if available, or our reproduced models using their officially released code repository. As shown in Table~\ref{app:tab:planar_sbm_error_bars}, the variances appear reasonable, and the conclusions remain the same.

\begin{table}[tbp]
    \centering
    \caption{Performances on the Planar and SBM datasets with error bars}\label{app:tab:planar_sbm_error_bars}
    \begin{subtable}{\textwidth}
    \caption{Planar}\label{app:tab:planar_error_bars}
        \begin{sc}
        \resizebox{\textwidth}{!}{
        \begin{tabular}{lcccccc}\toprule
        Model & Deg. & Clus. & Orbit & Spec. & Ratio & VUN  \\ \midrule
        DiGress   & 0.0003±0.0002 & 0.0415±0.0081 & 0.0056±0.0028 & 0.0078±0.0010 & 3.8±1.4     & 79.0±6.0     \\
        ESGG      & 0.0006±0.0004 & 0.0434±0.0154 & 0.0101±0.0073 & 0.0091±0.0018 & 6.3±4.0     & \textbf{90.5±5.6}     \\
        \method{} & 0.0004±0.0003 & 0.0533±0.0083 & 0.0005±0.0004 & 0.0066±0.0009 & \textbf{1.6±0.5}     & 80.3±6.8     \\ \bottomrule
        \end{tabular}
        }
        \end{sc}
    \end{subtable}
    \begin{subtable}{\textwidth}
    \caption{SBM}\label{app:tab:sbm_error_bars}
        \begin{sc}
        \resizebox{\textwidth}{!}{
        \begin{tabular}{lcccccc}\toprule
        Model & Deg. & Clus. & Orbit & Spec. & Ratio & VUN  \\ \midrule
        DiGress   & 0.0013±0.0009 & 0.0501±0.0009 & 0.0393±0.0104 & 0.0053±0.0007 & \textbf{1.6±0.3} & 66.0±5.6     \\
        ESGG      & 0.0468±0.0096 & 0.0554±0.0013 & 0.0699±0.0051 & 0.0085±0.0011 & 15.3±3.1    & 16.0±4.7     \\
        \method{} & 0.0081±0.0051 & 0.0525±0.0015 & 0.0687±0.0138 & 0.0048±0.0011 & 3.9±1.6     & \textbf{88.3±4.7}     \\ \bottomrule
        \end{tabular}
        }
        \end{sc}
    \end{subtable}
\end{table}

\subsection{Additional Results on molecular generation datasets}\label{app:sec:additional_results_molecules}
Due to space constraints, we provide results on the QM9 dataset in Table~\ref{tab:qm9}, and additional metrics on the MOSES dataset in Table~\ref{app:tab:moses}. It is worth noting that all results on MOSES and GuacaMol were obtained using the benchmarking tools from the original works, which might rely on outdated packages. For instance, using the latest FCD package gives a FCD score of 94.7 for the pre-trained \method{}.

Our results on all three benchmarks demonstrate the immense potential of \method{} for molecular generation. The fact that AutoGraph, by learning from graph data alone, can achieve validity scores competitive with these specialized models is a strong demonstration of its learning capabilities. It successfully infers the complex, implicit rules of molecular construction. Furthermore, some models, such as MCTS a search-based method, that guarantee 100\% validity do so at the cost of poor distributional similarity (e.g., a low FCD score). From a practical perspective, one can easily reject invalid graphs to achieve near-perfect validity while generating over-simplified and basic molecules that have low distributional similarity (FCD score) with the training data, which is of low practical interest. AutoGraph achieves a superior overall balance. 

\begin{table}[tbp]
    \centering
    \vskip -0.1in
    \caption{Benchmarking \method{} on the QM9 dataset}
    \label{tab:qm9}
    \begin{small}
    \begin{sc}
    \resizebox{.7\textwidth}{!}{
    \begin{tabular}{lccccc}\toprule
         & \multicolumn{5}{c}{QM9 with hydrogen atoms} \\ %
         & \multicolumn{5}{c}{$n_{\mathrm{graphs}}=100$K, $|V|_{\max}=29$, $|V|_{\avg}\approx 18$} \\ \cmidrule{2-6}
        Model & Valid$\shortuparrow$ & Unique$\shortuparrow$ & Novel$\shortuparrow$ & Atom stable$\shortuparrow$ & Mol stable$\shortuparrow$ \\ \midrule
        DiGress & 95.4 & \textbf{97.6} & 33.4 & 98.1 & 79.8 \\ \midrule
        \method{} & \textbf{97.7} & 96.7 & \textbf{45.5} & \textbf{98.6} &  \textbf{87.3}\\
        \bottomrule
    \end{tabular}
    }
    \end{sc}
    \end{small}
    \vskip -0.1in
\end{table}

\begin{table}[tbp]
    \centering
    \vskip -0.1in
    \caption{Benchmarking \method{} on the MOSES dataset}
    \label{app:tab:moses}
    \begin{small}
    \begin{sc}
    \resizebox{\textwidth}{!}{
    \begin{tabular}{llccccccc}\toprule
         & & \multicolumn{7}{c}{MOSES} \\ %
         & & \multicolumn{7}{c}{$n_{\mathrm{graphs}}= 1.58$M, $|V|_{\max}=27$, $|V|_{\avg}\approx 22$} \\ \cmidrule{3-9}
        Model & Type & Valid$\shortuparrow$ & Unique$\shortuparrow$ & Novel$\shortuparrow$ & Filters$\shortuparrow$ & FCD$\shortdownarrow$ & SNN$\shortdownarrow$ & Scaf$\shortuparrow$ \\ \midrule
        VAE & SMILES & 97.7 & 99.8 & 69.5 & 99.7 & 0.57 & 0.58 & 5.9 \\
        JT-VAE & Fragments & 100 & 100 & 99.9 & 97.8 & 1.00 & 0.53 & 10 \\
        GraphINVENT & Graph & 96.4 & 99.8 & – & 95.0  & 1.22 & 0.54 & 12.7 \\
        DiGress & Graph & 85.7 & 100 & 95.0 & 97.1 & 1.19 & 0.52 & 14.8 \\ \midrule
        \method{} & Graph & 87.4 & 100 & 85.9 & 98.6 & 0.91 & 0.55 & 10.2  \\
        \bottomrule
    \end{tabular}
    }
    \end{sc}
    \end{small}
    \vskip -0.1in
\end{table}

\subsection{Transfer Performance of \method{}}\label{app:sec:transfer}
\begin{figure}
    \centering
    \includegraphics[width=0.8\linewidth]{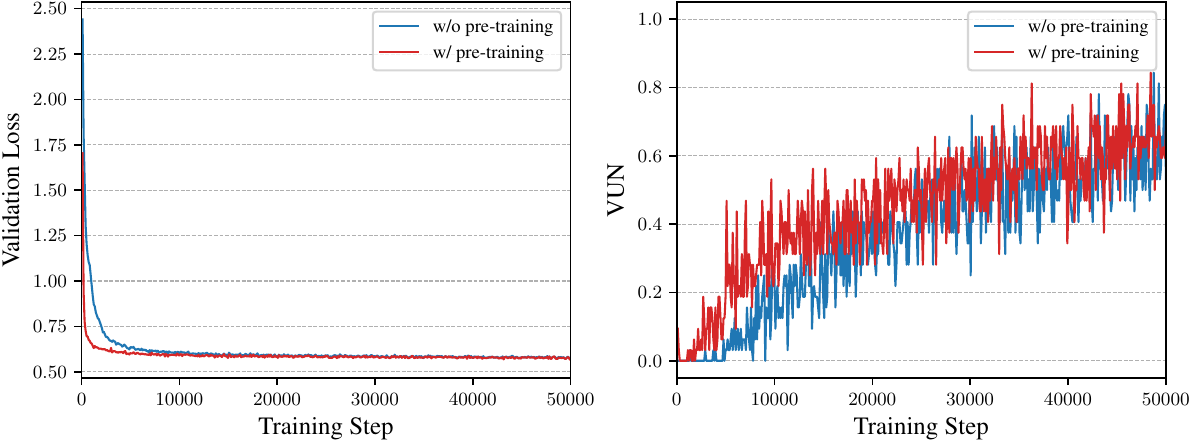}
    \caption{Comparison of \method{} with and without pre-training on the Planar dataset with 50000 training steps. The model with pre-training converges clearly faster than the model without pre-training.}
    \label{app:fig:with_vs_without_pretraining}
\end{figure}

We provide here additional results for the transfer learning of \method{}. We compare the training curves of \method{} models with and without pre-training on the Planar datasets in Figure~\ref{app:fig:with_vs_without_pretraining}. The result suggests that the model with pre-training converges clearly faster.

For the transfer experiment on molecular generation, we first pre-train \method{} on the PubChem-10M dataset~\citep{chithrananda2020chemberta}, and then fine-tune it on the GuacaMol dataset. In this experiment, we use richer node attributes including atom types, total number of hydrogens, and formal charges.

\subsection{Substructure Conditioned Generation}\label{app:sec:substructure_conditioned_generation}

\begin{figure}[!ht]
    \centering
    \includegraphics[width=\linewidth]{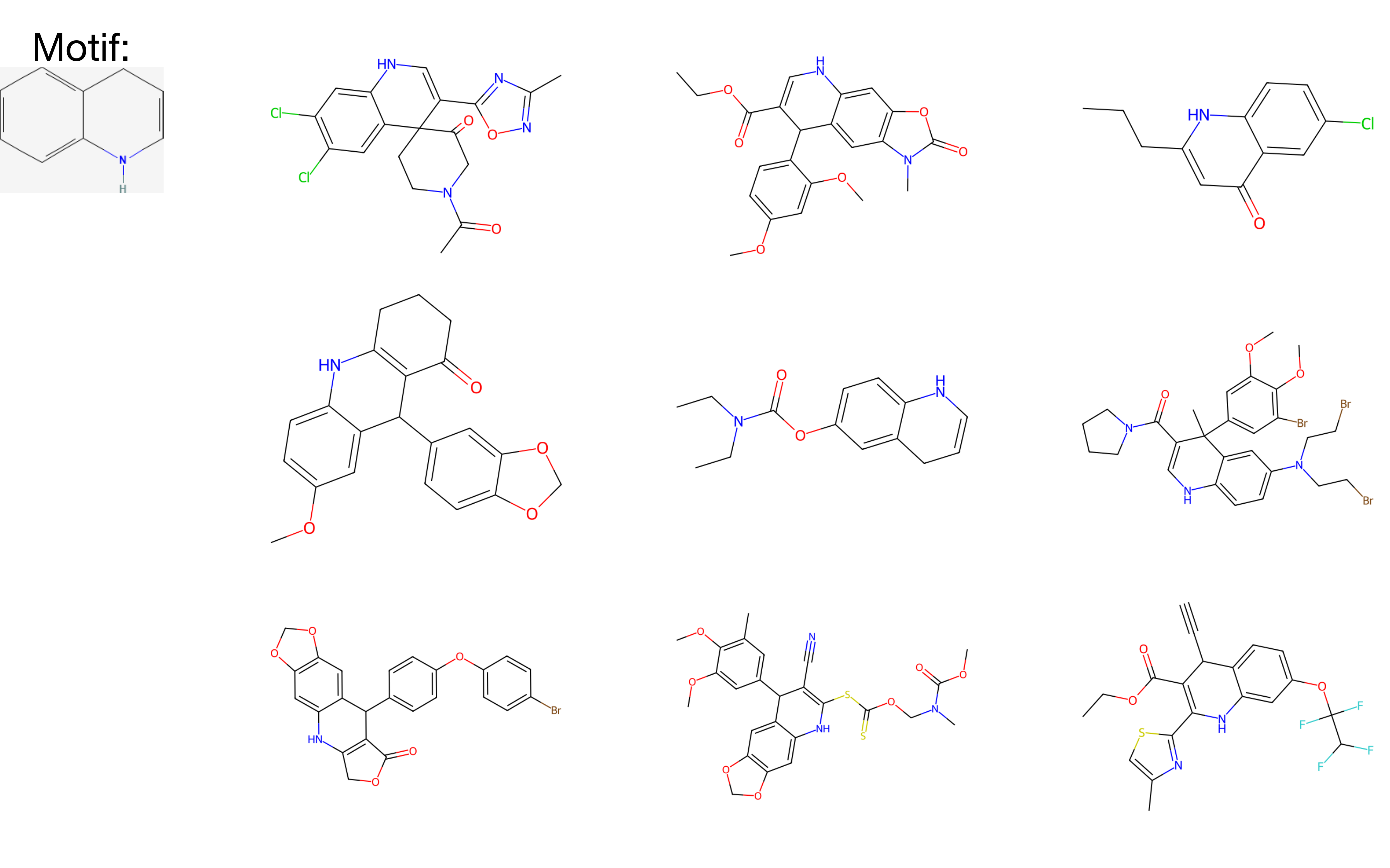}
    \caption{Substructure conditioned generation on one copy of the motif 1\_4-Dihydroquinoline.}
    \label{app:fig:1_motif_samples}
\end{figure}

\begin{figure}[!ht]
    \centering
    \includegraphics[width=\linewidth]{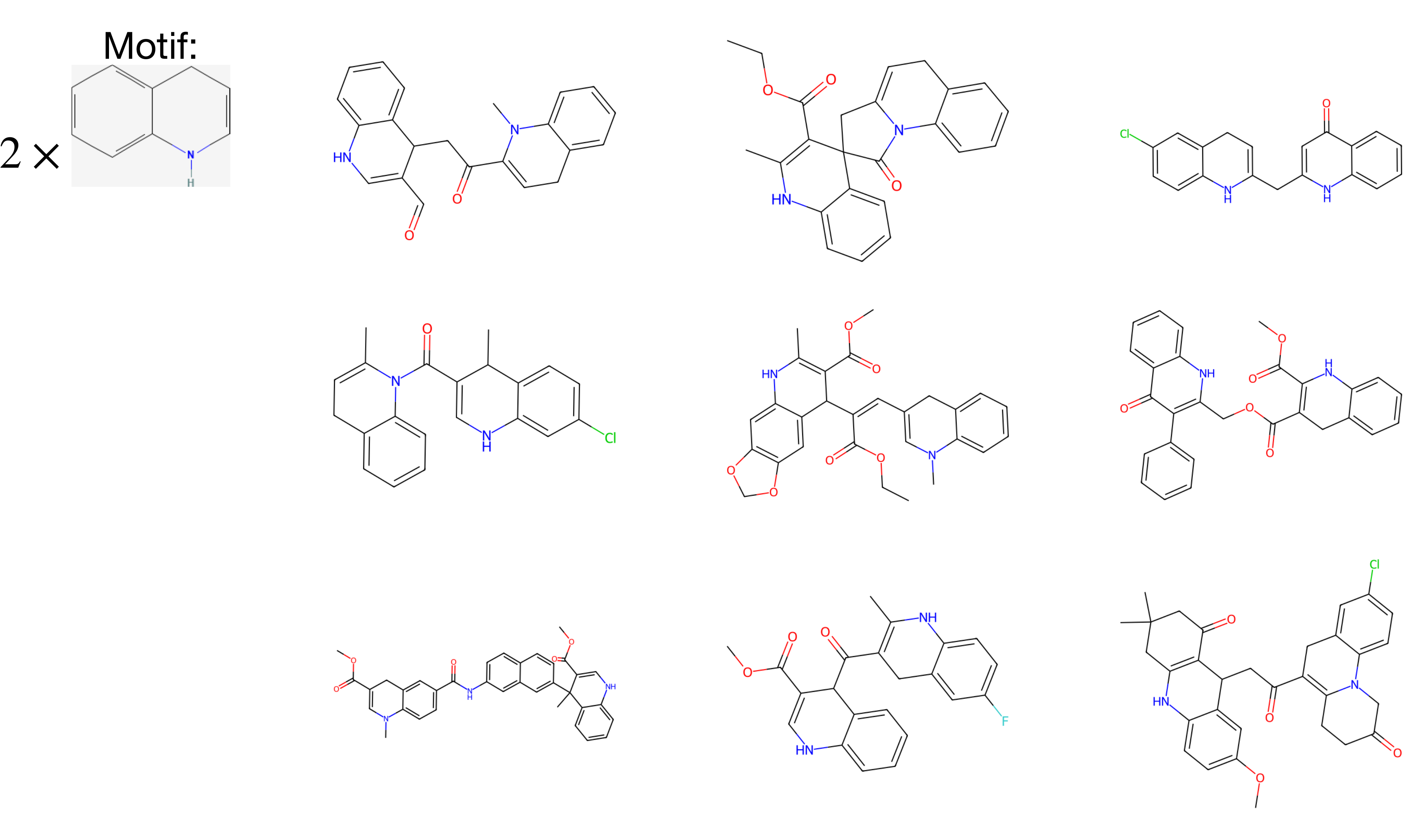}
    \caption{Substructure conditioned generation on two copies of the motif 1\_4-Dihydroquinoline.}
    \label{app:fig:2_motif_samples}
\end{figure}

\begin{figure}[!ht]
    \centering
    \includegraphics[width=\linewidth]{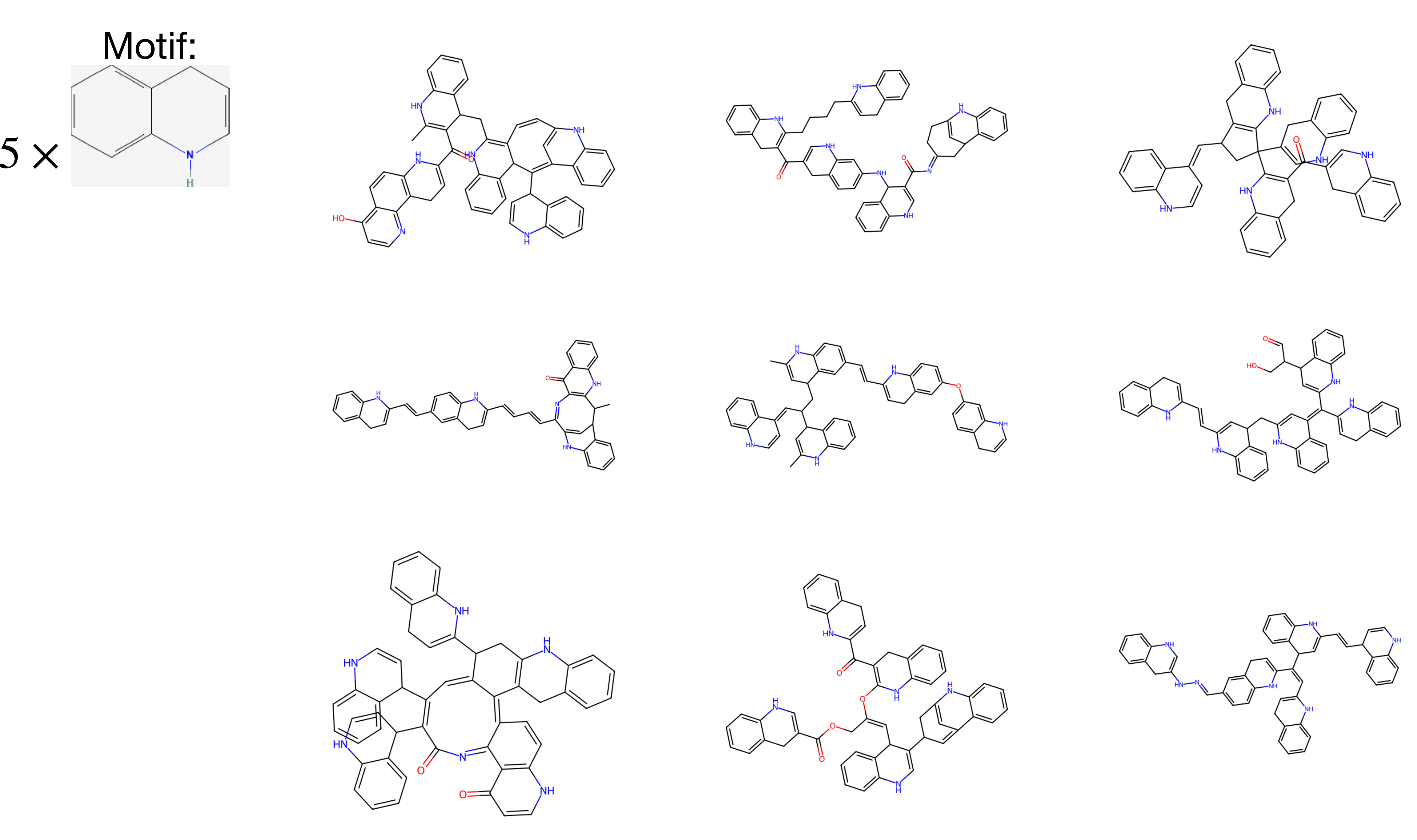}
    \caption{Substructure conditioned generation on five copies of the motif 1\_4-Dihydroquinoline.}
    \label{app:fig:5_motif_samples}
\end{figure}

As presented in Section~\ref{sec:substructure_conditioned_generation}, we test more extreme cases by replicating the same motif multiple times before initiating the conditional generation. Figure~\ref{app:fig:1_motif_samples}, \ref{app:fig:2_motif_samples}, and \ref{app:fig:5_motif_samples} demonstrate non-curated samples generated by \method{} (trained on the GuacaMol dataset without any additional fine-tuning) conditioned on $p$ copies of the same motif, where $p=1, 2, 5$ respectively.

\begin{figure}
    \centering
    \includegraphics[width=\textwidth]{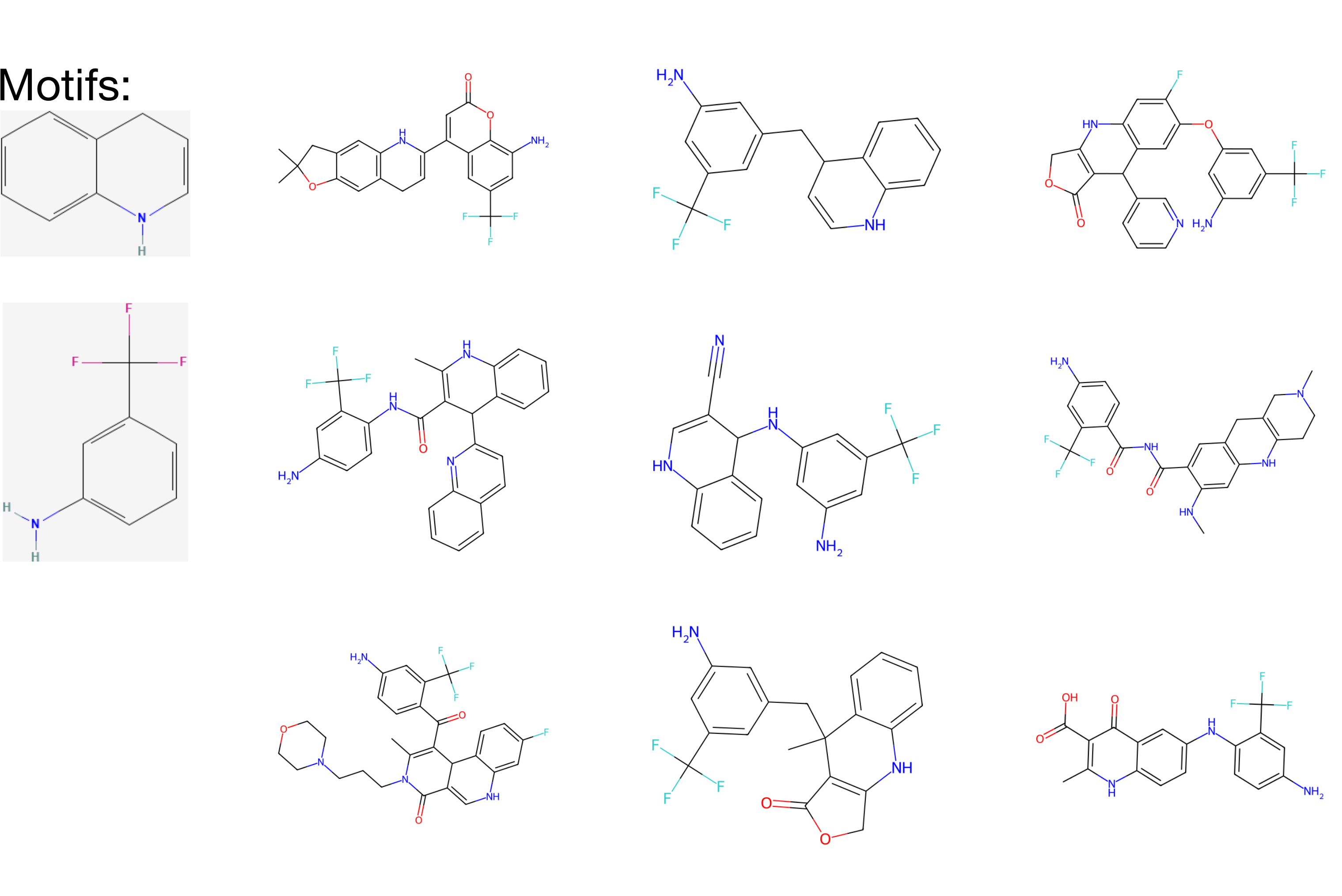}
    \caption{Substructure conditioned generation on two different motifs: 1\_4-Dihydroquinoline and 3-(Trifluoromethyl)aniline.}
    \label{app:fig:linker generation}
\end{figure}

To further showcase the flexibility of \method{}, we conduct the same experiments for two different motifs: 1,4-Dihydroquinoline\footnote{\url{https://pubchem.ncbi.nlm.nih.gov/compound/1_4-Dihydroquinoline}} and 3-(Trifluoromethyl)aniline\footnote{\url{https://pubchem.ncbi.nlm.nih.gov/compound/3-_Trifluoromethyl_aniline}}. This is a very relevant problem in drug discovery, usually termed linker design. The validity, uniqueness, and novelty for 1024 samples are respectively 97.4, 81.4, and 99.9. Visual examples are given in Figure~\ref{app:fig:linker generation}.

\subsection{Additional Ablation Experiments}\label{app:sec:ablation}

\begin{figure}[!ht]
    \centering
    \includegraphics[width=0.5\linewidth]{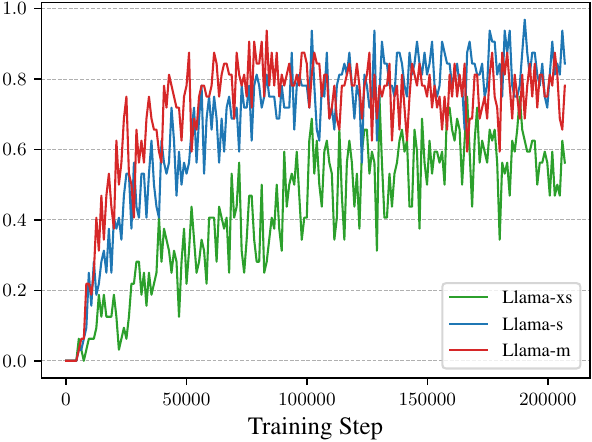}
    \caption{Comparison of model size on the Planar dataset: VUN score vs training steps. LLaMA-m appears to suffer from overfitting and LLaMA-xs appears to suffer from underfitting.}
    \label{app:fig:model_size}
\end{figure}

\paragraph{Impact of model size.}
Table~\ref{app:tab:model_size} and Figure~\ref{app:fig:model_size} compare the impact of model size. Larger models demonstrate better VUN scores. In this work, we use LLaMA-s in all our experiments to balance the trade-off between performance and speed.

\begin{table}[tbp]
    \centering
    \caption{Comparison of model size on the Planar dataset}
    \label{app:tab:model_size}
    \begin{sc}
    \resizebox{\textwidth}{!}{
    \begin{tabular}{lllcccccc}\toprule
        Model & \# Params & Configuration & Deg. & Clus. & Orbit & Spec. & Ratio & VUN  \\ \midrule
        LLaMA-xs & 25.2M & 6 layers, 512 dims & -0.0001 & 0.0570 & 0.0006 & 0.0063 & \textbf{1.0} & 60.0 \\ 
        LLaMA-s & 113M & 12 layers, 768 dims & 0.0005 & 0.0651 & 0.0005 & 0.0056 & 1.6 & \textbf{90.0} \\ 
        LLaMA-m & 402M & 24 layers, 1024 dims & 0.0001 & 0.0340 & 0.0013 & 0.0064 & 1.4 & 82.5 \\ 
        \bottomrule
    \end{tabular}
    }
    \end{sc}
\end{table}

\subsection{Visualization of Graphs Generated by \method{}}

\subsubsection{Results without Pre-training}
\begin{figure}[!ht]
    \centering
    \includegraphics[width=0.6\linewidth]{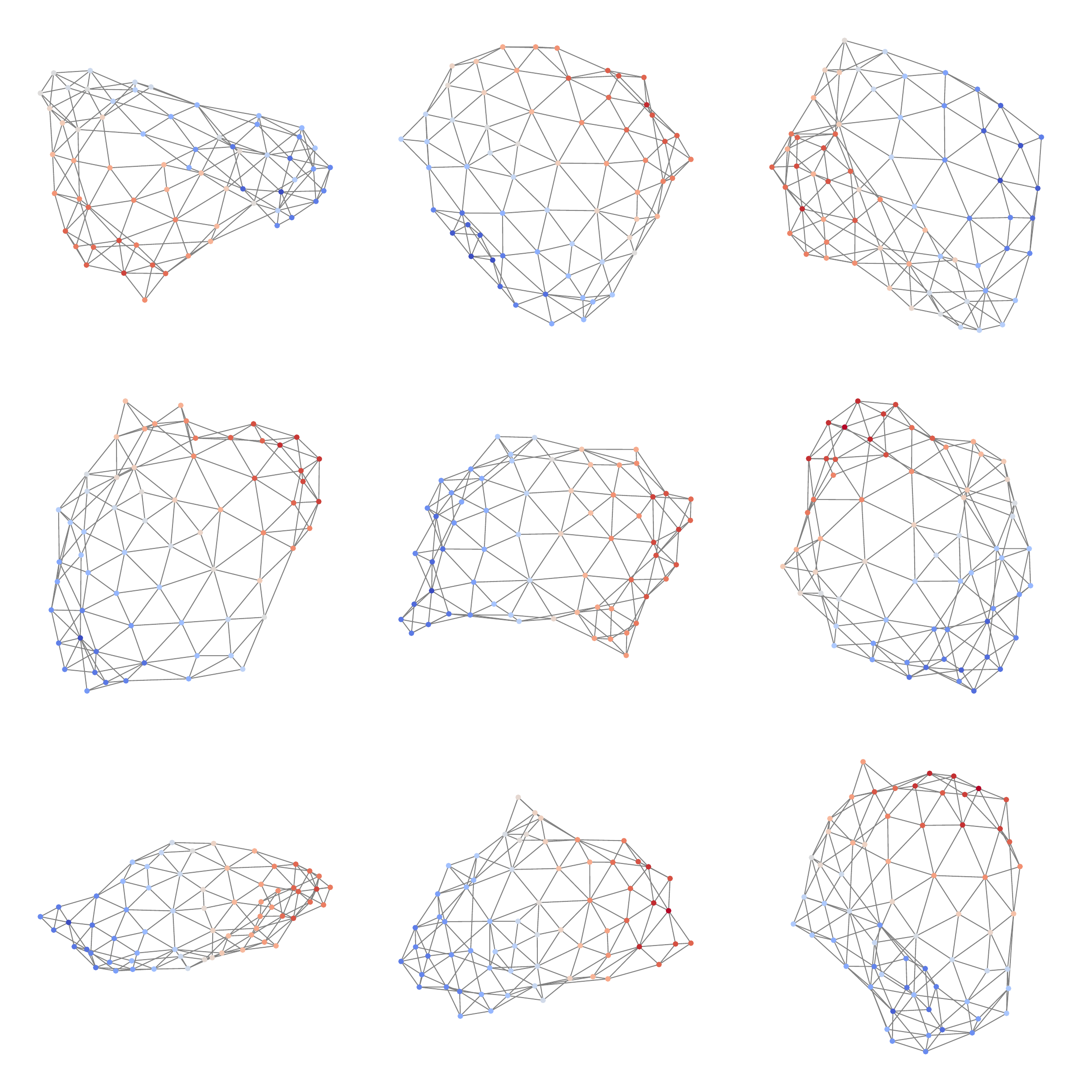}
    \caption{Non-curated samples generated by \method{} (without pre-training) trained on the Planar dataset.}
    \label{app:fig:planar_samples}
\end{figure}

\begin{figure}[!ht]
    \centering
    \includegraphics[width=0.6\linewidth]{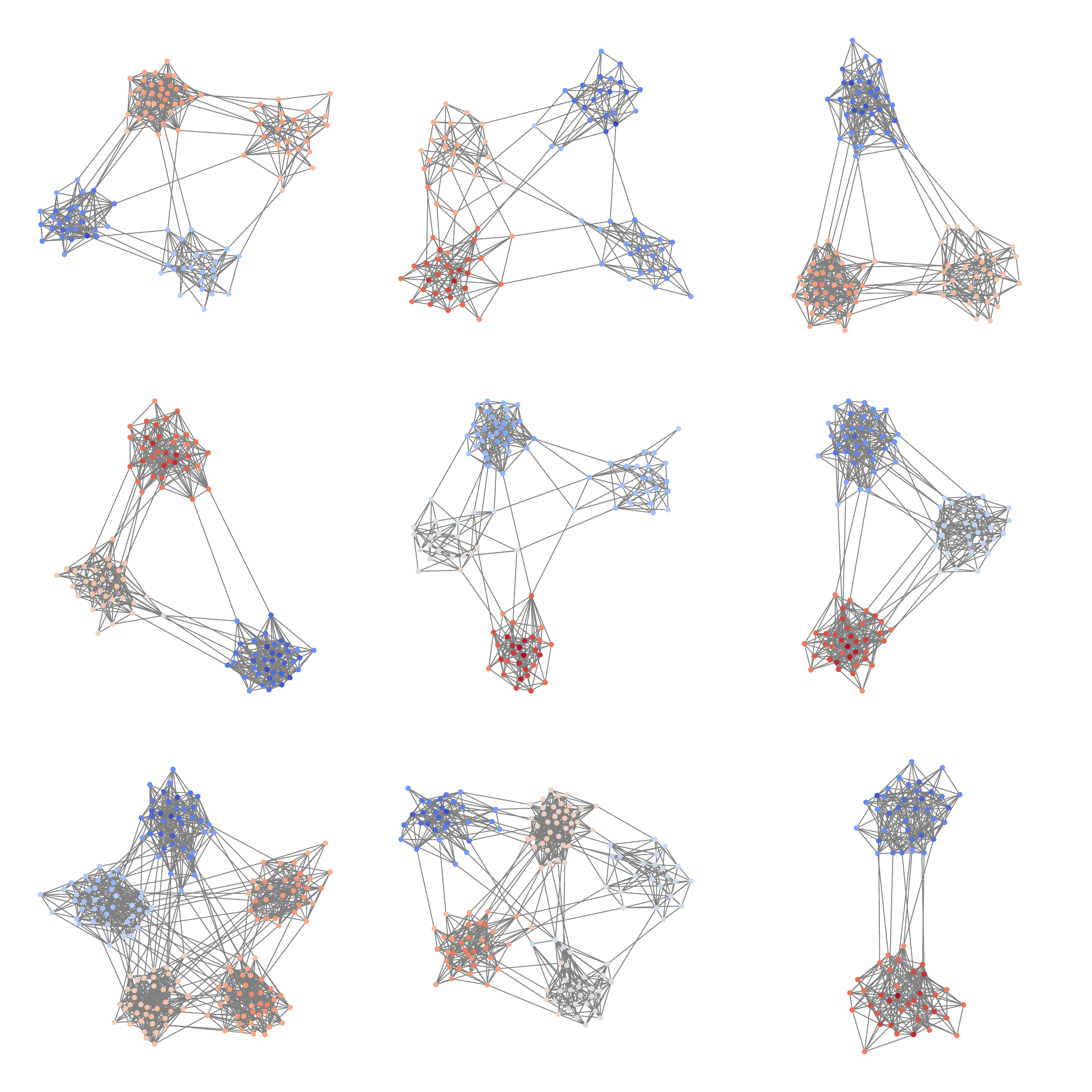}
    \caption{Non-curated samples generated by \method{} (without pre-training) trained on the SBM dataset.}
    \label{app:fig:sbm_samples}
\end{figure}

\begin{figure}[!ht]
    \centering
    \includegraphics[width=0.6\linewidth]{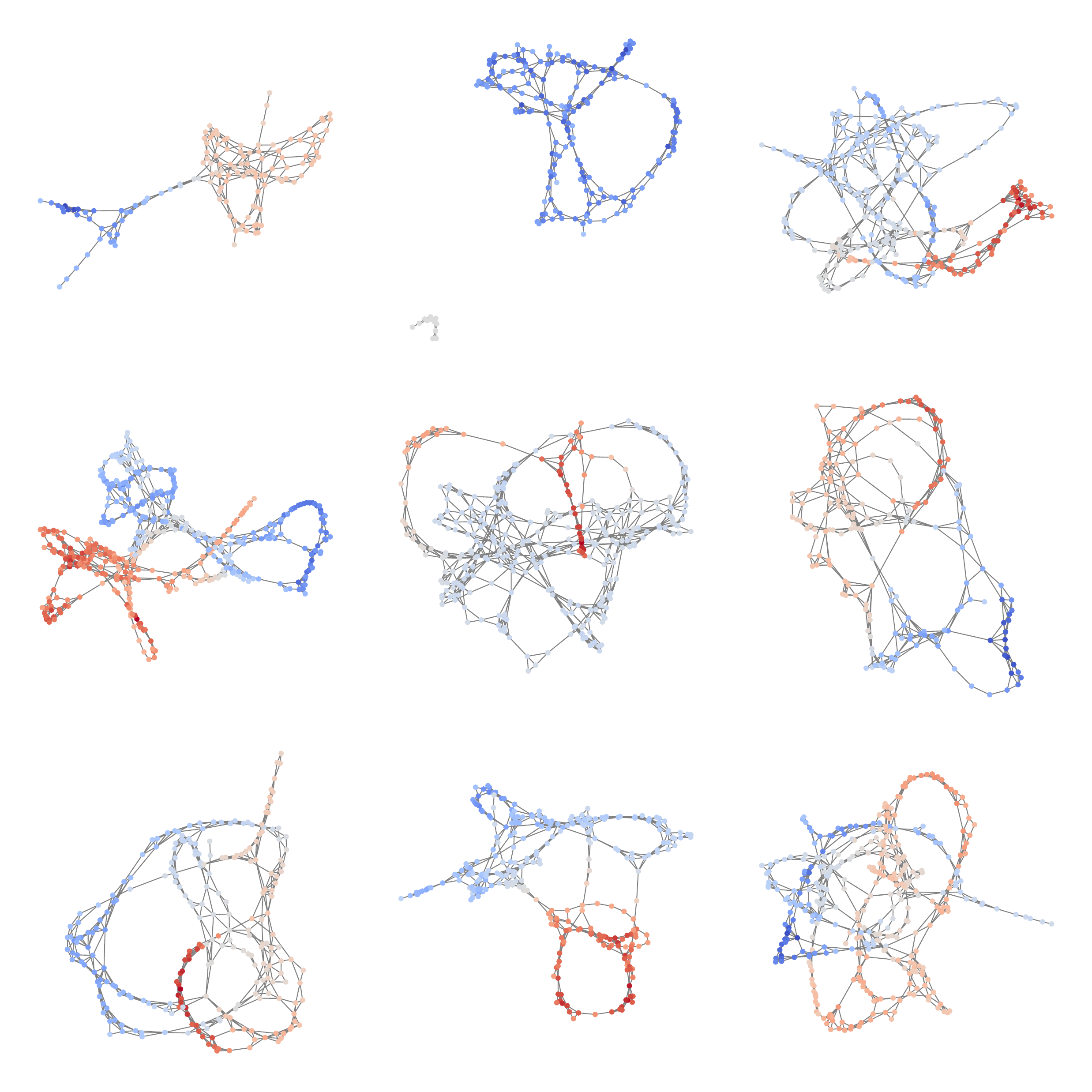}
    \caption{Non-curated samples generated by \method{} (without pre-training) trained on the Proteins dataset.}
    \label{app:fig:protein_samples}
\end{figure}

\begin{figure}[!ht]
    \centering
    \includegraphics[width=0.6\linewidth]{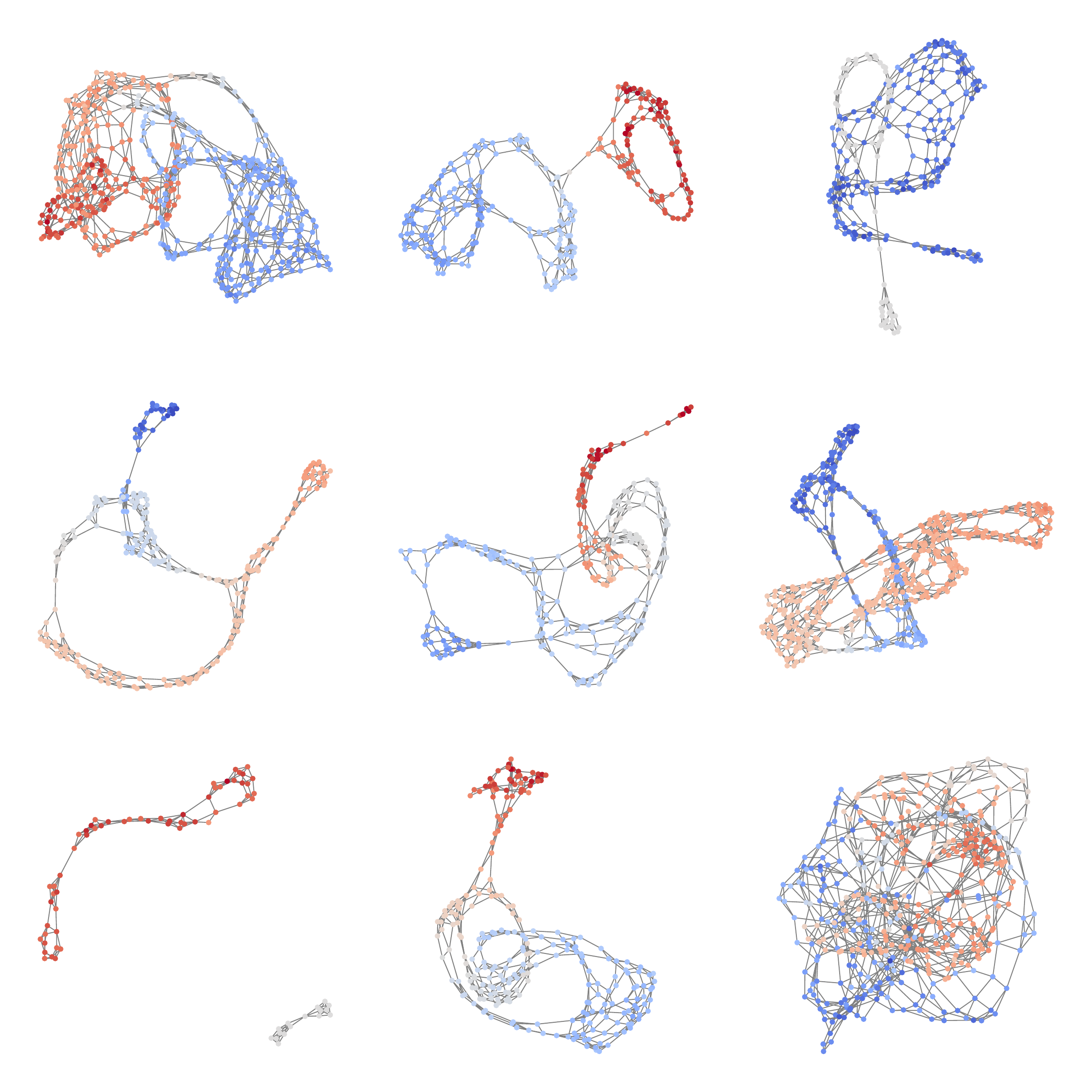}
    \caption{Non-curated samples generated by \method{} (without pre-training) trained on the Point Clouds dataset.}
    \label{app:fig:pointcloud_samples}
\end{figure}

\begin{figure}[!ht]
    \centering
    \includegraphics[width=.8\linewidth]{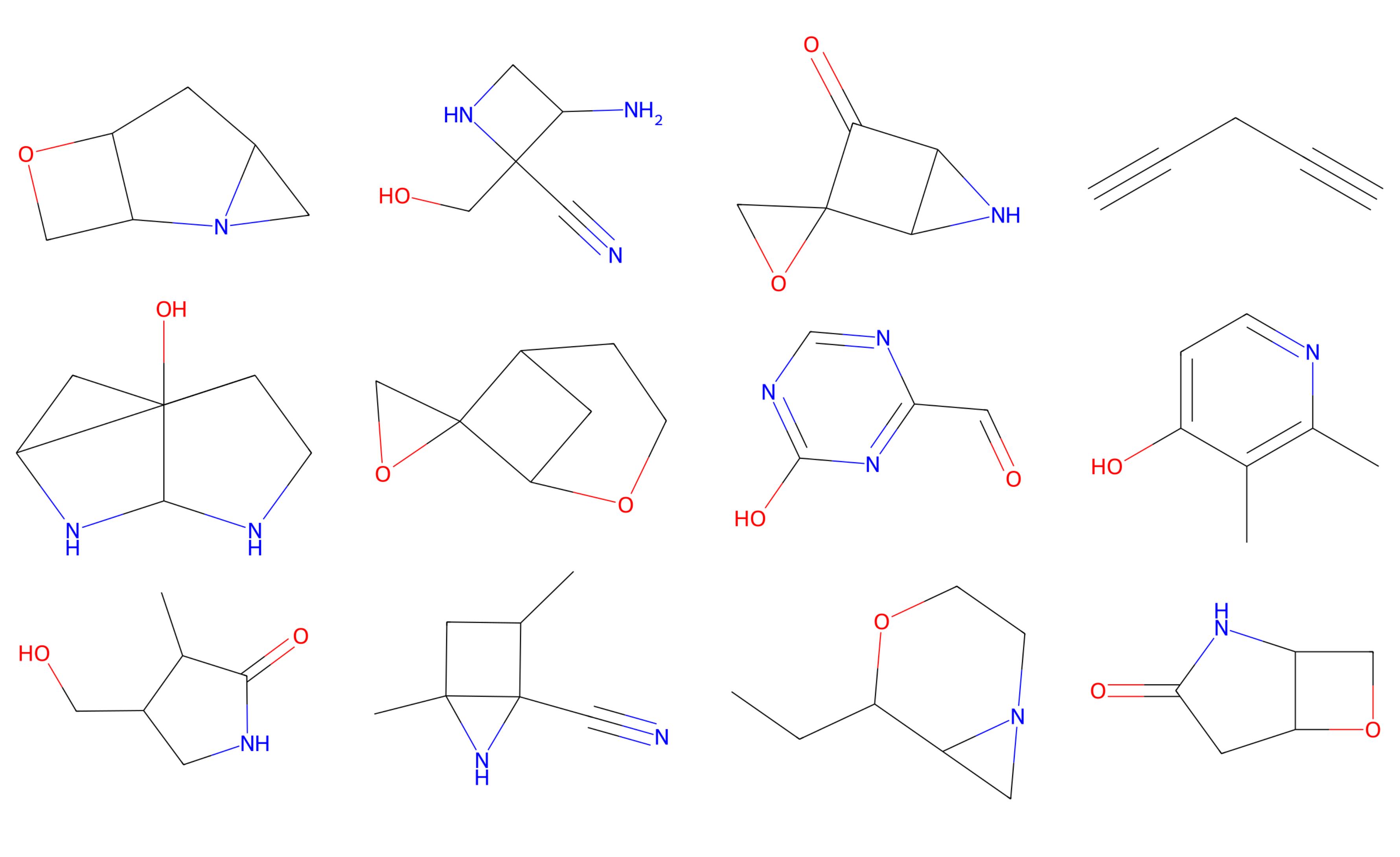}
    \caption{Non-curated samples generated by \method{} (without pre-training) trained on the QM9 dataset.}
    \label{app:fig:qm9_samples}
\end{figure}

\begin{figure}[!ht]
    \centering
    \includegraphics[width=\linewidth]{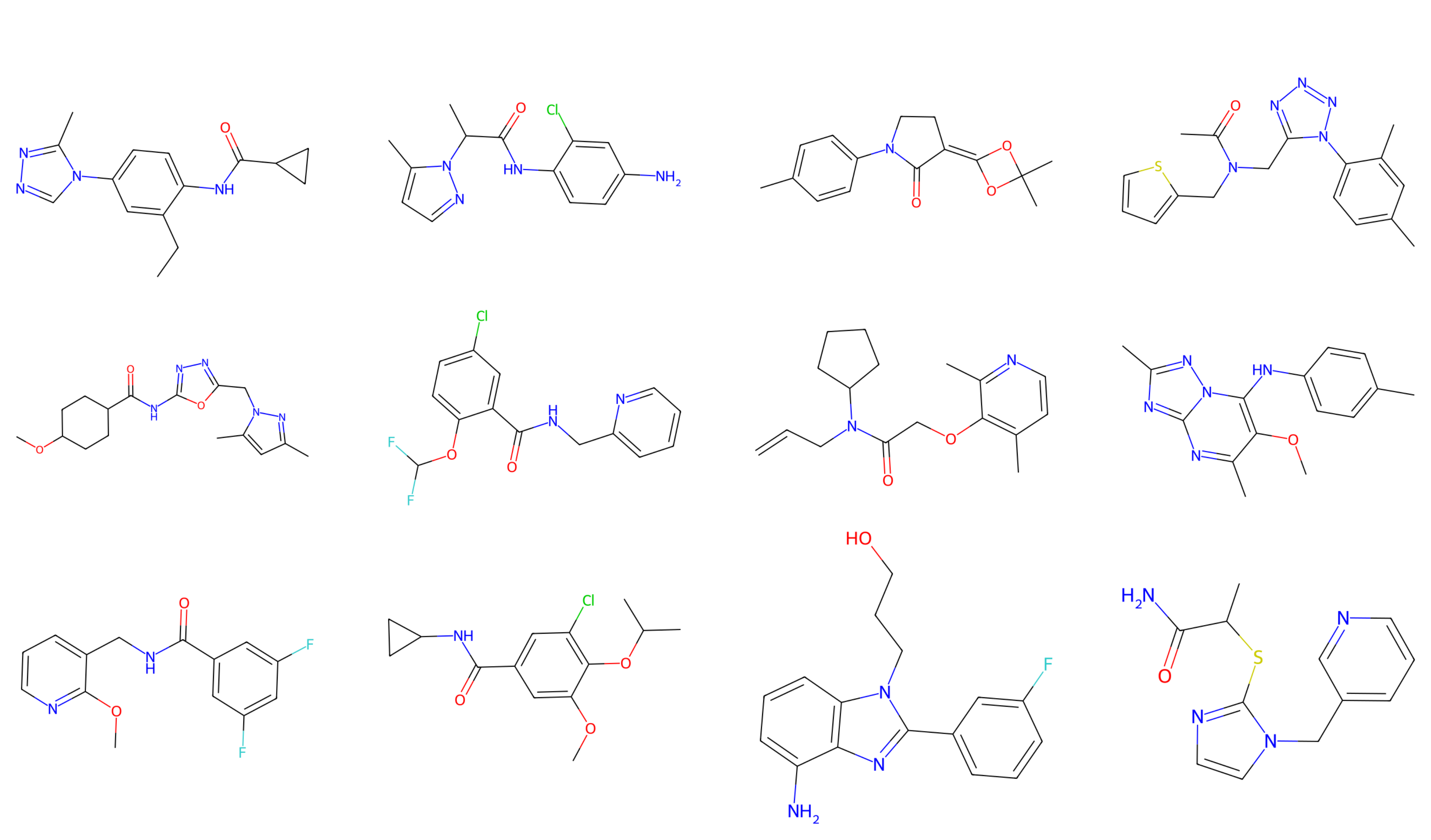}
    \caption{Non-curated samples generated by \method{} (without pre-training) trained on the MOSES dataset.}
    \label{app:fig:moses_samples}
\end{figure}

\begin{figure}[!ht]
    \centering
    \includegraphics[width=\linewidth]{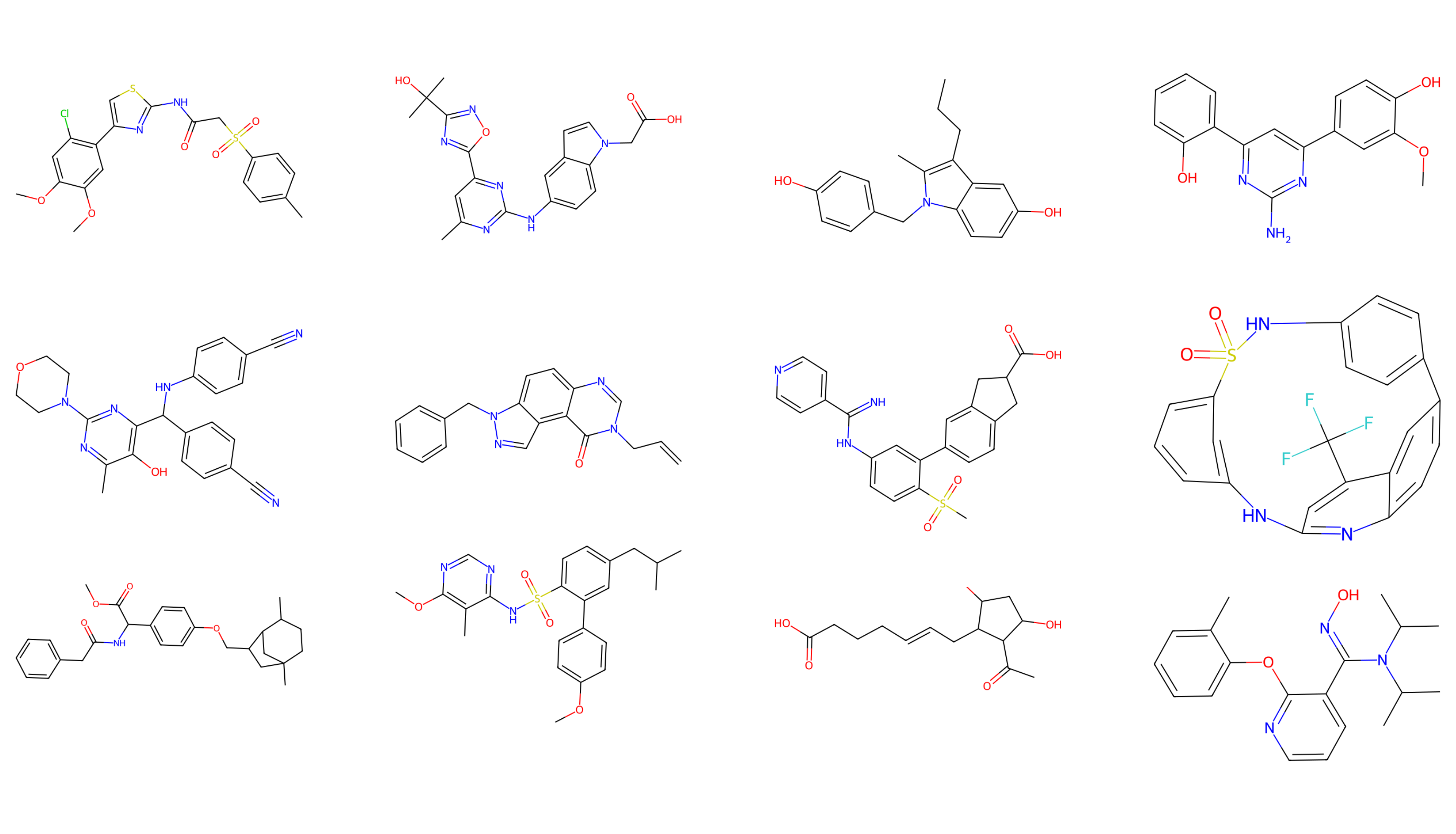}
    \caption{Non-curated samples generated by \method{} (without pre-training) trained on the GuacaMol dataset.}
    \label{app:fig:guacamol_samples}
\end{figure}

\begin{figure}[!ht]
    \centering
    \includegraphics[width=\linewidth]{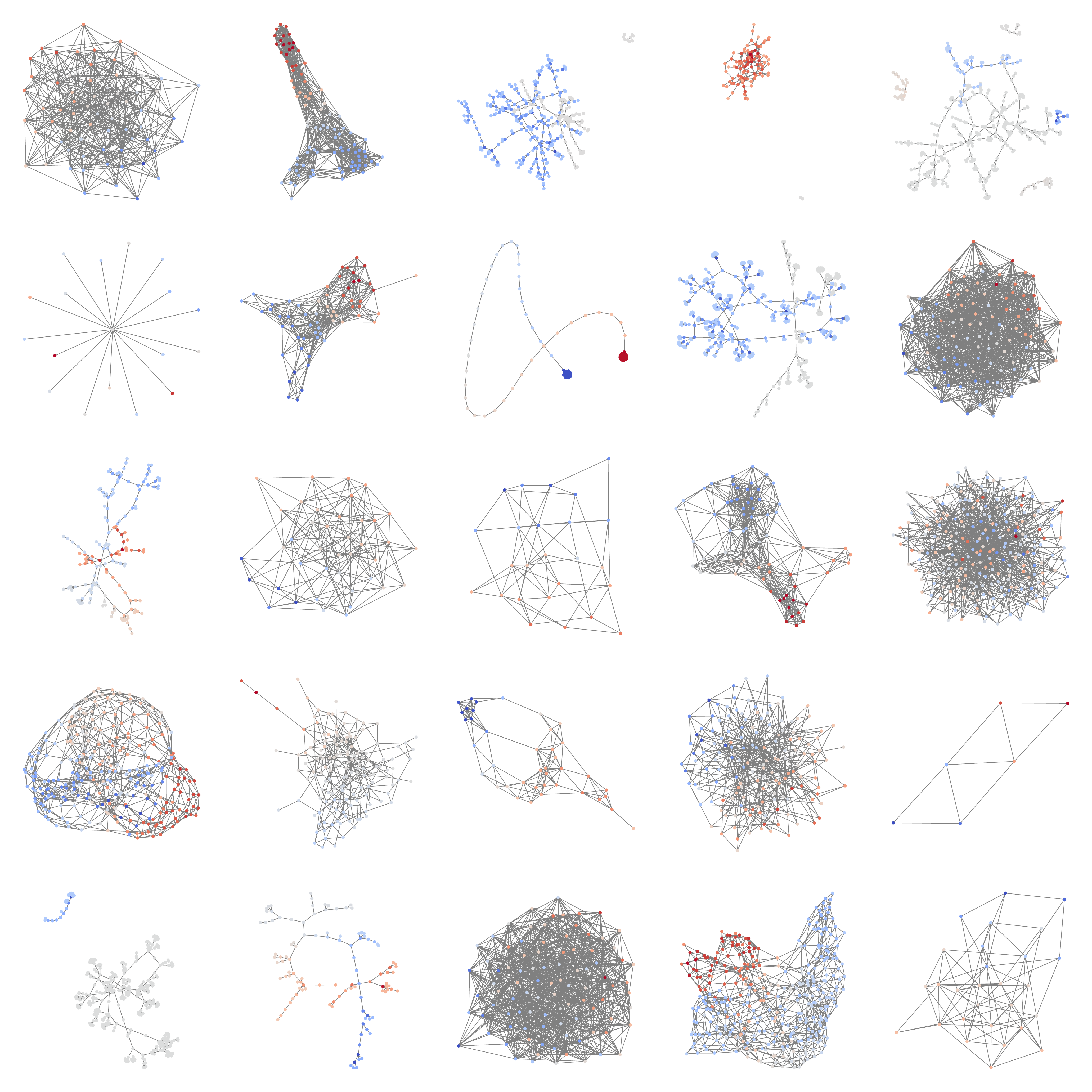}
    \caption{Non-curated samples generated by \method{} trained on the NetworkX dataset.}
    \label{app:fig:networkx_samples}
\end{figure}

We provide visualization of non-curated samples generated by \method{} without pre-training on all datasets in Figure~\ref{app:fig:planar_samples}, \ref{app:fig:sbm_samples}, \ref{app:fig:protein_samples}, \ref{app:fig:pointcloud_samples}, \ref{app:fig:qm9_samples}, \ref{app:fig:moses_samples}, and \ref{app:fig:guacamol_samples}. The results on NetworkX are illustrated in Figure~\ref{app:fig:networkx_samples}. Node colors in unattributed graphs represent the eigenvectors associated with the second-smallest eigenvalues of the graph Laplacian.

\subsubsection{Results with pre-training}
\begin{figure}[!ht]
    \centering
    \includegraphics[width=0.6\linewidth]{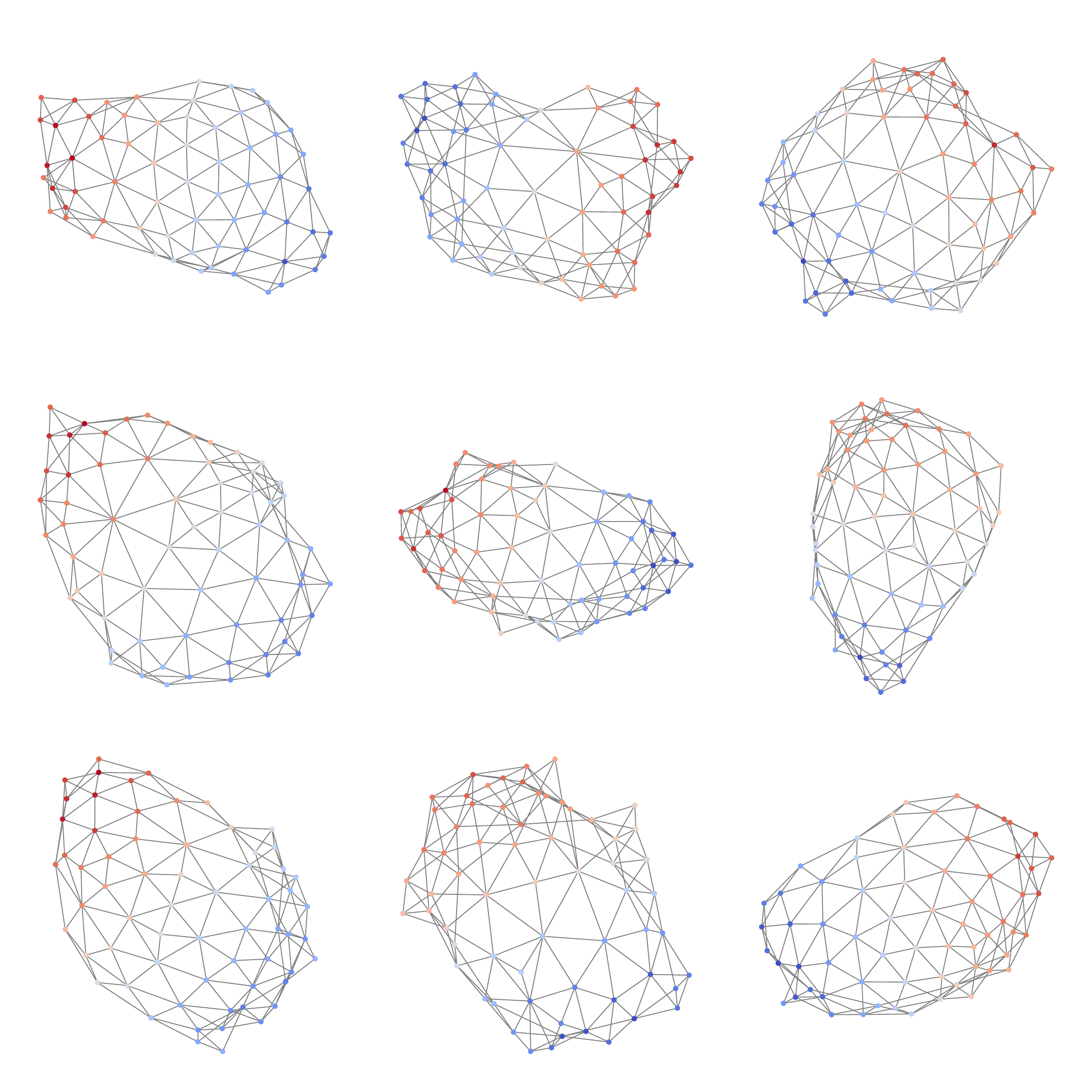}
    \caption{Non-curated samples generated by \method{} (with pre-training on the NetworkX dataset) trained on the Planar dataset.}
    \label{app:fig:planar_samples_pretrained}
\end{figure}

\begin{figure}[!ht]
    \centering
    \includegraphics[width=0.6\linewidth]{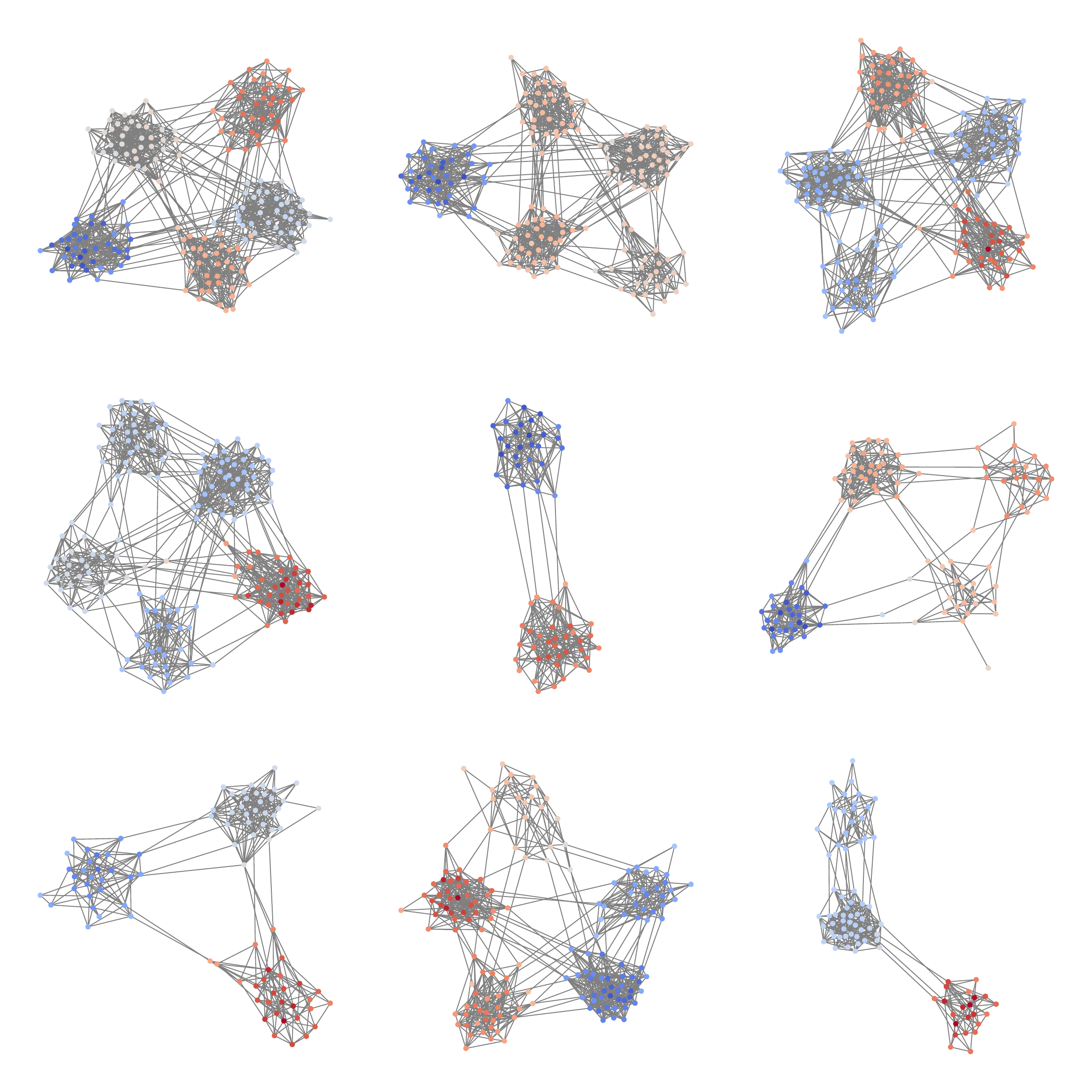}
    \caption{Non-curated samples generated by \method{} (with pre-training on the NetworkX dataset) trained on the SBM dataset.}
    \label{app:fig:sbm_samples_pretrained}
\end{figure}

\begin{figure}[!ht]
    \centering
    \includegraphics[width=0.6\linewidth]{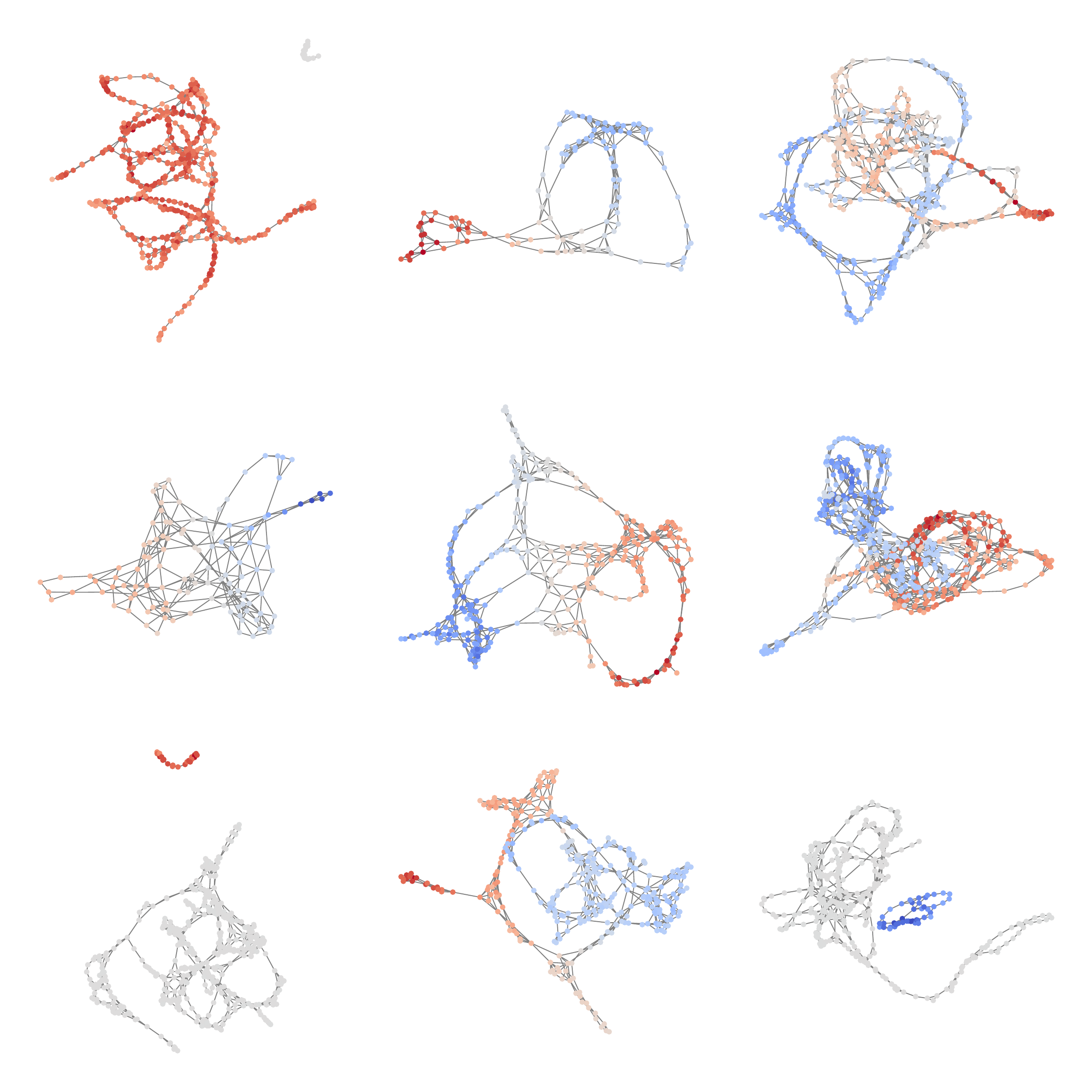}
    \caption{Non-curated samples generated by \method{} (with pre-training on the NetworkX dataset) trained on the Proteins dataset.}
    \label{app:fig:protein_samples_pretrained}
\end{figure}

\begin{figure}[!ht]
    \centering
    \includegraphics[width=0.6\linewidth]{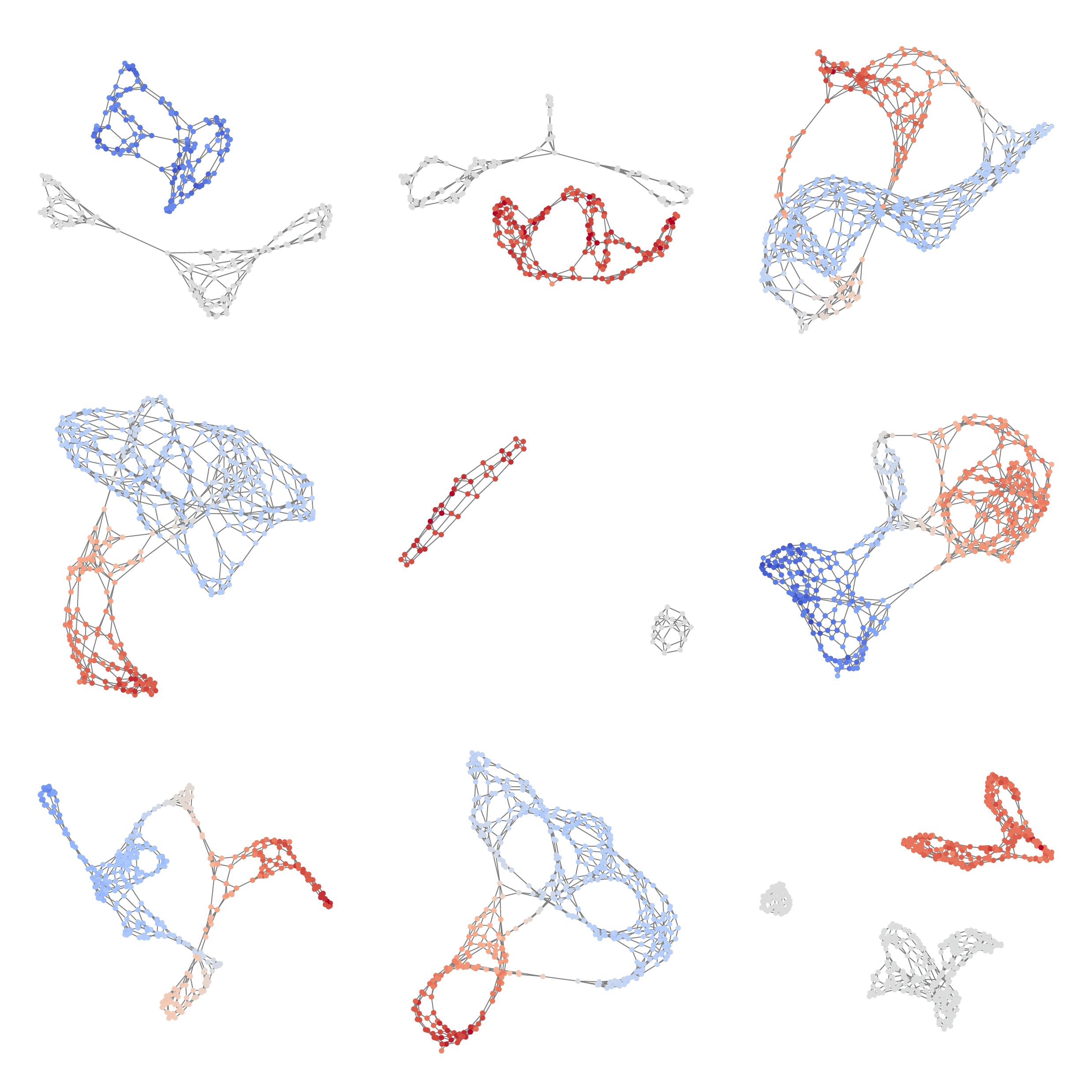}
    \caption{Non-curated samples generated by \method{} (with pre-training on the NetworkX dataset) trained on the Point Clouds dataset.}
    \label{app:fig:pointcloud_samples_pretrained}
\end{figure}

We provide visualization of non-curated samples generated by \method{} with pre-training (on the NetworkX dataset) trained on the non-attributed datasets including Planar, SBM, Proteins, and Point Clouds, illustrated in Figure~\ref{app:fig:planar_samples_pretrained}, \ref{app:fig:sbm_samples_pretrained}, \ref{app:fig:protein_samples_pretrained}, \ref{app:fig:pointcloud_samples_pretrained}.

\end{document}